\documentclass{article}

\usepackage[
  letterpaper,
  textheight=9in,
  textwidth=5.5in,
  top=1in
  ]{geometry}

\usepackage[utf8]{inputenc} % allow utf-8 input
\usepackage[T1]{fontenc}    % use 8-bit T1 fonts
\usepackage{url}            % simple URL typesetting
\usepackage{booktabs}       % professional-quality tables
\usepackage{amsfonts}       % blackboard math symbols
\usepackage{nicefrac}       % compact symbols for 1/2, etc.
\usepackage{microtype}      % microtypography

\usepackage{graphicx} % more modern
\usepackage{subfigure}
\usepackage{amssymb}
\usepackage{amsmath}
\usepackage{amsthm}
\usepackage{times}
\usepackage{helvet}
\usepackage{color}
\usepackage{algorithm}
\usepackage{algorithmic}
\usepackage{multirow}
\usepackage{bbm}
\usepackage{threeparttable}
\allowdisplaybreaks

\title{Stochastic Patching Process}

\author{
  Xuhui Fan\thanks{xhfan.ml@gmail.com}, Bin Li\thanks{libin82cn@gmail.com}, Yi Wang, Yang Wang, Fang Chen \\
  Data61, CSIRO \\
}
\date{}

\makeatletter

\newtheorem{proposition}{Proposition}

\newtheorem{theorem}{Theorem}

\begin{document}

\maketitle

\begin{abstract}
  Stochastic partition models tailor a product space into a number of rectangular regions such that the data within each region exhibit certain types of homogeneity. Due to constraints of partition strategy, existing models may cause unnecessary dissections in sparse regions when fitting data in dense regions. To alleviate this limitation, we propose a parsimonious partition model, named Stochastic Patching Process (SPP), to deal with multi-dimensional arrays. SPP adopts a ``bounding'' strategy to attach rectangular patches to dense regions. SPP is self-consistent such that it can be extended to infinite arrays. We apply SPP to relational modeling and use MCMC sampling for approximate inference. In particular, Conditional-SMC is adopted to sample new patches. The experimental results validate its merit compared to the state-of-the-arts.
\end{abstract}

%==========================================================
\section{Introduction}
%==========================================================

  Stochastic partition processes on a product space have found many real-world applications, such as relational modeling~\cite{kemp2006learning,airoldi2009mixed}, community detection~\cite{nowicki2001estimation,karrer2011stochastic}, collaborative filtering~\cite{porteous2008multi}, and random forests~\cite{LakRoyTeh2014a}. By tailoring the product space into rectangular regions, the partition model aims to fit data using these ``blocks'' such that the data within each block exhibit certain types of homogeneity. As one can choose an arbitrarily fine resolution of partition, the data can be fitted reasonably well.

\begin{figure}[t]
\centering
\includegraphics[width = 0.15 \textwidth, bb = 127 47 461 381, clip]{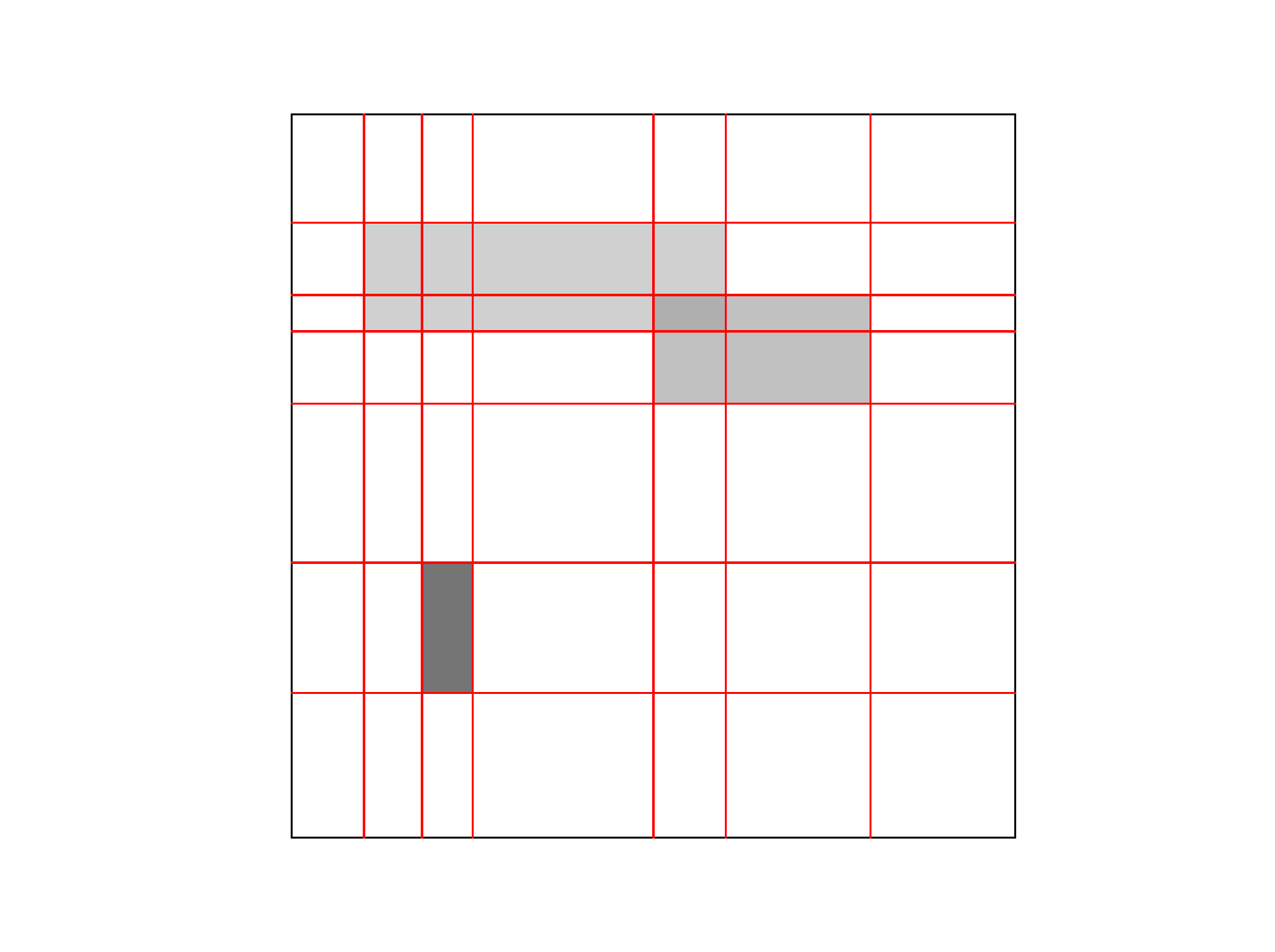} \quad
\includegraphics[width = 0.15 \textwidth, bb = 127 47 461 381, clip]{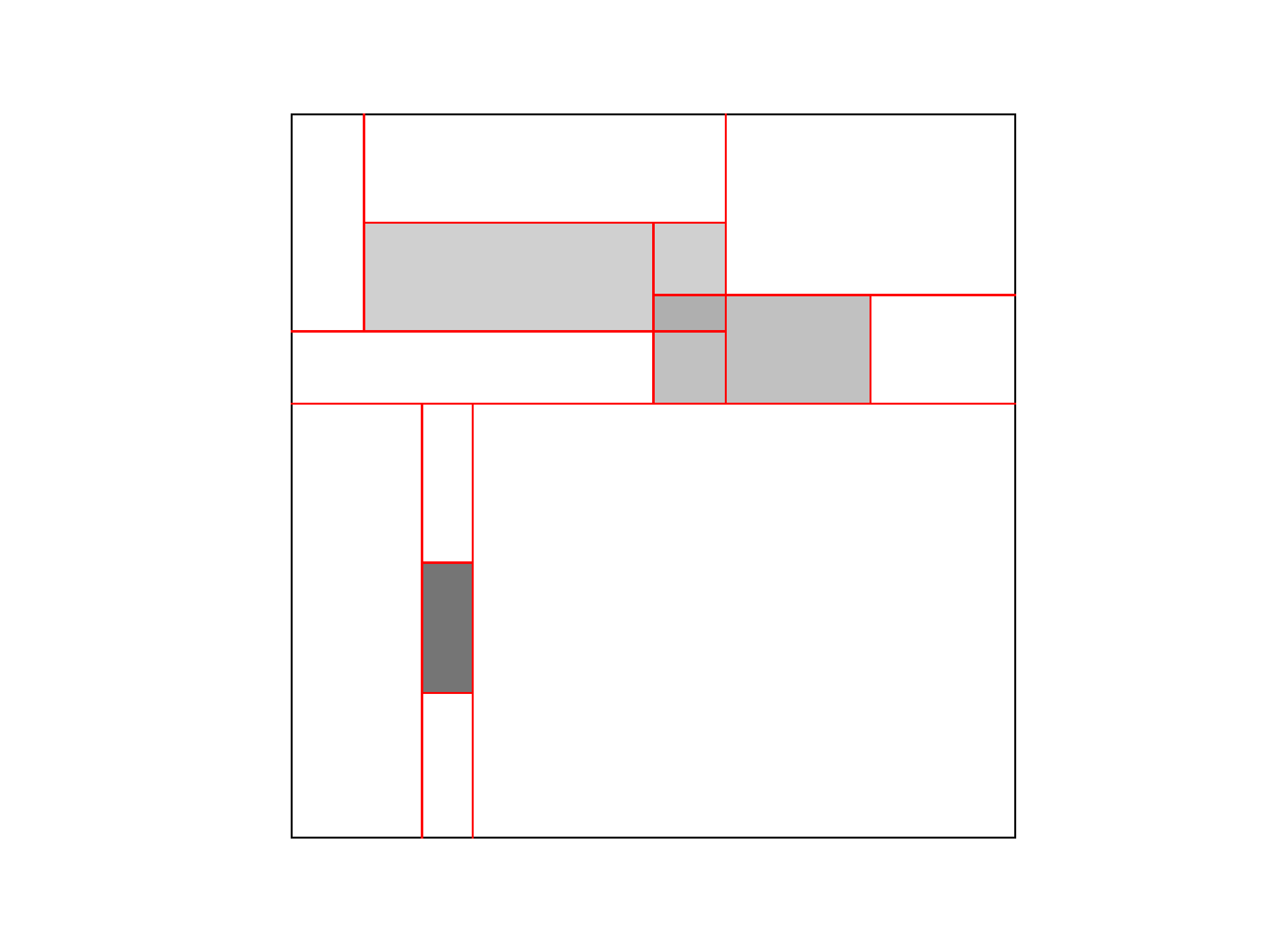} \quad
\includegraphics[width = 0.15 \textwidth, bb = 127 47 461 381, clip]{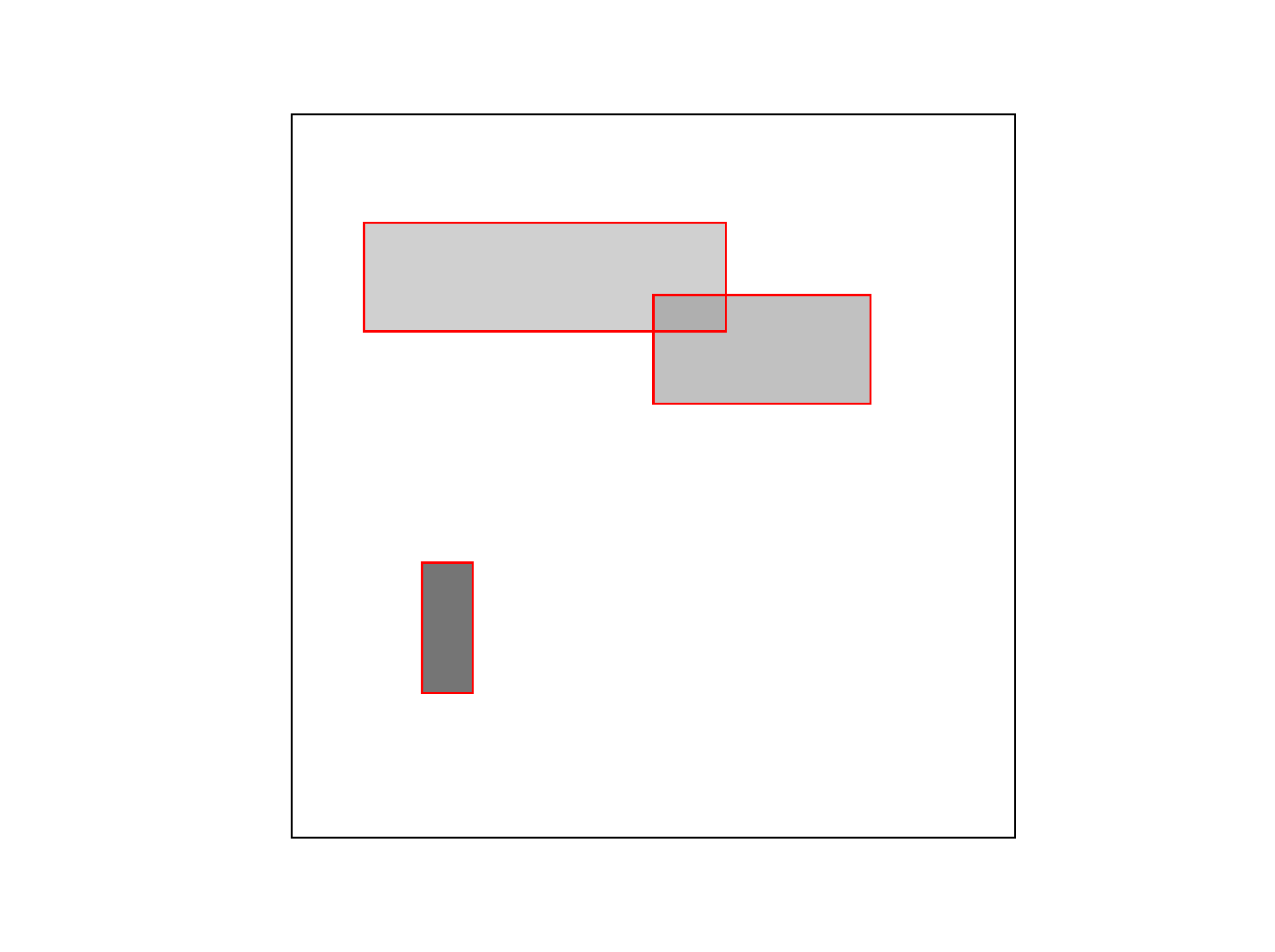}
\caption{(left) Regular-grid partition; (middle) Hierarchical partition; (right) SPP-based partition.}
\label{fig:motivation}
\end{figure}

  The cost of data fitness is that the partition model may cause unnecessary dissections in sparse regions. Compared to regular-grid partitions, the Mondrian process (MP)~\cite{roy2009mondrian} is a hierarchical partition process which has been more parsimonious for data fitting. However, the strategy of recursively cutting the space still cannot avoid unnecessary dissections in sparse regions. Take community detection for example, where a ``block'' corresponds to a community: When tailoring a block out of the relational matrix, cutting-based models will unavoidably separate some uninvolved users. As a result, some meaningless communities are generated as an undesired by-product (see Figure~\ref{fig:motivation} for an illustration).

  Instead of ``cutting'', we propose a bounding-based partition process, named Stochastic Patching Process (SPP), to alleviate the above limitation. SPP attaches patches on a multi-dimensional array to enclose dense regions. In this way, ``significant'' regions of the space can be comprehensively modeled. Each patch can be generated by an outer product of multiple binary vectors, with a segment of consecutive ``1'' entries to indicate the initial and terminal positions of the patch. As patches are independently generated, the layout of patches can be quite flexible. This improves its expressiveness to describe those regions with complicated patterns. As a result, SPP is able to use fewer blocks (thus a more parsimonious expression of model) than those cutting-based partition models to achieve similar modeling capability.

  An important property of SPP is self-consistency. This means that, by restricting the patches generated from an SPP on a multi-dimensional array $Y$ to its sub-array $X$, the resulting patches are distributed as if they are directly generated on $X$ through another SPP (given the same budget). The property will be verified in three steps: (1) the number distribution of nonempty patches is self-consistent; (2) the position distribution of a nonempty patch is self-consistent; (3) based on the above, the self-consistency of SPP can be verified. This property suggests that SPP can be extended to infinite multi-dimensional arrays according to the Kolmogorov extension theorem~\cite{bochner1955harmonic}.

  The merit of SPP can be seen in many applications. In this paper we investigate its merit in relational modeling, where patches can be viewed as communities while the ``price'' of a patch (cost per unit area) can be viewed as interacting intensity within the community. An MCMC based approximate inference method is proposed for the SPP relational model. In particular, Conditional-Sequential Monte Carlo~\cite{andrieu2010particle} is adopted to sample completely new patches in each iteration. The experimental results on a number of real-world relational data sets demonstrate that SPP can achieve parsimonious partitions with competitive performance compared to the state-of-the-arts.

%==========================================================
\section{Preliminaries} \label{sec:rw}
%==========================================================

\subsection{Stochastic Partition Processes}

  Stochastic partition processes partition a product space into blocks. A popular application of such processes is modeling relational data such that the intensity of interactions is homogeneous within each block. In terms of partitioning strategy, state-of-the-art stochastic partition processes can be roughly categorized into regular-grid partitions and flexible axis-aligned partitions.

  A regular-grid stochastic partition process is constituted by two separate partition processes on each dimension of the multi-dimensional array. The resulting orthogonal interactions between two dimensions will exhibit regular grids, which can represent interacting intensities. Typical regular-grid partition models include the infinite relational model (IRM)~\cite{kemp2006learning} and the infinite extension of mixed-membership stochastic blockmodels~\cite{airoldi2009mixed}. Regular-grid partition models are widely used in real-world applications for modeling graph data~\cite{ishiguro2010dynamic,nonpa2013schmidt}.

  To our knowledge, only the Mondrian process (MP)~\cite{roy2009mondrian,roy2011thesis} and the rectangular tiling process (RTP)~\cite{nakano2014rectangular} can produce flexible axis-aligned partitions on a product space. MP recursively generates axis-aligned cuts on a unit hypercube and partitions the space in a hierarchical fashion known as $k$d-tree (\cite{roy2007learning} also considers a tree-consistent partition model but it is not Bayesian nonparametric). Different from the hierarchical partitioning strategy, RTP generates a flat partition structure on a two-dimensional array by assigning each entry to an existing block or a new block in sequence, without violating the rectangular restriction on the blocks.

\subsection{Kolmogorov Extension Theorem} \label{sub:kolmogorov}

  We consider a projective system of stochastic partitions: Let $\{(\Omega_X, \mathcal{B}_X)\}_{X \in\mathcal{F}(\mathbb{N}^D)}$ be a family of measurable spaces, where $\Omega_X$ is the partition space, $\mathcal{B}_X$ is a $\sigma$-algebra on $\Omega_X$, and $\mathcal{F}(\mathbb{N}^D)$ denotes the collection of all finite sub-arrays of the infinite $D$-dimensional array $\mathbb{N}^D$. For each $X \in \mathcal{F}(\mathbb{N}^D)$, $P_{\boxplus}^X$ is a probability measure on $\mathcal{B}_X$. $\mathcal{F}(\mathbb{N}^D)$ is a partially ordered set; while $X \preccurlyeq Y \in \mathcal{F}(\mathbb{N}^D)$ the projection $\pi_{Y,X}$ restricts the partition $\boxplus_Y$ on $Y$ into $X$, by keeping $\boxplus_Y$'s entries within $X$ unchanged and removing the remaining entries. For $B_X \in\mathcal{B}_X$, the pre-image under projection is defined as $\pi_{Y,X}^{-1} B_X = \{\boxplus_Y \in \Omega_Y |\pi_{Y,X} \boxplus_Y \in B_X\}$ and the projection also satisfies $\pi_{Y,X} \circ \pi_{X,W}=\pi_{Y,W}, W \preccurlyeq X \preccurlyeq Y$. This family defines the projective limit measurable space $(\Omega_{\mathbb{N}^D}, \mathcal{B_{\mathbb{N}^D})}$.
\begin{theorem}[Theorem 3.3.6 in~\cite{chung2001course}]
  For a set of probability spaces $\{(\Omega_X, \mathcal{B}_X, P_{\boxplus}^X\}_{X \in \mathcal{F}(\mathbb{N}^D)}$ such that projection $\pi_{Y,X}: \Omega_Y \to \Omega_X, X \preccurlyeq Y \in \mathcal{F}(\mathbb{N}^D)$ and $P_{\boxplus}^Y(\pi_{Y,X}^{-1}B_X) = P_{\boxplus}^X(B_X)$ holds for all $B_X \in\mathcal{B}_X$. Then $P_{\boxplus}^X$ can be uniquely extended to measure $P_{\boxplus}^{\mathbb{N}^D}$ on $(\Omega_{\mathbb{N}^D}, \mathcal{B}_{\mathbb{N}^D})$ as the projective limit measurable space.
\end{theorem}
  The Kolmogorov extension theorem provides us a constructive way to extend SPP to the infinite $D$-dimensional array $\mathbb{N}^D$, which will be discussed in Section~\ref{sec:sc}.

\subsection{Exchangeable Arrays} \label{sub:graphon}

  The Aldous--Hoover theorem~\cite{hoover1979relations,aldous1981representations} provides the theoretical foundation to model exchangeable multi-dimensional arrays conditioned on a stochastic partition model. A random 2-dimensional array is called separately exchangeable if its distribution is invariant under separate permutations of rows and columns.
\begin{theorem}[Theorem 3.2 in~\cite{TPAMI2014peterdaniel}] \label{theoremgraphon}
  A random array $(R_{ij})$ is separately exchangeable if and only if it can be represented as follows: There exists a random measurable function $F:[0,1]^3 \mapsto \mathcal{X}$ such that $(R_{ij}) \overset{d}{=} \left(F(\xi_i^{\textrm{row}}, \eta_j^{\textrm{col}}, \nu_{ij})\right)$, where $\{\xi_i^{\textrm{row}}\}_i, \{\eta_j^{\textrm{col}}\}_j$ and $\{\nu_{ij}\}_{i,j}$ are, respectively, two sequences and an array of i.i.d. uniform random variables in $[0,1]$.
\end{theorem}

  Relational modeling based on a stochastic partition model is a typical application of the Aldous--Hoover theorem. By defining $W(\xi_i^{\textrm{row}},\eta_j^{\textrm{col}}) := P(F(\xi_i^{\textrm{row}}, \eta_j^{\textrm{col}}, \nu_{ij})=1|F)$, every exchangeable array can be represented by a random graph function~\cite{TPAMI2014peterdaniel}. The SPP relational model introduced in Section~\ref{sec:rm} is implicitly built on this theorem.

%==========================================================
\section{Stochastic Patching Process} \label{sec:spp}
%==========================================================

  SPP is defined on a measurable space $(\Omega_X, \mathcal{B}_X),X \in \mathcal{F}(\mathbb{N}^D)$. Each element in $\Omega_X$ denotes a partition $\boxplus_X$, constituted by a collection of rectangular \emph{nonempty} patches $\{\Box_k\}_k$ with corresponding costs $\{m_k\}_k$, where $k \in \mathbb{N}$ indexes the patch number in $\boxplus_X$. In particular, a patch is defined by an outer product $\Box_k := \bigotimes_{d=1}^D u_k^{(d)}$, where $u_k^{(d)} \in \{0,1\}^{N_X^{(d)}}$ ($N_X^{(d)}$ denotes the length of the $d$th dimension in $X$) is a position indicator vector for the $d$th dimension of $\Box_k$, with the constraint that $u_k^{(d)}$ only comprises a segment of $l_k^{(d)} \in \{1,\ldots,N_X^{(d)}\}$ consecutive ``1'' entries which starts at an initial position $s_k^{(d)} \in \{1,\ldots,N_X^{(d)}\}$.

  Given an array $X$ and a budget $\tau$, we can sample a random partition from an SPP: $\boxplus_X \sim \mbox{SPP}(X,\tau)$. We assume that the costs of patches are \emph{i.i.d.} sampled from the same exponential distribution, which implies there exists a homogeneous Poisson process on the time (cost) line. The generating time of each patch is uniform in $(0,\tau]$ and the number of patches has a Poisson distribution. We represent a random partition as $\boxplus_X := \{m_k,\Box_k\}_{k=1}^{K_\tau} \in \Omega_X$, which is generated as follows\footnote{An equivalent construction is to directly generate \emph{nonempty} patches through thinning the Poisson process which is used for generating candidate patches -- See ``Alternative Construction of SPP'' in Appendix.}:
\begin{enumerate}
  \item Sample the number of candidate patches $\hat{K}_{\tau}\sim\mbox{Poisson}(\tau S_X)$, where $S_X=\prod_{d=1}^D N_X^{(d)}$;
  \item Sample $\hat{K}_{\tau}$ \emph{i.i.d.} candidate patches. For $k' = 1,\ldots,\hat{K}_{\tau}$, $d = 1,\ldots,D$
\begin{enumerate}
  \item Sample the initial position $s_{k'}^{(d)}$ of the $k'$th candidate uniformly from $\{1, \cdots, N_X^{(d)}\}$;
  \item If $s_{k'}^{(d)}=1$, the side-length $l_{k'}^{(d)}$ increments from 0 to 1 with probability $1$; otherwise $l_{k'}^{(d)}$ increments from 0 to 1 with probability $(1-\theta)$, where $\theta\in[0,1]$;
  \item If $l_{k'}^{(d)}$ has incremented from 0 to 1, generate the side-length using the distribution
      %\footnote{{\color{red}$P(l_{k'}^{(d)})$ follows a geometric distribution for $1\le l_{k'}^{(d)} < L_*$; while the probability for $l_{k'}^{(d)} = L_*$ is the sum of the probability mass of the geometric distribution for $l_{k'}^{(d)} \ge L_*$.}}
\begin{eqnarray}
  P(l_{k'}^{(d)})=\left\{\begin{array}{ll}
  \theta^{l_{k'}^{(d)}-1}(1-\theta), & 1\le l_{k'}^{(d)} < L_*; \\
  \theta^{l_{k'}^{(d)}-1}, & l_{k'}^{(d)}= L_*, \\
  \end{array}\right. \nonumber
  \end{eqnarray}
  where $L_*=N_X^{(d)}-s_{k'}^{(d)}+1$;
\end{enumerate}
\item Remove all empty patches and retain $K_\tau$ \emph{nonempty} patches $\{\Box_k|S_{\Box_k}=\prod_{d=1}^D l_k^{(d)}>0\}_{k=1}^{K_\tau}$. Sample $K_{\tau}$ \emph{i.i.d.} time points uniformly in $(0,\tau]$ and index them to satisfy $t_1<\ldots<t_{K_\tau}$. Set the cost of $\Box_k$ as $m_k = t_{k}-t_{k-1}$ ($t_0=0$) and the rate\footnote{In Section~\ref{sec:rm} we will show that such rate is useful when we use $\Box_k$ and $\omega_k$ as the priors of a community and the intensity of interactions within the community (large communities have relatively weak interactions).} of $\Box_k$ as $\omega_k = m_k/S_{\Box_k}$.
\end{enumerate}
  We use the initial position $s_k^{(d)}$ and the side-length $l_k^{(d)}$ of $\Box_k$ in the $d$th dimension to determine the position indicator vector $u_k^{(d)} \in \{0,1\}^{N_X^{(d)}}$, which further constitutes $\Box_k$. Thus, given $K_\tau$, all patches are \emph{i.i.d.} generated and the layout of patches can be quite flexible. Patches can be overlapped or even contained by others. In the two-dimensional case, a partition sampled from an SPP has the following interpretation: $X$ can be viewed as a piece of cloth while $\Box_k$ can be viewed as a patch; the material of $\Box_k$ has rate (price) $\omega_k$ and the number of ``patches'' on the ``cloth'' is determined by $S_X$ and the budget $\tau$ -- This is where the name of ``Stochastic Patching Process'' comes from.
  %The rate of an overlapped part are synthesized by the rates of the involved patches.

  SPP has some favorable properties for being used to manipulate the prior partition via the hyper-parameters. Given array $X$, one can use the budget $\tau$ to control the expected ``volume'' covered by the patches on $X$ (overlapped parts are counted separately). Given array $X$ and budget $\tau$, the expected volume is a fixed constant, $\mathbb{E}(K_{\tau})\cdot \mathbb{E}(S_\Box) = constant$, and the expected number of patches $\mathbb{E}(K_{\tau})$ and the expected volume of each patch $\mathbb{E}(S_\Box)$ can be balanced through $\theta$. Formally, we have the following property (see Appendix for the proof).

\begin{proposition}
  Let $\boxplus_X := \{m_k,\Box_k\}_{k=1}^{K_\tau} \sim \mbox{SPP}(X,\tau)$, the expected volume of all patches is a constant in terms of the budget $\tau$ and the volume of the array $X$, that is, $\mathbb{E}(K_{\tau})\cdot \prod_{d=1}^D \mathbb{E}(l^{(d)}) = \tau\cdot \prod_{d=1}^D N_X^{(d)}$.
\end{proposition}

\begin{figure}[t]
    \centering
    \includegraphics[width = 0.25 \textwidth, bb = 174 43 462 361, clip]{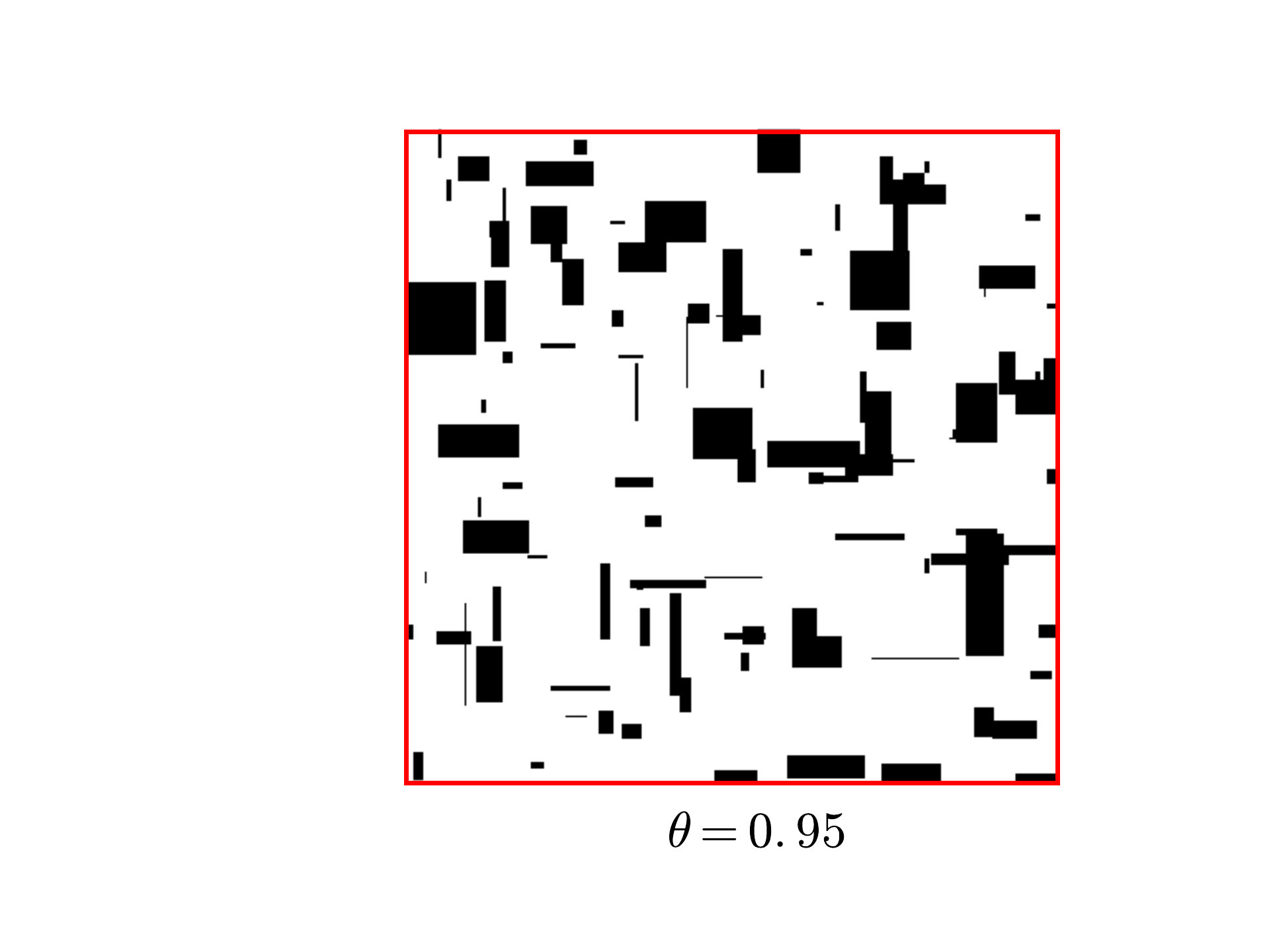} \quad
    \includegraphics[width = 0.25 \textwidth, bb = 174 43 462 361, clip]{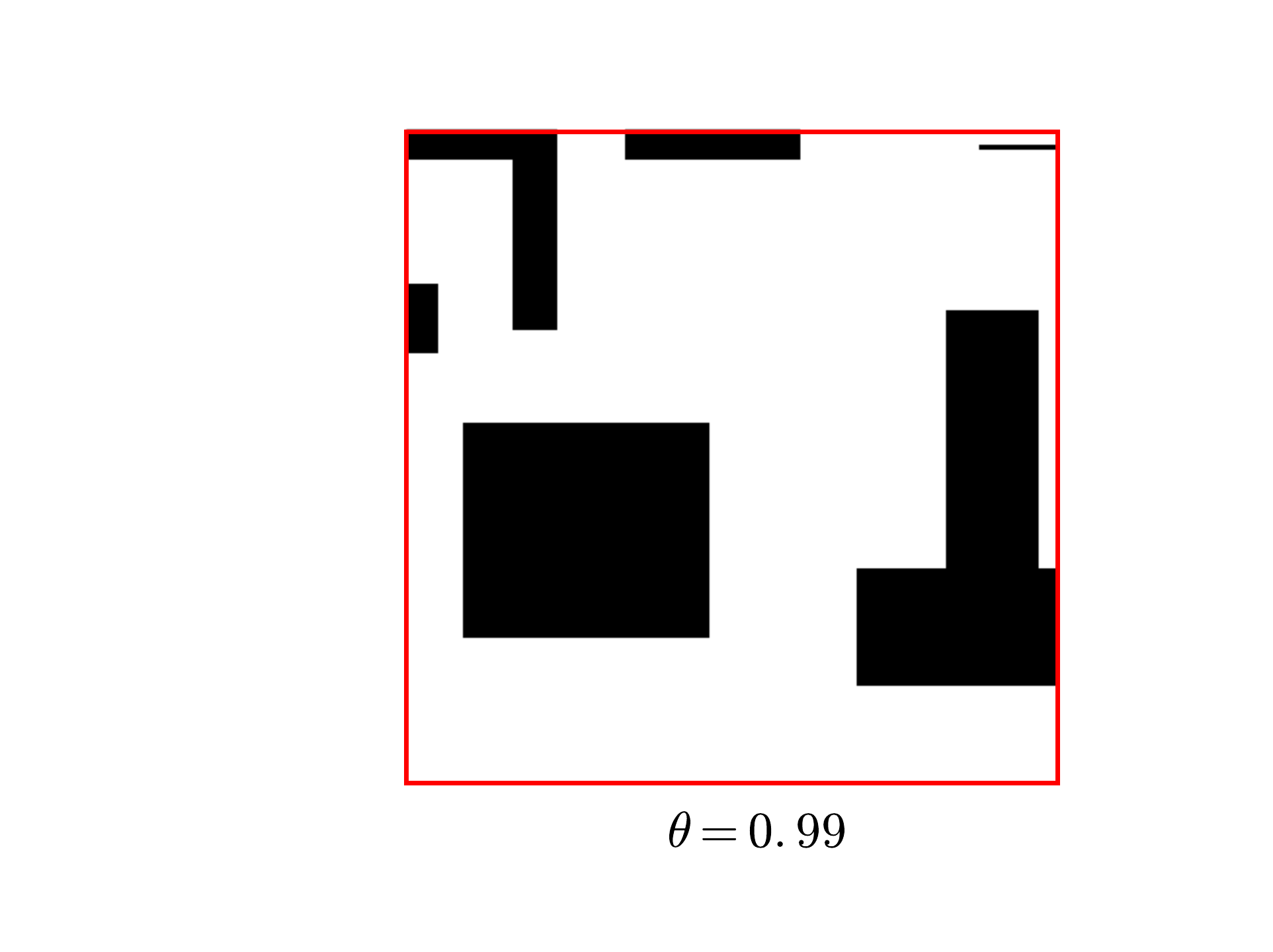}
    \caption{Two example SPP-based prior partitions on a 2-array ($N_X^{(1)}=N_X^{(2)}=500, \tau = 0.2$). In the case of $\theta=0.95$ (left) many small patches are generated while in the case of $\theta=0.99$ (right) few large patches are generated.}
    \label{fig:thetainfluence}
    \end{figure}

  Figure~\ref{fig:thetainfluence} gives an illustration of the above property based on a toy two-dimensional array, with fixed value of $\tau$ but two different values of $\theta$. We can see that, in both cases, the volumes (black areas) covered by the patches are similar while the number of nonempty patches varies inversely to the average side-length of patch -- A larger (smaller) value of $\theta$ results in a longer (shorter) expected side-length and a smaller (larger) number of nonempty patches.

  Due to such a flexible layout of patches, SPP is parsimonious to model multi-dimensional arrays, especially in sparse scenarios -- SPP is able to describe ``significant'' parts of the array (e.g. active communities in a social network) through small patches; after patching these ``significant'' parts, the rest are usually large and irregular sparse areas which may be neglected.

%==========================================================
\section{Self-Consistency} \label{sec:sc}
%==========================================================

  Section~\ref{sec:spp} has defined SPP on a finite array given a budget. To further extend SPP to the infinite array $\mathbb{N}^D$, an essential property of SPP is self-consistency. That is to say, while restricting an SPP on a finite $D$-dimensional array $Y$, say $\mbox{SPP}(Y,\tau)$, to its sub-array $X$, $X \subset Y \in \mathcal{F}(\mathbb{N}^D)$, the resulting patches restricted to $X$ are distributed as if they are directly generated on $X$ through $\mbox{SPP}(X,\tau)$. A typical application scenario is social network, where $X$ and $Y$ successively represent two snapshots of a growing network.

  The self-consistency property is verified in three steps: (1) the number distribution of nonempty patches is self-consistent; (2) the position distribution of a nonempty patch is self-consistent; (3) SPP is self-consistent. Following the notations used in Sections~\ref{sub:kolmogorov} and~\ref{sec:spp}, we use $\pi_{Y,X}$ to denote the projection that restricts $\boxplus_Y \in \Omega_Y$ to $X$ by keeping $\boxplus_Y$'s entries in $X$ unchanged and removing the rest. An ``empty patch'' is referred to the case $S_{\Box}=0$ ($\exists d, l^{(d)}=0$), where $S_\Box$ denotes the volume of the candidate patch.

\subsection{Number of Nonempty Patches}

\begin{proposition} \label{expectednumber}
  While restricting $\mbox{SPP}(Y,\tau)$ to $X$, $X \subset Y \in \mathcal{F}(\mathbb{N}^D)$, the time points of nonempty patches crossing into $X$ from $Y$ follows the same Poisson process for generating the time points of nonempty patches in $\mbox{SPP}(X,\tau)$.
\end{proposition}

  According to the definition, the candidate patches sampled from $\mbox{SPP}(Y,\tau)$ (or $\mbox{SPP}(X,\tau)$) follows a homogeneous Poisson process with intensity $S_Y$ (or $S_X$). Since there exists possibility to generate empty patches, we use intensity $S_X\cdot P(S_{\Box^X}>0)$ for thinning the Poisson process to generate nonempty patches. Given the same budget $\tau$, Proposition~\ref{expectednumber} holds if we can prove the following equality of the two Poisson process intensities
\begin{eqnarray} \label{eq:poissonrateequality}
  S_Y\cdot P(S_{\pi_{Y,X}(\Box^Y)}>0) = S_X\cdot P(S_{\Box^X}>0)
\end{eqnarray}

  Due to the independence of dimensions, we have $P(S_{\Box^X}>0)=\prod_d P(l^{(d)}_{X}>0)$. W.l.o.g, we assume that the two arrays, $X$ and $Y$, have the same shape apart from the $d'$th dimension where $Y$ has one additional column (the general case of more columns follows by induction). Then we can discuss two cases: $X$ and $Y$ 1) share the terminal boundary or 2) share the initial boundary in the $d'$th dimension. Eq.~(\ref{eq:poissonrateequality}) can be proved in both cases (see Appendix for the complete proof).

  Because of the same Poisson process intensity in Eq.~(\ref{eq:poissonrateequality}), the following equality also holds
\begin{eqnarray}
  P^Y_{K_{\tau}, \{m_k\}_k}\left(\pi_{Y,X}^{-1}\left(K_{\tau}^{X}, \{m_{k}^{X}\}_{k=1}^{K_{\tau}^{X}}\right)\right) 	
  =P^X_{K_{\tau}, \{m_k\}_k}\left(K_{\tau}^{X}, \{m_{k}^{X}\}_{k=1}^{K_{\tau}^{X}}\right) \label{k_tau_m_k}
\end{eqnarray}

\subsection{Position of Nonempty Patches}

\begin{proposition} \label{expecteddistribution}
  While restricting $\mbox{SPP}(Y,\tau)$ to $X$, $X \subset Y \in \mathcal{F}(\mathbb{N}^D)$, the marginal probability of the pre-images of a nonempty patch $\Box^X$ in $Y$ (given the patches in $Y$ crossing into $X$ that are nonempty) equals to the probability of $\Box^X$ directly sampled from $\mbox{SPP}(X,\tau)$ (given the patches in $X$ that are nonempty), that is $ P^Y_{\Box}(\pi_{Y,X}^{-1}(\Box^X) \bigm| S_{\pi_{Y,X}(\Box^Y)}>0) =  P^X_{\Box}(\Box^X \bigm| S_{\Box^X}>0)$.
\end{proposition}

  W.l.o.g, we assume that the two arrays, $X$ and $Y$, have the same shape apart from the $d'$th dimension where $Y$ has one additional column (the general case follows by induction). For dimensions $d \neq d'$, it is obvious that the law of patches is consistent under projection because the projection is the identity. Given the same budget $\tau$, Proposition~\ref{expecteddistribution} holds if we can prove the following equality
\begin{eqnarray} \label{eq_position}
  P^Y_{u}\left(\pi_{Y,X}^{-1}(u^{(d')}_{X}) \bigm| |\pi_{Y,X}(u^{(d')}_{Y})| \geq 1\right)
  =  P^X_{u}(u^{(d')}_{X} \bigm| |u^{(d')}_{X}|\geq 1)
\end{eqnarray}
  where $u^{(d')}_{X}$ indicates the initial position, $s^{(d')}_{X}$, and the side-length, $l^{(d')}_{X}$, of the $d'$th side of $\Box^X$; $|u_X^{(d')}|\geq1$ means that there is at least one ``1'' entry in $u_X^{(d')}$.

  Eq.~(\ref{eq_position}) involves four cases in total: $X$ and $Y$ 1) share the initial boundary or 2) share the terminal boundary in the $d'$th dimension; in each case, there are two sub-cases regarding whether the terminal position (for Case 1) or the initial position (for Case 2) of the $d'$th side of $\Box^X$ locates at the boundary of $X$. We can prove that Eq.~(\ref{eq_position}) holds in all cases. Consider all $D$ dimensions we have $ P^Y_{\Box}(\pi_{Y,X}^{-1}(\Box^X) \bigm| S_{\pi_{Y,X}(\Box^Y)}>0) =  P^X_{\Box}(\Box^X \bigm| S_{\Box^X}>0)$ (see Appendix for the complete proof).

\begin{figure*}[ht]
\centering
\includegraphics[width = 1\textwidth]{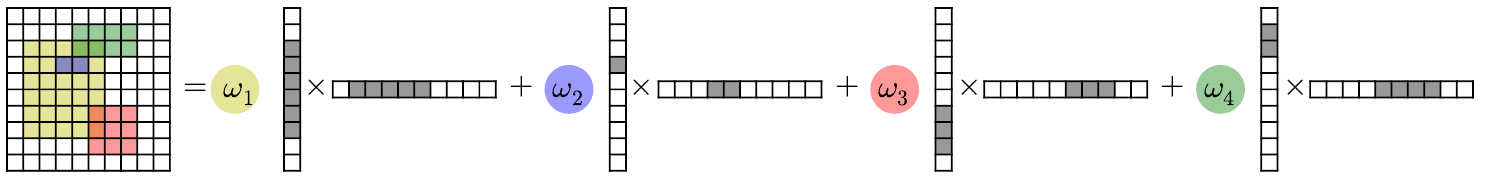}
\caption{SPP relational model: $\{\omega_k\}_k$ denote intensities of relations within communities $\{\Box_k\}_k$.}
\label{SPPDecomposition}
\end{figure*}

\subsection{Self-Consistency of SPP}

  Now we are ready to prove $P^Y_\boxplus(\pi_{Y,X}^{-1}(\boxplus_X)) = P^X_\boxplus(\boxplus_X)$.
\begin{eqnarray}
  & & P^Y_\boxplus(\pi_{Y,X}^{-1}(\boxplus_X)) \nonumber \\
  & = &
  P^Y_{K_{\tau}, \{m_k\}_k}\left(\pi_{Y,X}^{-1}\left(K_{\tau}^{X}, \{m_{k}^{X}\}_{k=1}^{K_{\tau}^{X}}\right)\right) \cdot\prod_{k=1}^{K_{\tau}^{X}}P^Y_{\Box}\left(\pi_{Y,X}^{-1}\left(\Box_{k}^{X}\right) \bigm| S_{\pi_{Y,X}(\Box^Y)}>0 \right) \label{eq:SPPcon1} \nonumber \\
  & \overset{(*)}{=} & P^X_{K_{\tau}, \{m_k\}_k}\left(K_{\tau}^{X}, \{m_{k}^{X}\}_{k=1}^{K_{\tau}^{X}}\right) \cdot\prod_{k=1}^{K_{\tau}^{X}}P^X_{\Box}\left(\Box_{k}^{X} \bigm| S_{\Box^X}>0 \right) \label{eq:SPPcon2} \nonumber\\
  & = & P^X_\boxplus(\boxplus_X) \nonumber
\end{eqnarray}
  where $P^X_{K_{\tau}, \{m_k\}_k}(\cdot)$ and $\prod_{k=1}^{K_{\tau}^{X}}P^X_{\Box}(\cdot)$ in the step marked by $(*)$ are obtained by applying Propositions~\ref{expectednumber} and~\ref{expecteddistribution}, respectively. According to the Kolmogorov extension theorem (see Section~\ref{sub:kolmogorov}), we have the following result.
\begin{theorem}
  The probability measure $P_{\boxplus}^X$ on a measurable space $(\Omega_{X}, \mathcal{B}_{X})$ of SPP, $X\in\mathcal{F}(\mathbb{N}^D)$, can be uniquely extended to $P_{\boxplus}^{\mathbb{N}^D}$ on $(\Omega_{\mathbb{N}^D}, \mathcal{B}_{\mathbb{N}^D})$ as the projective limit measurable space.
\end{theorem}

%==========================================================
\section{Application to Relational Modeling} \label{sec:rm}
%==========================================================

\subsection{SPP Relational Model}

  A typical application of SPP is relational modeling. Given the relational data as an asymmetric matrix $R \in \{0,1\}^{N \times N}$, with $R_{ij}$ indicating the relation from node $i$ to node $j$, the patches $\{\Box_k\}_k$ with different rates $\{\omega_k\}_k$ of a partition $\boxplus$ are used for modeling communities with different intensities of relations. Because $\{\Box_k\}_k$ can be overlapped, the intensity of relations in an overlapped part on $R$ is synthesized by the rates of the involved patches (see Figure~\ref{SPPDecomposition}).

  The generative process of an SPP relational model is as follows: (1) Generate a partition $\boxplus$; (2) For $i = 1,\ldots, N$, generate row index $r_i$ of $R$; (3) for $j = 1,\ldots, N$, generate column index $c_j$ of $R$; (4) for $i,j = 1,\ldots, N$, generate relational data $R_{r_i c_j} \sim \mbox{Bernoulli}(\sigma(\sum_{k=1}^{K_{\tau}} \frac{\omega_k}{\gamma}\cdot u_{k,i}^{(1)} u_{k,j}^{(2)}))$, where $\sigma(x)=\frac{\exp(x+e^{-6})-1}{\exp(x+e^{-6})+1}$ is a selected function for mapping the aggregated rate from $[0,\infty)$ to $(0, 1)$ as intensity of relations and $\gamma$ is a scaling parameter. While here we instantiate an SPP relational model with binary interactions (Bernoulli likelihood), other types of relations (e.g., Categorical likelihood) can also be plugged in.

  Actually, SPP and the mapping function $\sigma(\cdot)$ play together as the role of random function $W(\cdot)$ defined in Section~\ref{sub:graphon}. The uniformly exchanged row and column indices ($r_i$ and $c_j$) resemble the row and column indices ($\xi_i^{row}$ and $\eta_j^{col}$) which are uniformly sampled in $[0, 1]$. By re-arranging the rows and columns of $R$ according to the inferred indices, the SPP relational model is expected to uncover homogeneous interactions in $R$ as compact patches.

\subsection{Sampling for SPP Relational Model}

\begin{algorithm}[t]
\caption{Sampling for SPP Relational Model}
\label{detailInference}
\begin{algorithmic}
  \REQUIRE Relational data $R$, budget $\tau$, hyper-parameters $\theta, \gamma$, iteration time $T$
  \ENSURE $K_{\tau}$, $\{m_{k},u^{(1)}_k, u^{(2)}_k\}_k$, $\{r_i\}_i$ and $\{c_j\}_j$
  \FOR{$t=1,\cdots, T$}
    \STATE Sample $K_\tau$;
    \STATE Sample $\{m_k\}_{k=1}^{K_{\tau}}$; // Metropolis-Hastings
    \FOR{$k=1,\cdots,K_{\tau}$}
      \STATE Sample $(u_k^{(1)}, u_k^{(2)})$; // C-SMC (Algorithm \ref{patch_pg})
    \ENDFOR
      \STATE Sample $\{r_i\}_{i=1}^N,\{c_j\}_{j=1}^N$; // Multiple-Try Metropolis
  \ENDFOR
\end{algorithmic}
\end{algorithm}

  The joint probability of the data $\{R_{ij}\}_{i,j}$, the number of nonempty patches $K_{\tau}$, the variables of the nonempty patches $\{m_k, u^{(1)}_k, u^{(2)}_k\}_{k=1}^{K_\tau}$, and the indices $\{r_i\}_i, \{c_j\}_j$ gives
{\small\begin{eqnarray}
   P(\{R_{ij}\}_{i,j}, K_{\tau}, \{m_{k},u^{(1)}_k, u^{(2)}_k\}_k, \{r_i\}_i, \{c_j\}_j|\theta,\tau,\gamma,N) \nonumber \\
  = \prod_{i,j} P(R_{r_ic_j}|K_{\tau},\{m_k, u^{(1)}_k, u^{(2)}_k\}_k,\gamma) \nonumber \\
  \cdot P(\{r_i\}_i|N) \cdot P(\{c_j\}_j|N) \cdot P(K_{\tau},\{m_k\}_{k}|\tau,\theta,N) \nonumber \\
   \cdot \prod_k P(u^{(1)}_{k}|\theta,N) P(u^{(2)}_{k}|\theta,N)  \nonumber
\end{eqnarray}}
  where $P(R_{r_ic_j}|K_{\tau},\{m_k, u^{(1)}_k, u^{(2)}_k\}_k,\gamma) = \rho_{ij}^{R_{r_ic_j}}(1-\rho_{ij})^{1-R_{r_ic_j}}=\ell(r_i,c_j,\rho_{ij}) $ refers to the probability of $R_{r_ic_j}$ and $\rho_{ij} = \sigma(\sum_{k=1}^{K_{\tau}} \frac{\omega_k}{\gamma}\cdot u_{k,i}^{(1)} u_{k,j}^{(2)})$ denotes the parameter of the Bernoulli likelihood; $P(\{r_i\}_i|N)=P(\{c_j\}_j|N)=\frac{1}{N!}$ denotes the probability of row and column indices; $P(K_{\tau},\{m_k\}_{k}|\tau,\theta,N) = (\gamma N^2\theta_*)^{K_{\tau}}e^{-\tau\gamma N^2\theta_*}$ (where $\theta_*=\frac{1}{N^2} \cdot \left[\theta+N(1-\theta)\right]^2$) denotes the joint probability of the number and the time points of the nonempty patches.

%Let $\ell(r_i,c_j,\rho_{ij}) = P(R_{r_ic_j}|K_{\tau},\{m_k, u^{(1)}_k, u^{(2)}_k\}_k,\gamma) = \rho_{ij}^{R_{r_ic_j}}(1-\rho_{ij})^{1-R_{r_ic_j}}$ be the likelihood function, where $\rho_{ij} = \sigma(\sum_{k=1}^{K_{\tau}} \frac{\omega_k}{\gamma}\cdot u_{k,i}^{(1)} u_{k,j}^{(2)})$ denotes the parameter of the Bernoulli likelihood; $P(\{r_i\}_i|N)=P(\{c_j\}_j|N)=\frac{1}{N!}$ denotes the probability of indexing orders; $P(K_{\tau},\{m_k\}_{k}|\tau,\theta_0,\theta,N) = (\gamma N^2\theta_*)^{K_{\tau}}e^{-\tau\gamma N^2\theta_*}$ denotes the joint probability of $K_{\tau}$ and $\{m_k\}_{k=1}^{K_{\tau}}$ (where $\theta_*=\frac{\theta_0^2}{N^2} \cdot \left[\theta+N(1-\theta)\right]^2$).

  We adopt MCMC methods for sampling the posteriors of $K_{\tau}$, $\{m_{k},u^{(1)}_k, u^{(2)}_k\}_k$, $\{r_i\}_i$ and $\{c_j\}_j$. By an abuse of notation, in the following $\rho^{x \rightarrow x^*}_{ij}$ is used to represent the case that the likelihood is updated by replacing $x$ with $x^*$, keeping the other variables unchanged. Also, we use $\rho_{ij}^{\neg k}$ to denote the likelihood computed excluding the $k$th patch. The sampling algorithm is outlined in Algorithm~\ref{detailInference}.
\begin{algorithm}[t]
\caption{C-SMC Sampler for $\Box^*_k$}
\label{patch_pg}
\begin{algorithmic}
  \REQUIRE Relational data $R$, hyper-parameters $\theta$, current patches $\{\Box_k\}_k$, number of particles $C$, maximum length of the sequence in C-SMC $I$
  \ENSURE New position of patch $\Box^*_k = (u_k^{(1)}, v_k^{(2)})$
    \FOR{$c=2,\cdots,C$}
      \STATE Sample random initial positions for $\mathcal{P}_c(0)$;
    \ENDFOR
      \FOR{$i=1,\cdots, I$}
    \STATE Set $\mathcal{P}_1(i)=\Box_k(i)$ and $j_1=1$;
    \FOR{$c=2,\cdots,C$}
      \STATE Sample $\mathcal{P}_c(i)$ from $q(\cdot|\mathcal{P}_c(i-1))$;
    \ENDFOR
    \FOR{$c=1,\cdots,C$}
      \STATE Update weight $\omega_c(i)$ according to Eq.~(\ref{pg_weight_update});
    \ENDFOR
    \STATE Normalize $\{\omega_c(i)\}_{c=1}^C$ and obtain $\{\bar{\omega}_c(i)\}_{c=1}^C$;
     \STATE Resample indices $\{j_c\}_{c=2}^C$ from $\sum_{c'=1}^C\bar{\omega}_{c'}(i)\delta_{c'}$;
     \STATE $\forall c$, Assign $\mathcal{P}_{c}(i)=\mathcal{P}_{j_c}(i)$;
  \ENDFOR
  \STATE Sample $c^*$ from $\sum_{c'=1}^C\bar{\omega}_{c'}(I)\delta_{c'}$ and let $\Box^*_k = \mathcal{P}_{c^*}(I)$;
\end{algorithmic}
\end{algorithm}

\paragraph{Sample $K_{\tau}$} \label{eq_SamplingNtau}
  We use a similar strategy of~\cite{adams2009tractable} for updating $K_{\tau}$. First, we use probability $P_0 = \frac{1}{2}$ (or $1-P_0$) to choose proposing adding (or removing) a nonempty patch. The proposal probability of adding a nonempty patch is
\begin{eqnarray}
  q_{\textrm{add}}(K_{\tau} \rightarrow K_{\tau}+1) = P_0\cdot\frac{1}{\tau}\cdot P(\Box_*)
\end{eqnarray}
  where $\Box_*$ denotes a newly added patch; the proposal probability of deleting an existing patch is
\begin{eqnarray}
q_{\textrm{del}}(K_{\tau} \rightarrow K_{\tau}-1) = \frac{1-P_0}{K_{\tau}}\cdot
\end{eqnarray}
  We accept adding or removing a patch with a ratio of $\min(1,\alpha_{add})$ or $\min(1,\alpha_{del})$, where
\begin{eqnarray}
  \alpha_{\textrm{add}}  = \frac{\prod_{i,j}\ell(r_i,c_j,\rho_{ij}^{ K_{\tau} \rightarrow K_{\tau}+1})}{\prod_{i,j}\ell(r_i,c_j,\rho_{ij})}\cdot\frac{\tau\gamma N^2\theta_*}{K_{\tau}+1}\cdot\frac{1-P_0}{P_0}
\end{eqnarray}
\begin{eqnarray}
\alpha_{\textrm{del}}  = \frac{\prod_{i,j}\ell(r_i,c_j,\rho_{ij}^{ K_{\tau} \rightarrow K_{\tau}-1})}{\prod_{i,j}\ell(r_i,c_j,\rho_{ij})}\cdot\frac{K_{\tau}}{\tau\gamma N^2\theta_*}\cdot\frac{P_0}{1-P_0}
\end{eqnarray}
  It is worth noting that in the proposal of adding a new nonempty patch, it is generated by following Step 2 (a) and (b) of ``Alternative Construction of SPP'' in Appendix.

\paragraph{Sample $\{m_k\}_k$} \label{eq_SamplingPiK}
  For the $k$th patch, $k\in\{1,\cdots,K_{\tau}\}$, a new $m_k^*$ is sampled from the proposal distribution, which is a truncated Exponential distribution $f(m_k^*)\propto e^{-\gamma N^2\theta_*m_k^*}\mathbbm{1}[m_k^*\in(0,\tau-\sum_{k'\ne k}m_{k'})]$. We then accept $m_k^*$ with a ratio of $\min(1, \alpha)$, where
\begin{equation}
  \alpha=\frac{\prod_{i,j}\ell(r_i,c_j,\rho_{ij}^{ m_k \rightarrow m_k^*})}{\prod_{i,j}\ell(r_i,c_j,\rho_{ij})}\cdot \frac{e^{-\gamma N^2\theta_* m_k}}{e^{-\gamma N^2\theta_* m_k^*}}
\end{equation}

\paragraph{Sample $\{u_k^{(1)}, u_k^{(2)}\}_k$} \label{eq_SamplingUK}

A straightforward way to update $\{u_k^{(1)}, u_k^{(2)}\}_k$ is to use the Metropolis-Hastings (MH) algorithm. Since MH only proposes local changes to update the current positions of $\{\Box_k\}_k$ in each iteration, the sampler can explore limited space in $\Omega_X$ given a reasonable sampling budget. To overcome this problem, we adopt Conditional-Sequential Monte Carlo~\cite{andrieu2010particle} (C-SMC) for sampling completely new positions of $\{\Box_k\}_k$ in each iteration.

The description of the C-SMC sampler for $\Box_k$ is shown in Algorithm~\ref{patch_pg}. C-SMC works similarly as SMC to propose high dimensional variables ($u_k^{(1)}, u_k^{(2)}$ in our case) at each stage of the sequence, except for clamping the first particle as the current patch $\mathcal{P}_1(i)=\Box_k(i)$ (where $\Box_k(i)$ represents the patch updated at the $i$th stage). The advantage of C-SMC for our problem is that it is able to propose a completely new candidate patch to replace the current one with high acceptance ratio. The key connection between C-SMC and patch sampling is to view the generative process of a patch (introduced in Section~\ref{sec:spp}) as a sequence of state variables (similar as~\cite{LakOryTeh2015ParticleGibbs} using C-SMC for sampling a tree). Starting from a randomly chosen initial position, the patch undergoes entry-wise growing in both directions of row and column. Let $q(\cdot|\mathcal{P}_c(i-1))$ denote the proposal distribution of the $c$th particle at the $i$th stage, $q(\cdot|\mathcal{P}_c(i-1))$ is set as the conditional distribution of $\mathcal{P}_c(i)$ given $\mathcal{P}_c(i-1)$ under the generative prior, in both directions of row and column. As a result, the weight $\omega_c(i)$ for the $c$th particle at the $i$th stage is updated as
\begin{eqnarray} \label{pg_weight_update}
\omega_c(i) = \frac{\prod_{i',j'}\ell(r_{i'},c_{j'},\rho_{i'j'}^{\mathcal{P}_c(i-1)\rightarrow \mathcal{P}_c(i)})}{\prod_{i',j'}\ell(r_{i'},c_{j'},\rho_{i'j'})}
\end{eqnarray}
where $i'$ and $j'$ refer to the row and column indices whose likelihoods are influenced by the particle updating.

Although the complexity of the C-SMC sampler (Algorithm \ref{patch_pg}) is higher than the MH algorithm, C-SMC can propose completely new patches with high acceptance ratio. This ability enables it to fast explore the partition space $\Omega_X$. In this sense, C-SMC can provide a much better approximation to the posterior distribution of the patches.

\paragraph{Sample $\{r_i\}_i, \{c_j\}_{j}$} \label{eq_Samplingruv}
To cooperate with the C-SMC sampler for higher acceptance ratio, we adopt the Multiple-Try Metropolis method~\cite{liu2000multiple} for sampling the row and column indices of the relational data (please refer to Appendix for details).

\subsection{Experiments} \label{sec:exp}

\begin{figure*}[ht]
\centering
  \includegraphics[width = 0.95 \textwidth, bb = 31 291 884 408, clip]{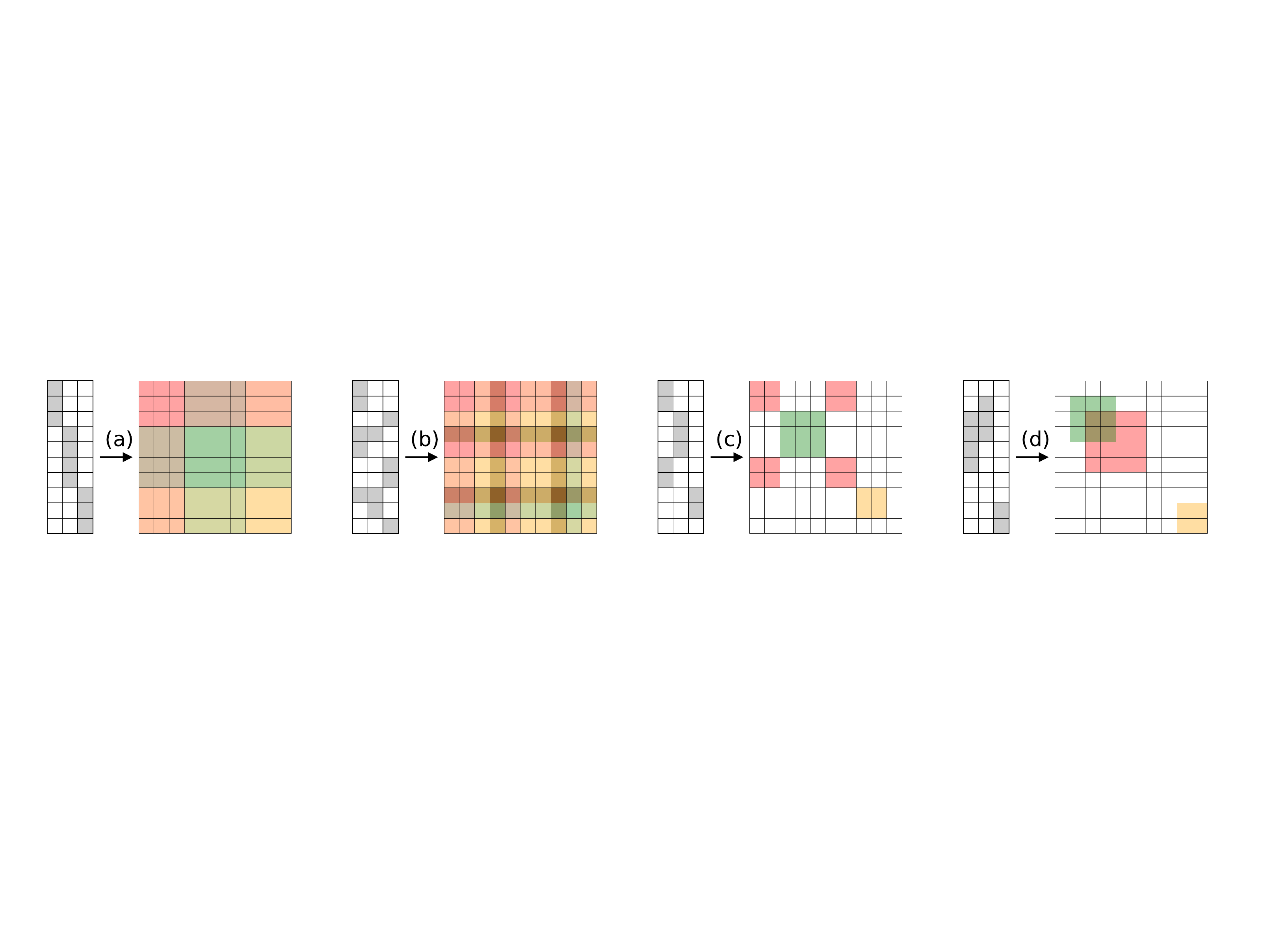}
  \caption{Partition illustration with three toy binary latent feature vectors for each method: (a) IRM presents regular grids; (b) LFRM presents plaid patterns; (c) MTA-RM presents non-overlapped noncontiguous tiles; (d) SPP-RM presents overlapped contiguous patches.}
\label{fig:models}
\end{figure*}

\begin{table*}[th]
\centering \small
\caption{Relational modeling (link prediction) comparison results (AUC$\pm$std)}
\begin{tabular}{l|ccccc}
  \hline
  Data Sets  &{IRM} & {LFRM} & {MP-RM} & {MTA-RM} & {SPP-RM} \\
  \hline
  Digg     & $0.792 \pm 0.011 $  & $0.801 \pm 0.031 $  & $0.784 \pm 0.020 $  & $0.793 \pm 0.005 $  & $\textbf{0.815} \pm 0.011 $ \\
  Flickr   & $0.870 \pm 0.003 $  & $0.881 \pm 0.006 $   & $0.868 \pm 0.011 $ & $0.872 \pm 0.004 $  & $\textbf{0.885} \pm 0.008 $ \\
  Gplus     & $0.857 \pm 0.002 $  & $0.860 \pm 0.008 $   & $0.855 \pm 0.007 $ & $0.857 \pm 0.002 $  & $\textbf{0.868} \pm 0.002 $ \\
  Facebook  & $0.872 \pm 0.013 $  & $0.881 \pm 0.023 $  & $0.876 \pm 0.028 $  & $0.885 \pm 0.010 $  & $\textbf{0.889} \pm 0.018 $ \\
  Twitter  & $0.860 \pm 0.003 $  & $0.868 \pm 0.021 $  & $0.815 \pm 0.055 $ & $0.870 \pm 0.006 $   & $\textbf{0.870} \pm 0.016 $ \\
  \hline
\end{tabular}
\label{rm_dataset}
\end{table*}

  We empirically test the SPP relational model (SPP-RM) for link prediction. We compare SPP-RM with four state-of-the-arts: (1) IRM~\cite{kemp2006learning} (regular grids); (2) LFRM~\cite{miller2009nonparametric} (plaid grids); (3) MP Relational Model (MP-RM)~\cite{roy2009mondrian} (hierarchical $k$d-tree); (4) Matrix Tile Analysis Relational Model (MTA-RM)~\cite{Givoni06} (noncontiguous tiles). All these models except MP-RM can be represented as a (weighted) sum of outer products of binary latent feature (community) vectors (see Figure~\ref{fig:models}). For IRM and LFRM, we adopt the collapsed Gibbs sampling algorithms for inference; for MP-RM, we adopt the reversible-jump MCMC algorithm for inference~\cite{wang2011nonparametric}; for MTA-RM, we adopt the Iterative Conditional Modes algorithm used in~\cite{Givoni06}.

  \textbf{Data Sets}: Five social network data sets are used: Digg, Flickr~\cite{Zafarani+Liu:2009}, Gplus~\cite{facebook_mcauley2012learning}, and Facebook, Twitter~\cite{leskovec2010predicting}. We extract a subset of nodes (top 1000 active nodes based on their interactions with others) from each data set for constructing the relational data matrix.

  \textbf{Experimental Setting}: We set the hyper-parameters for each method as follows: In IRM, we let $\alpha$ be sampled from a gamma prior $\Gamma(1, 1)$ and the row and column partitions be sampled from two independent Dirichlet processes; In LFRM, we let $\alpha$ be sampled from a gamma prior $\Gamma(2, 1)$. As the budget parameter of MP-RM is hard to sample~\cite{balaji2016aistats}, we set it to $3$, which suggests that around $(3+1)\times(3+1)$ blocks would be generated. For parametric model MTA-RM, we simply set the number of tiles to 16; In SPP-RM, we set $\theta=0.99$ and $\tau=0.5,\gamma=10^{-2}$, which leads to an expectation of 12.5 patches. We use $5$ particles in Algorithm~\ref{patch_pg} and set the maximum length of the sequence to $500$ (half number of rows/columns). The reported performance is averaged over 10 randomly selected hold-out test sets ($\text{Train}:\text{Test} = 9:1$).

  \textbf{Results}: Table~\ref{rm_dataset} reports the performance comparison results on the five data sets. We can see that SPP-RM consistently outperforms the other four methods in all cases, with around 0.01 improvement compared to the runner-up in prediction AUC. The overall results validate that SPP-RM is effective in relational modeling due to its flexibility via attaching patches to dense regions.
  %MTA-RM also performs well -- One reason may be that MTA-RM and SPP-RM have similar modeling philosophy (see Figure~\ref{fig:models}); another reason is that, being a parametric model, MTA-RM converges faster and can be fitted better than MP-RM and SPP-RM using same wall-clock time.

\begin{figure*}[p]
\centering
  \includegraphics[width = 0.18 \textwidth, bb = 110 210 497 597, clip]{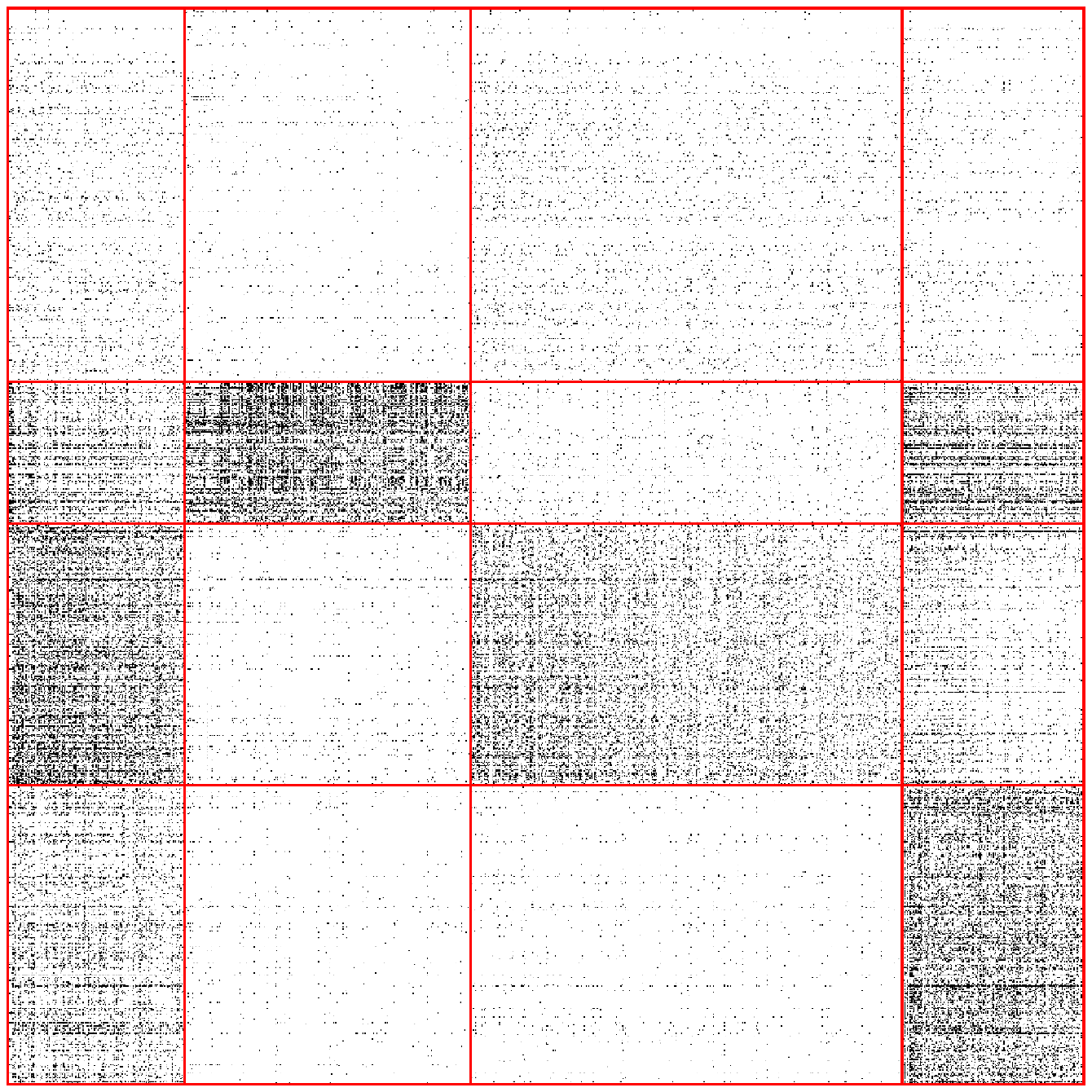}
  \includegraphics[width = 0.18 \textwidth, bb = 110 210 497 597, clip]{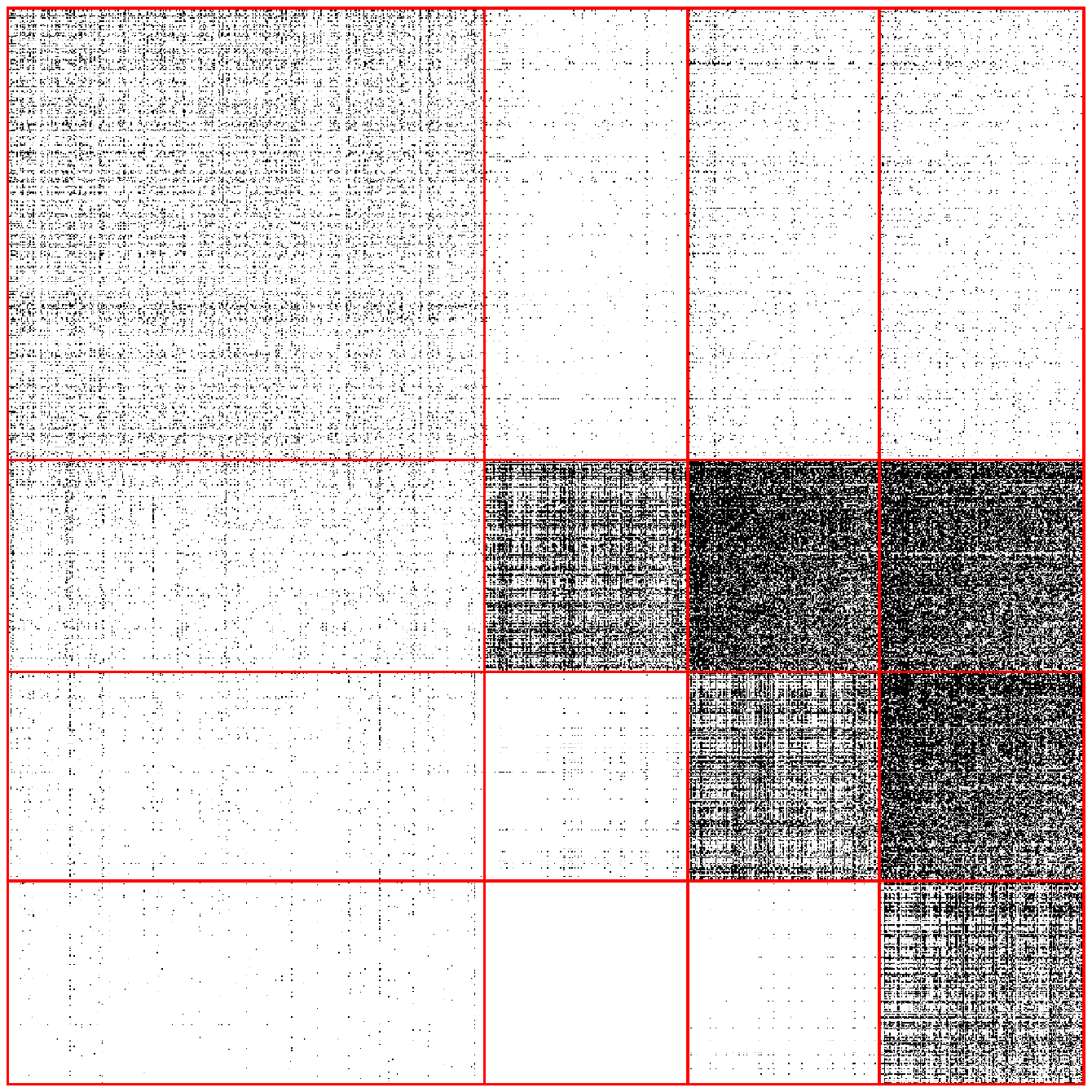}
  \includegraphics[width = 0.18 \textwidth, bb = 110 210 497 597, clip]{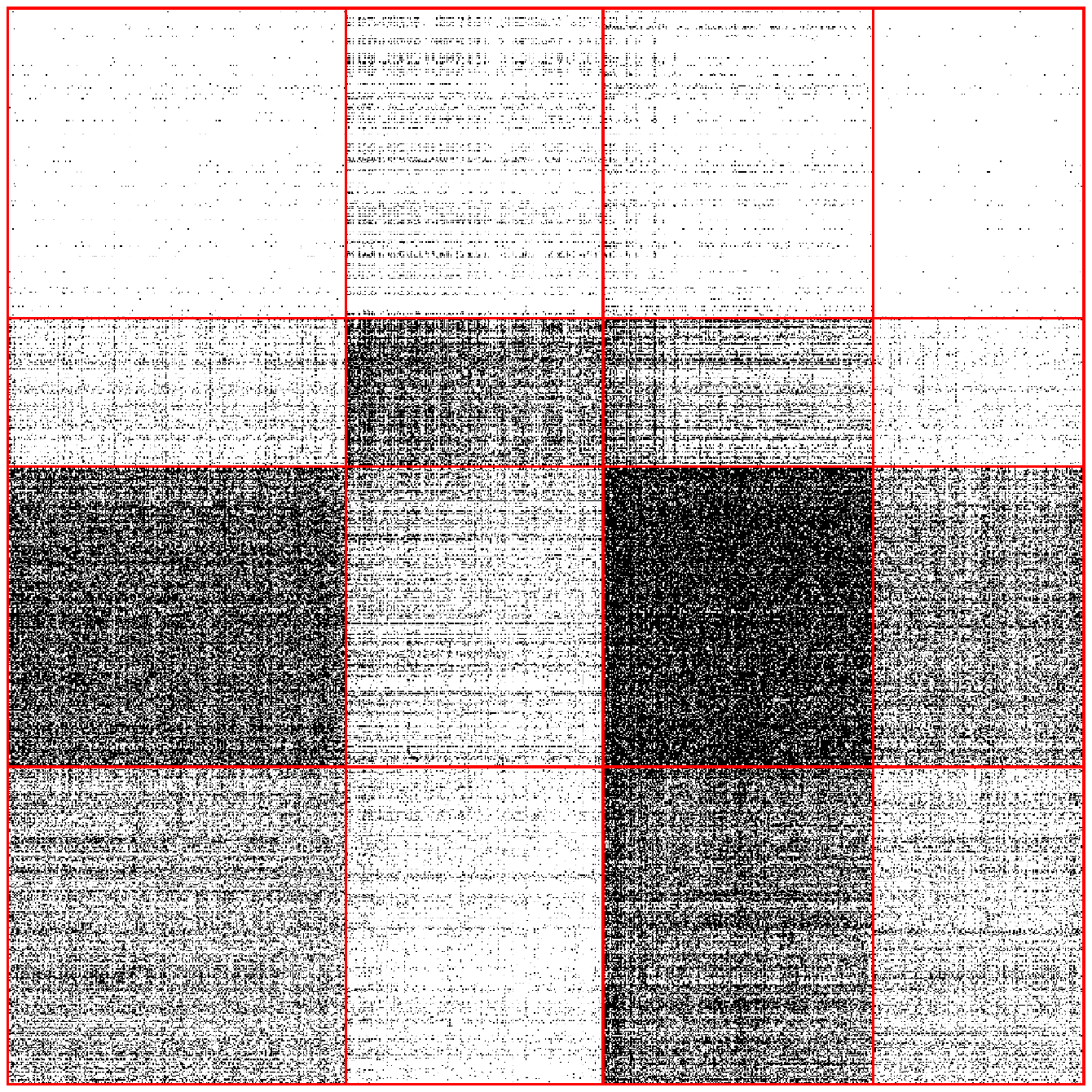}
  \includegraphics[width = 0.18 \textwidth, bb = 110 210 497 597, clip]{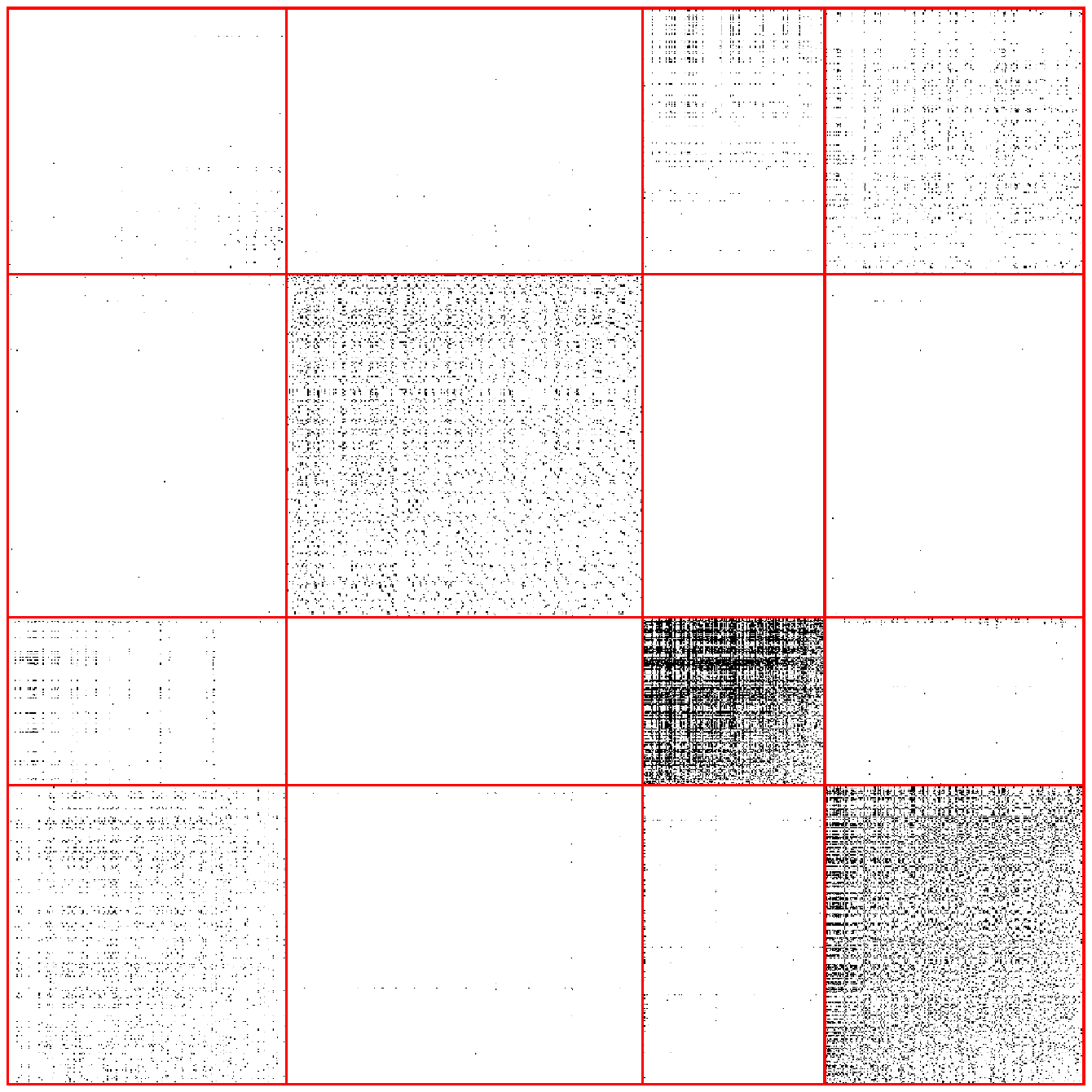}
  \includegraphics[width = 0.18 \textwidth, bb = 110 210 497 597, clip]{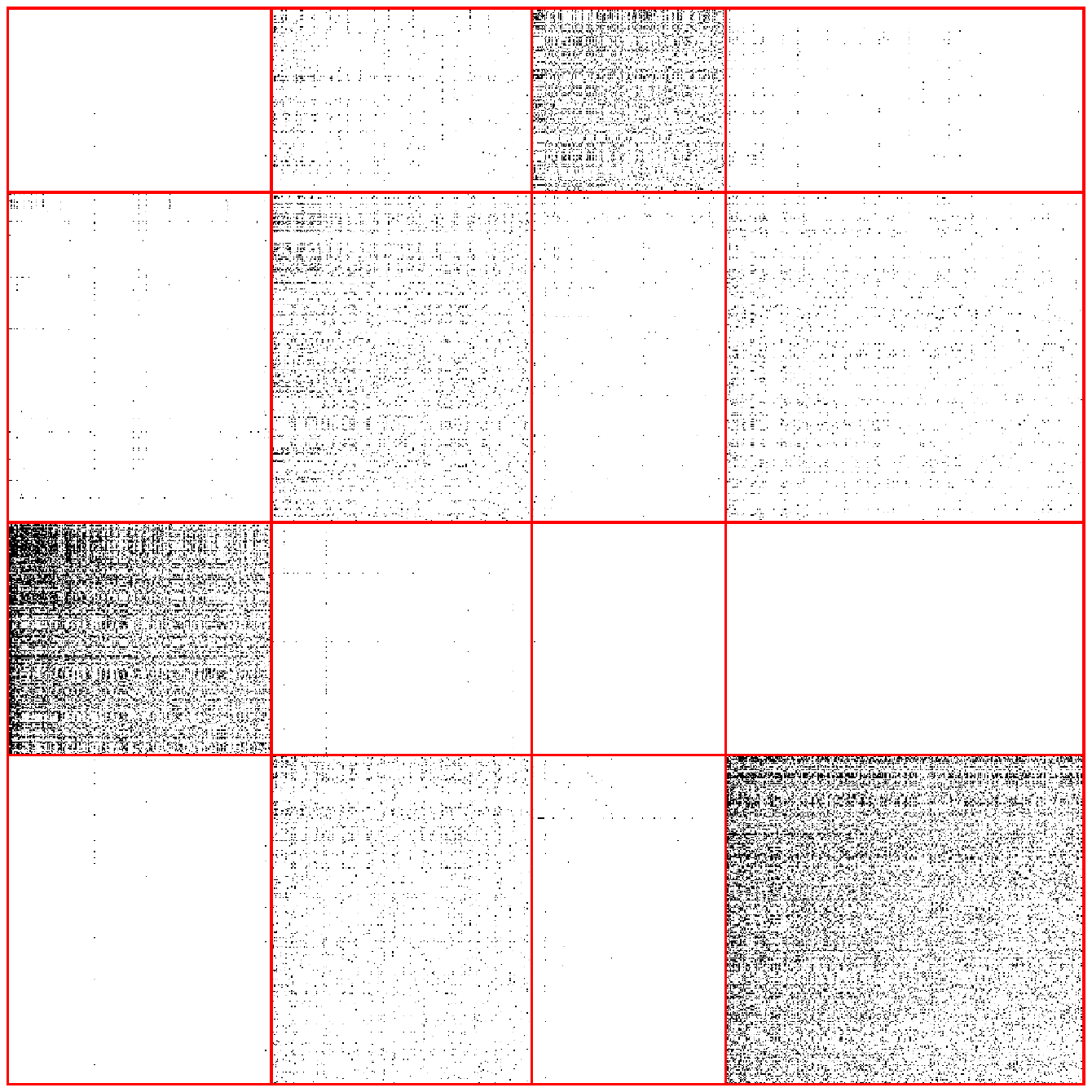}

  \includegraphics[width = 0.18 \textwidth, bb = 110 210 497 597, clip]{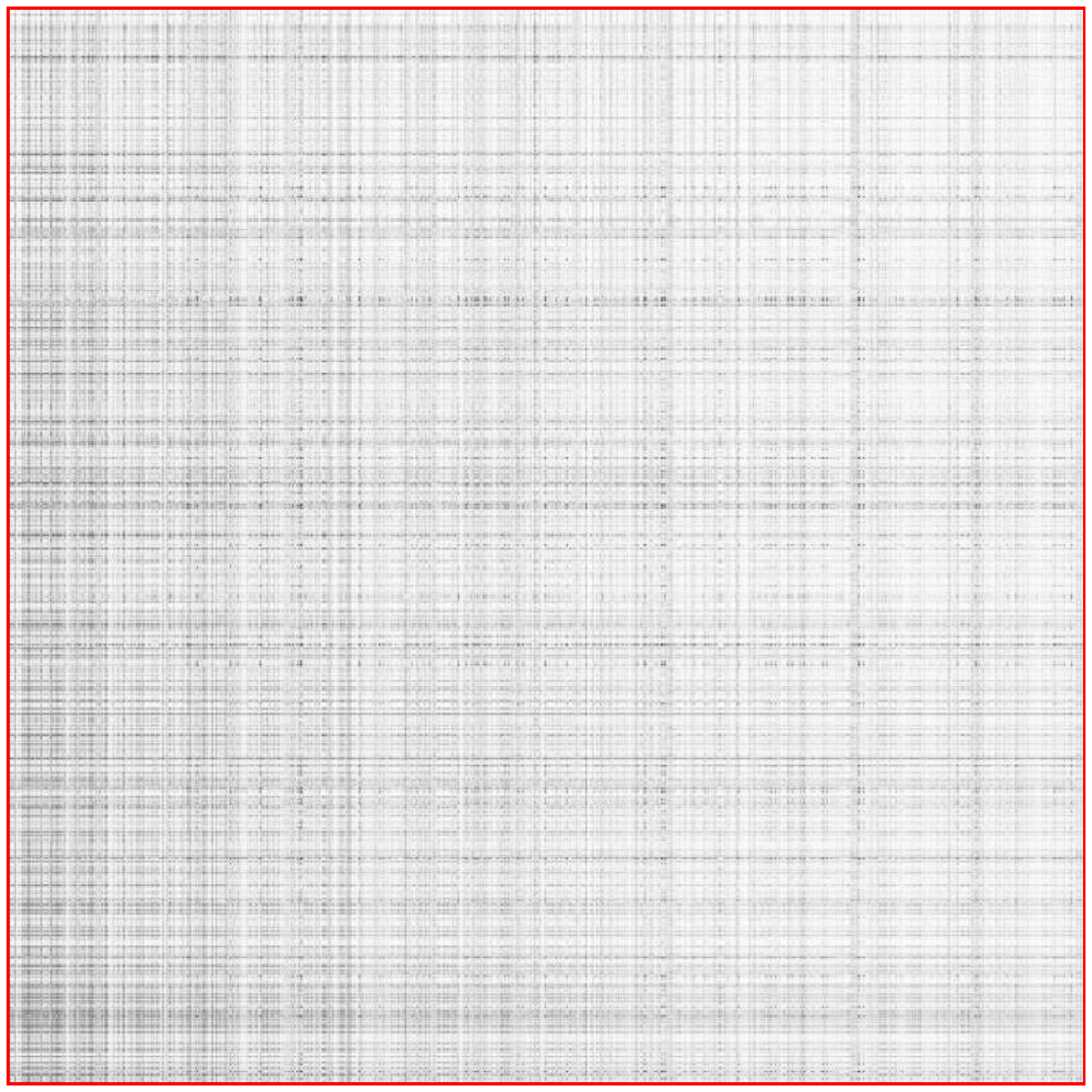}
  \includegraphics[width = 0.18 \textwidth, bb = 110 210 497 597, clip]{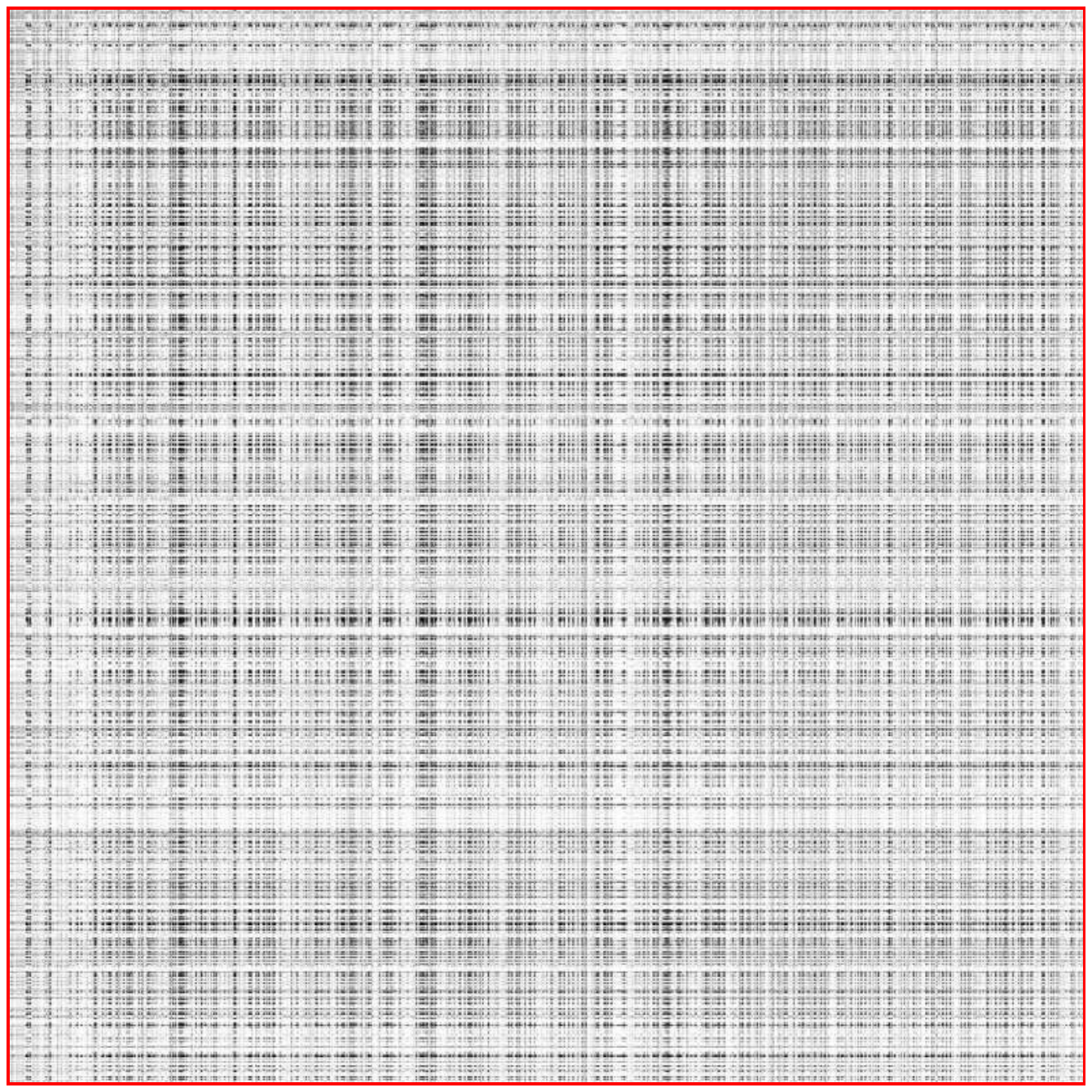}
  \includegraphics[width = 0.18 \textwidth, bb = 110 210 497 597, clip]{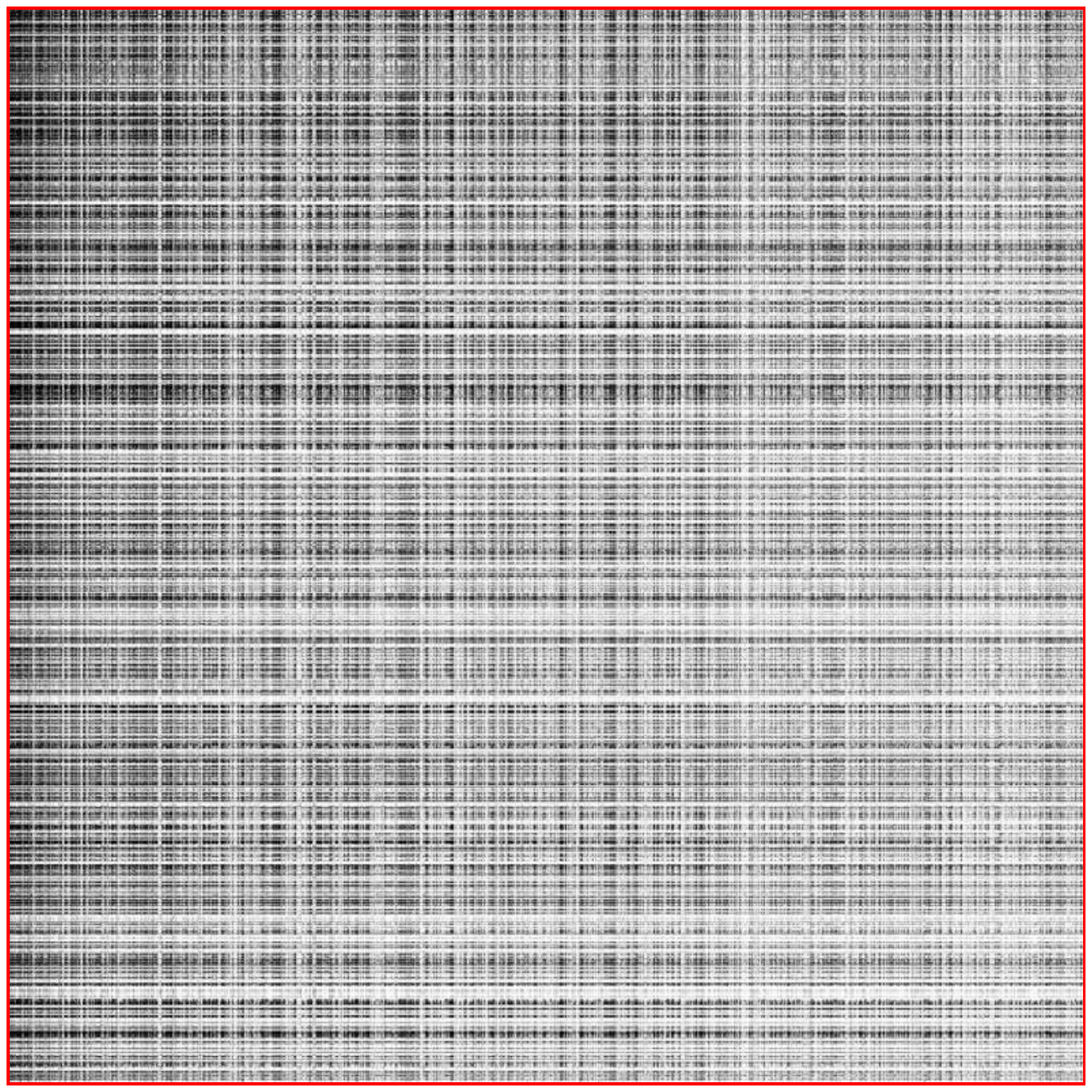}
  \includegraphics[width = 0.18 \textwidth, bb = 110 210 497 597, clip]{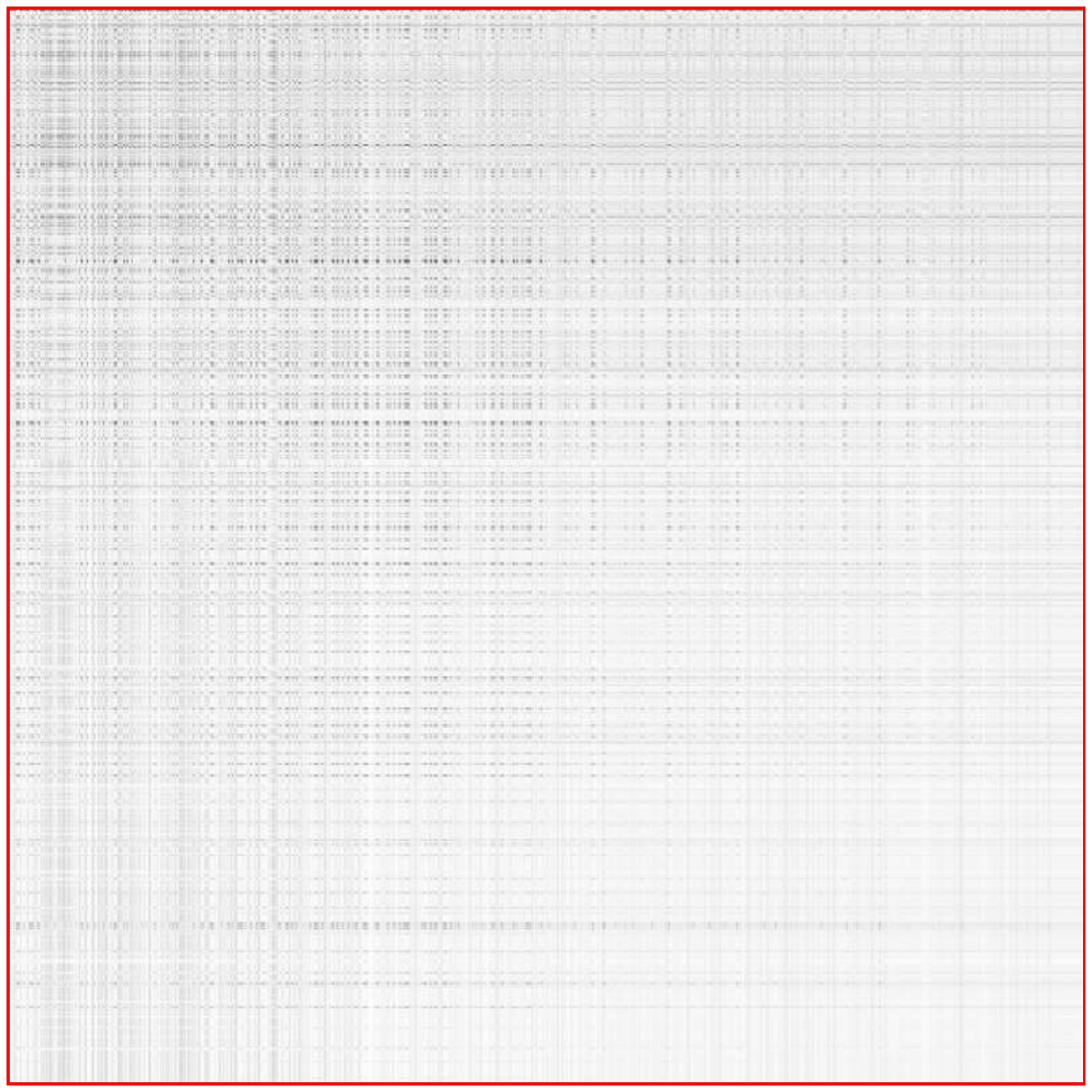}
  \includegraphics[width = 0.18 \textwidth, bb = 110 210 497 597, clip]{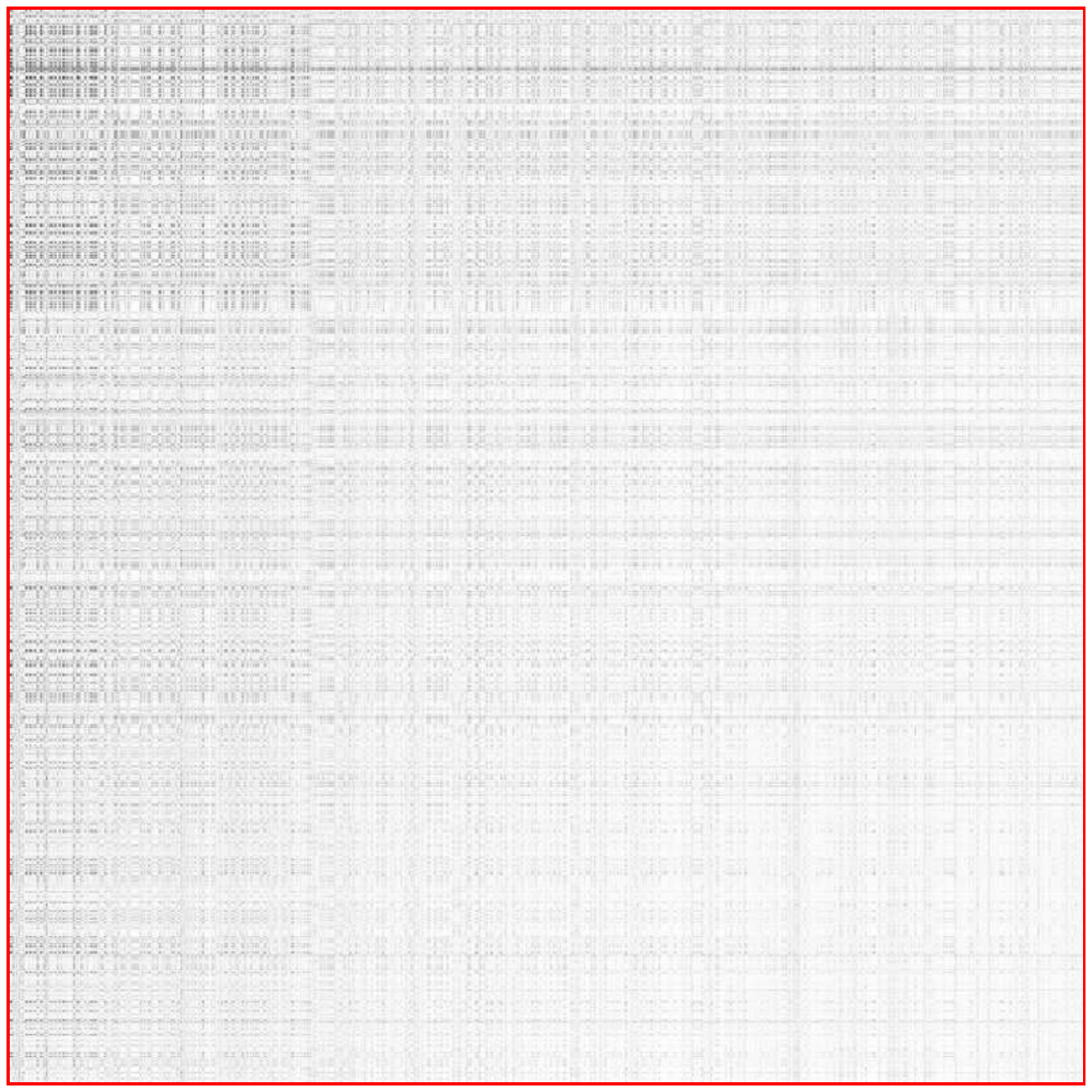}

  \includegraphics[width = 0.18 \textwidth, bb = 110 210 497 597, clip]{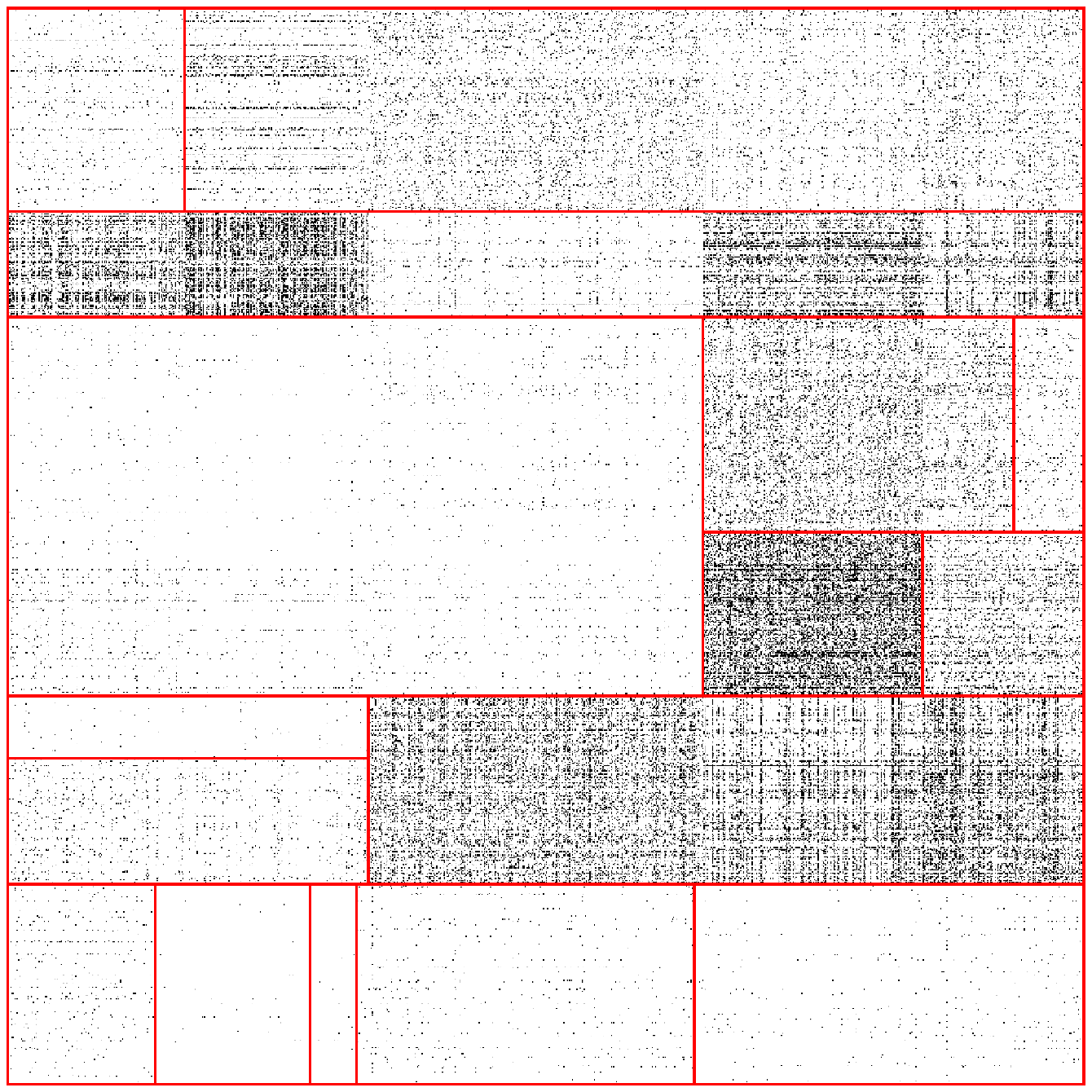}
  \includegraphics[width = 0.18 \textwidth, bb = 110 210 497 597, clip]{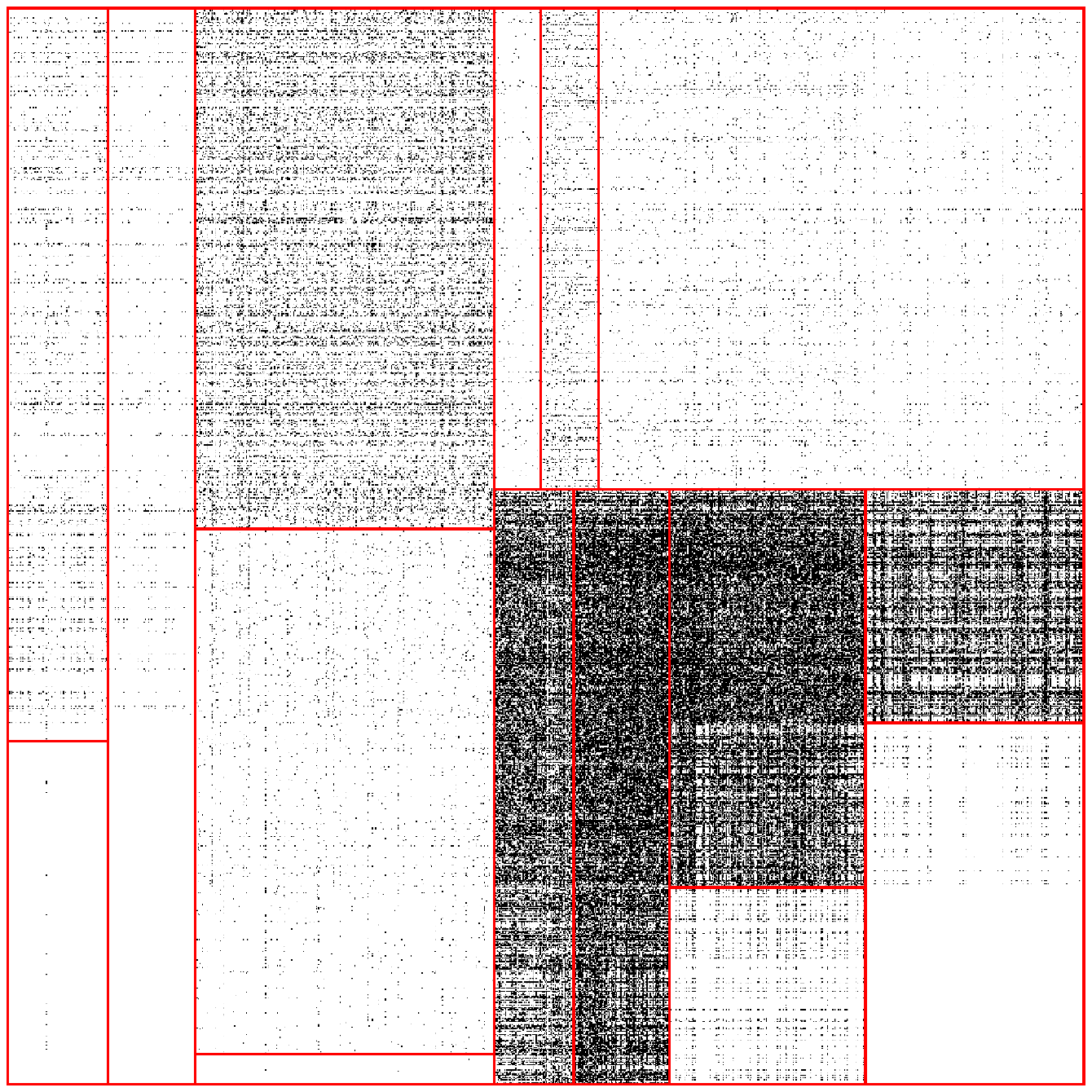}
  \includegraphics[width = 0.18 \textwidth, bb = 110 210 497 597, clip]{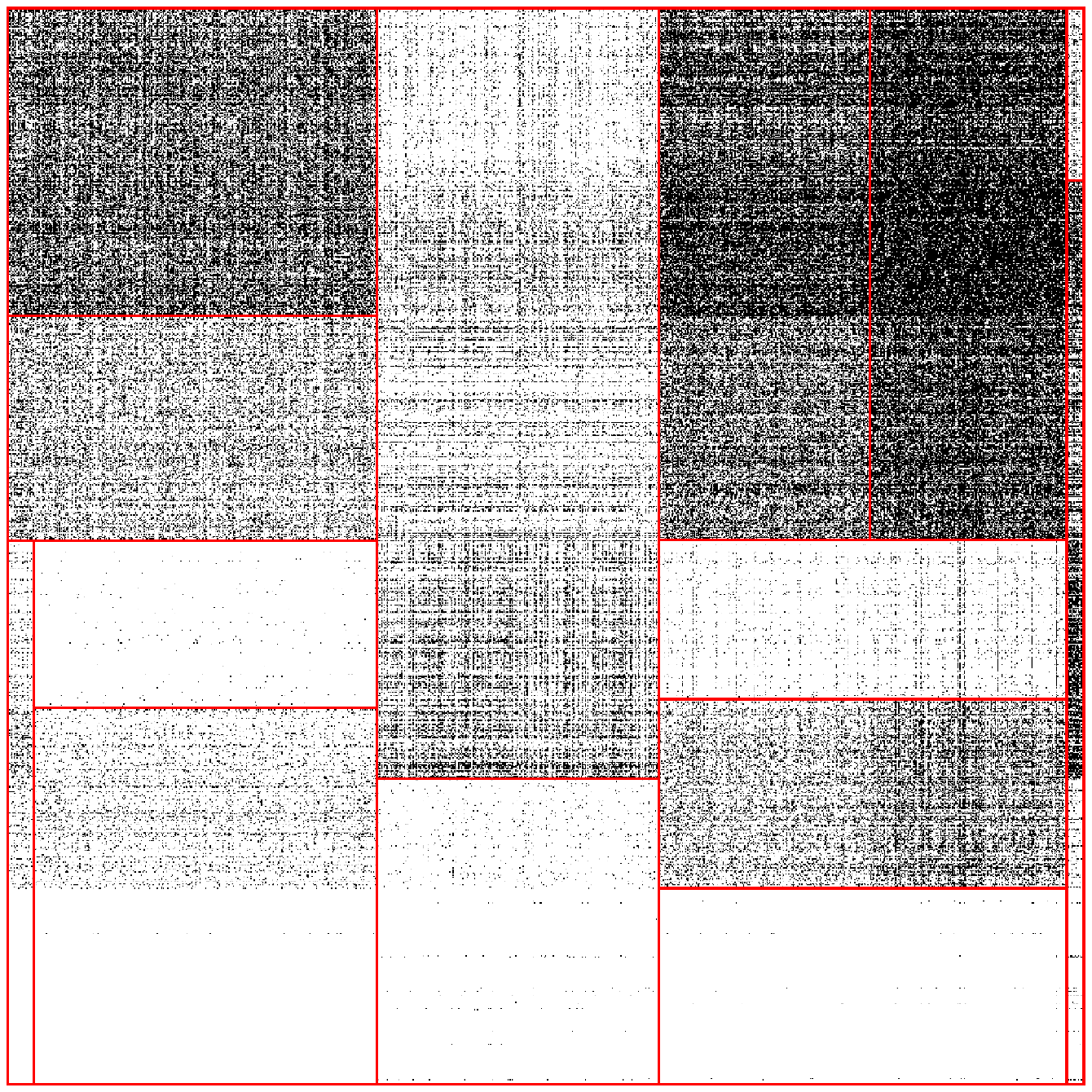}
  \includegraphics[width = 0.18 \textwidth, bb = 110 210 497 597, clip]{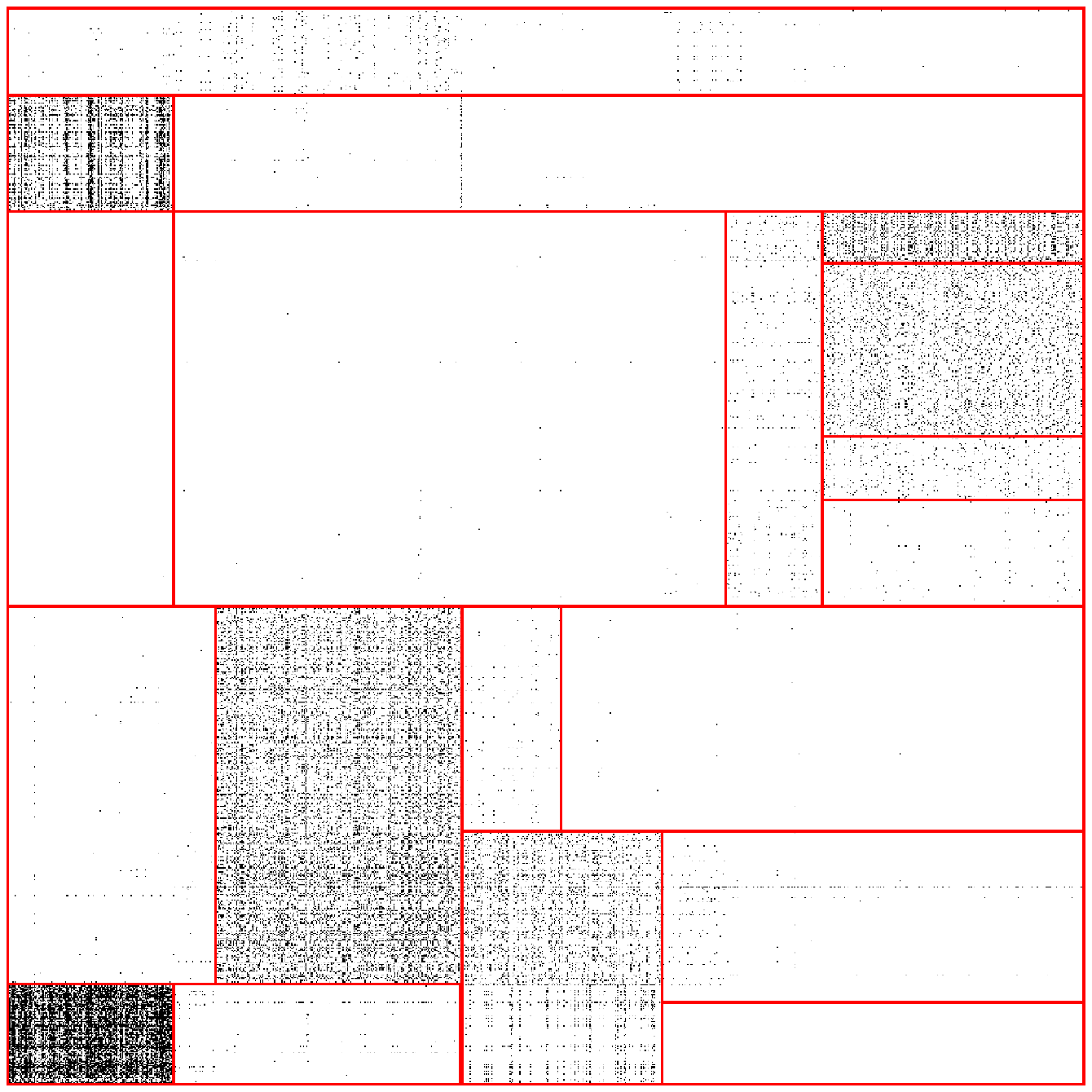}
  \includegraphics[width = 0.18 \textwidth, bb = 110 210 497 597, clip]{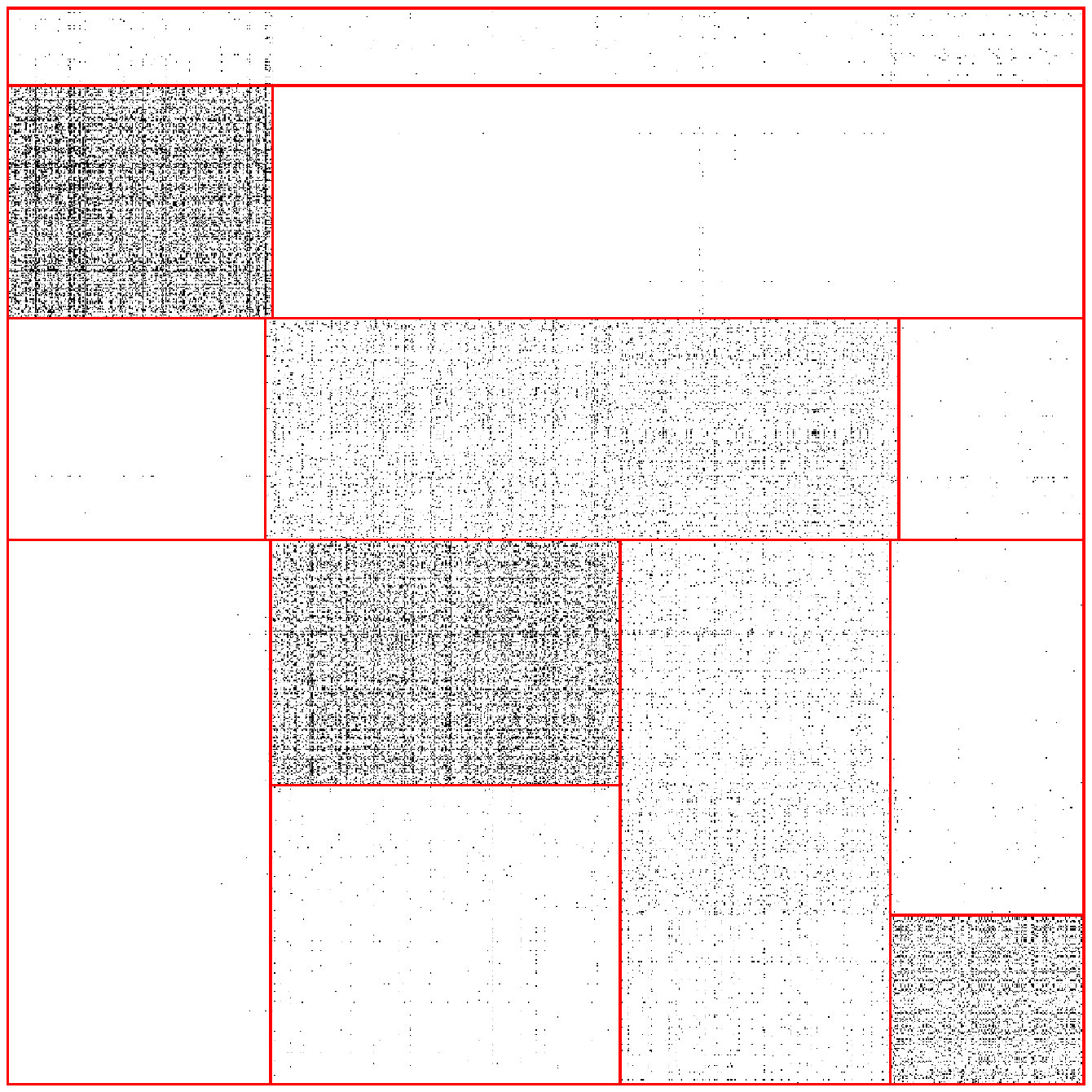}

  \includegraphics[width = 0.18 \textwidth, bb = 110 210 497 597, clip]{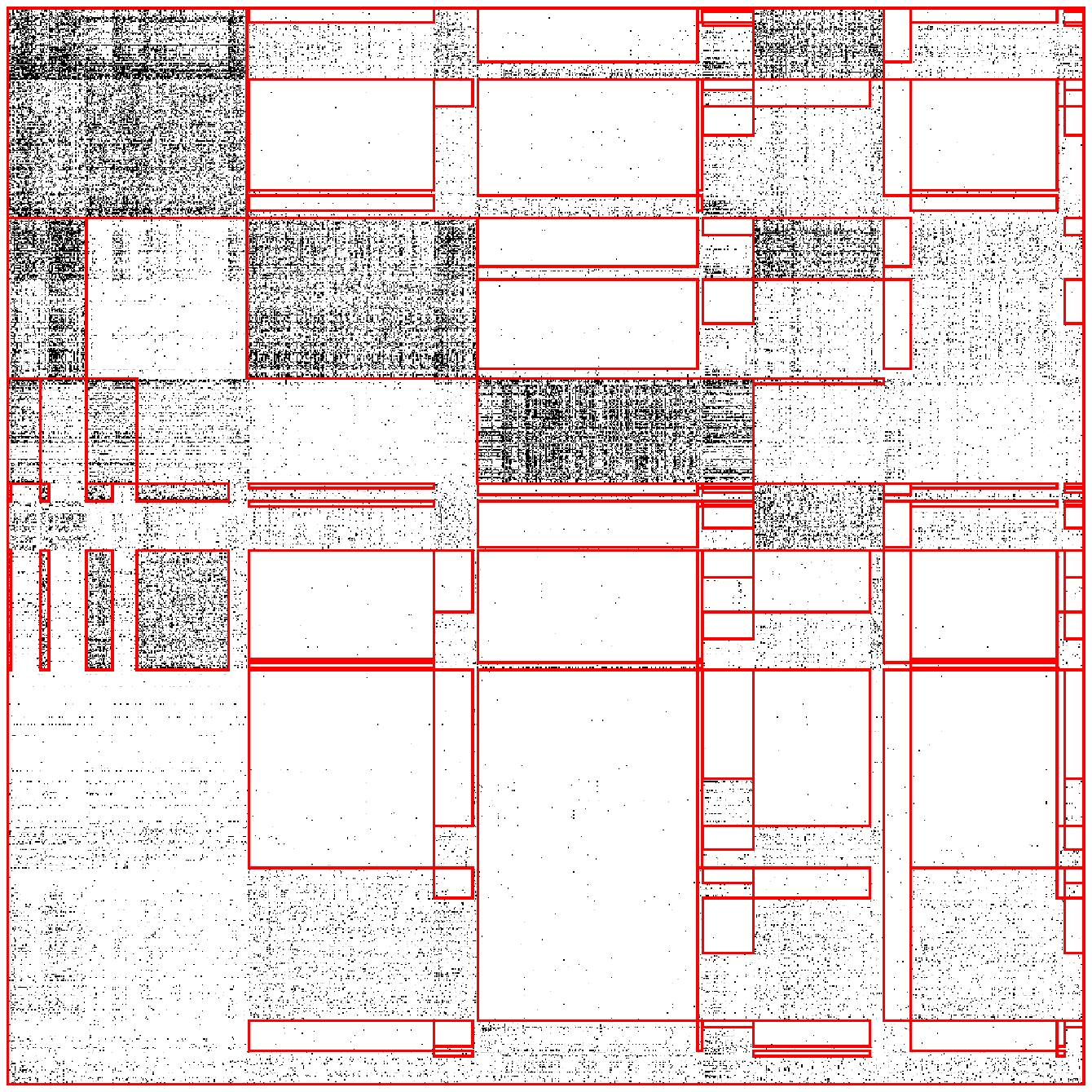}
  \includegraphics[width = 0.18 \textwidth, bb = 110 210 497 597, clip]{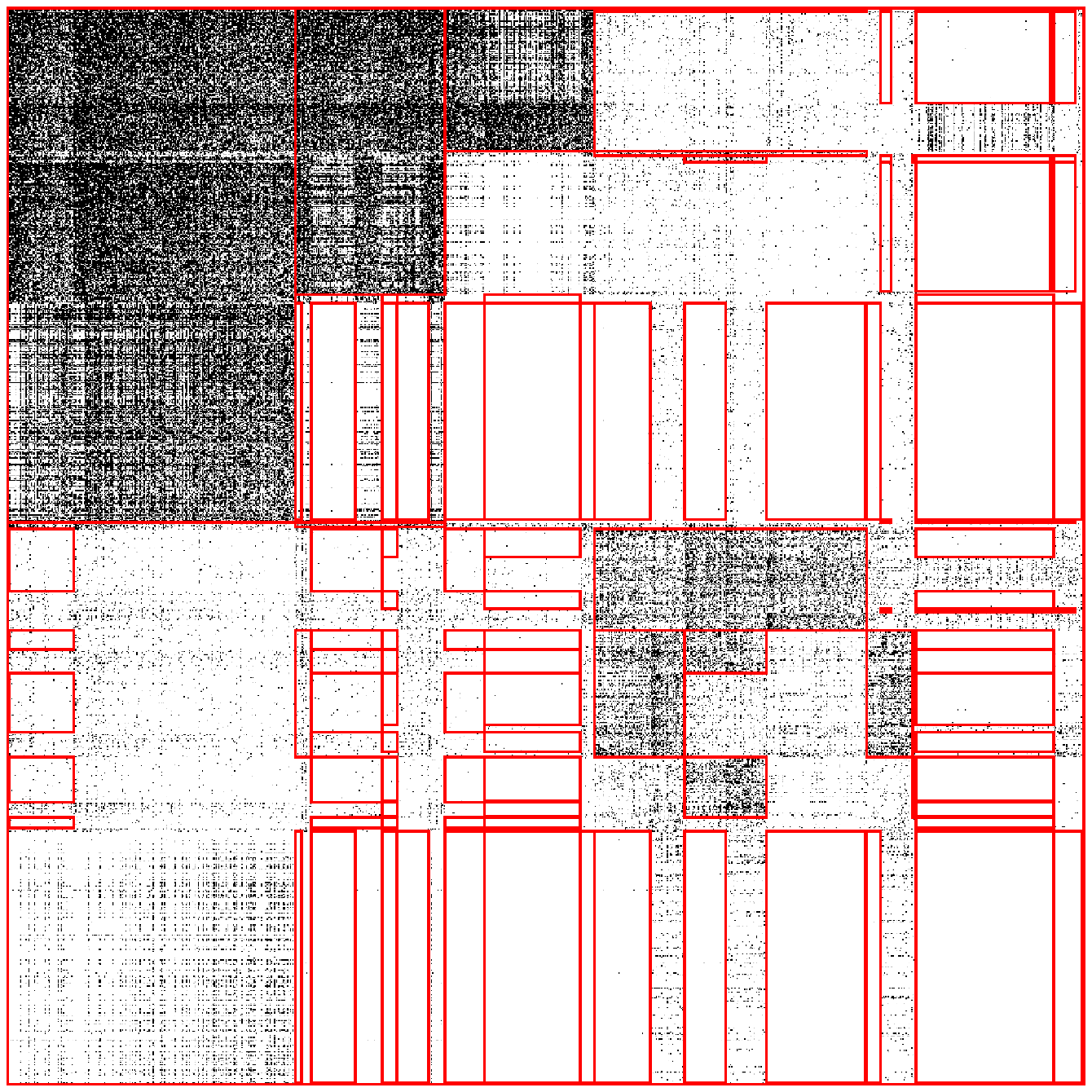}
  \includegraphics[width = 0.18 \textwidth, bb = 110 210 497 597, clip]{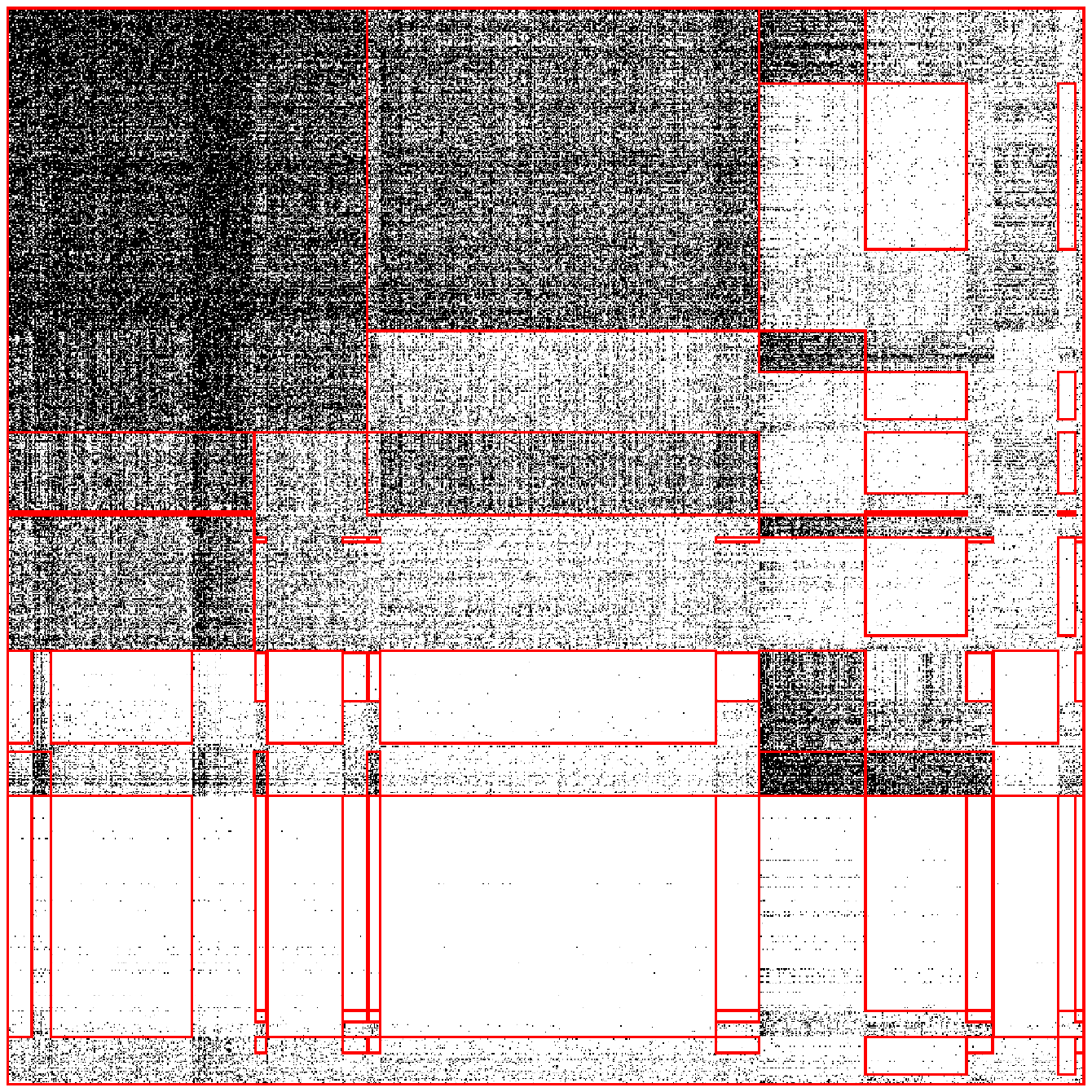}
  \includegraphics[width = 0.18 \textwidth, bb = 110 210 497 597, clip]{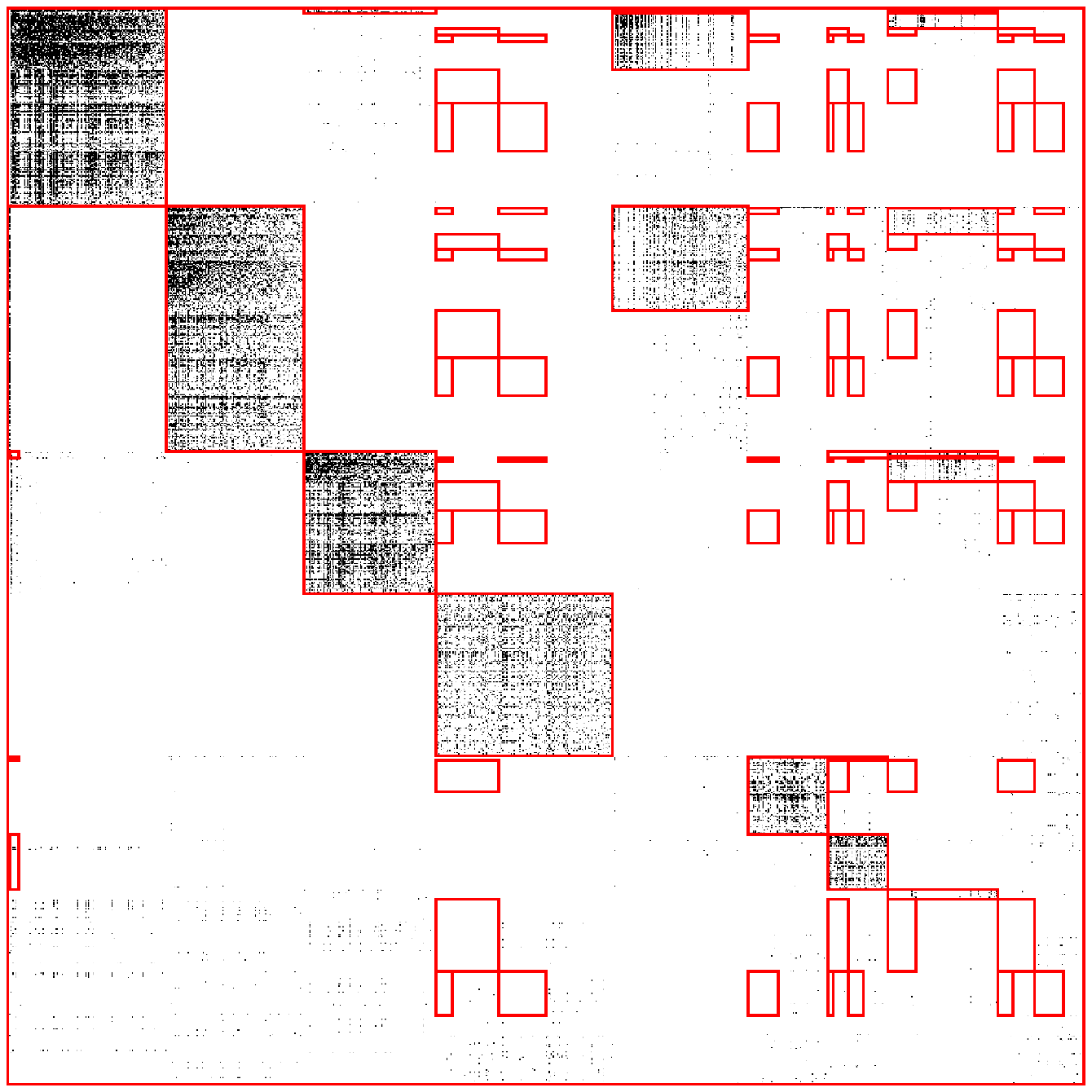}
  \includegraphics[width = 0.18 \textwidth, bb = 110 210 497 597, clip]{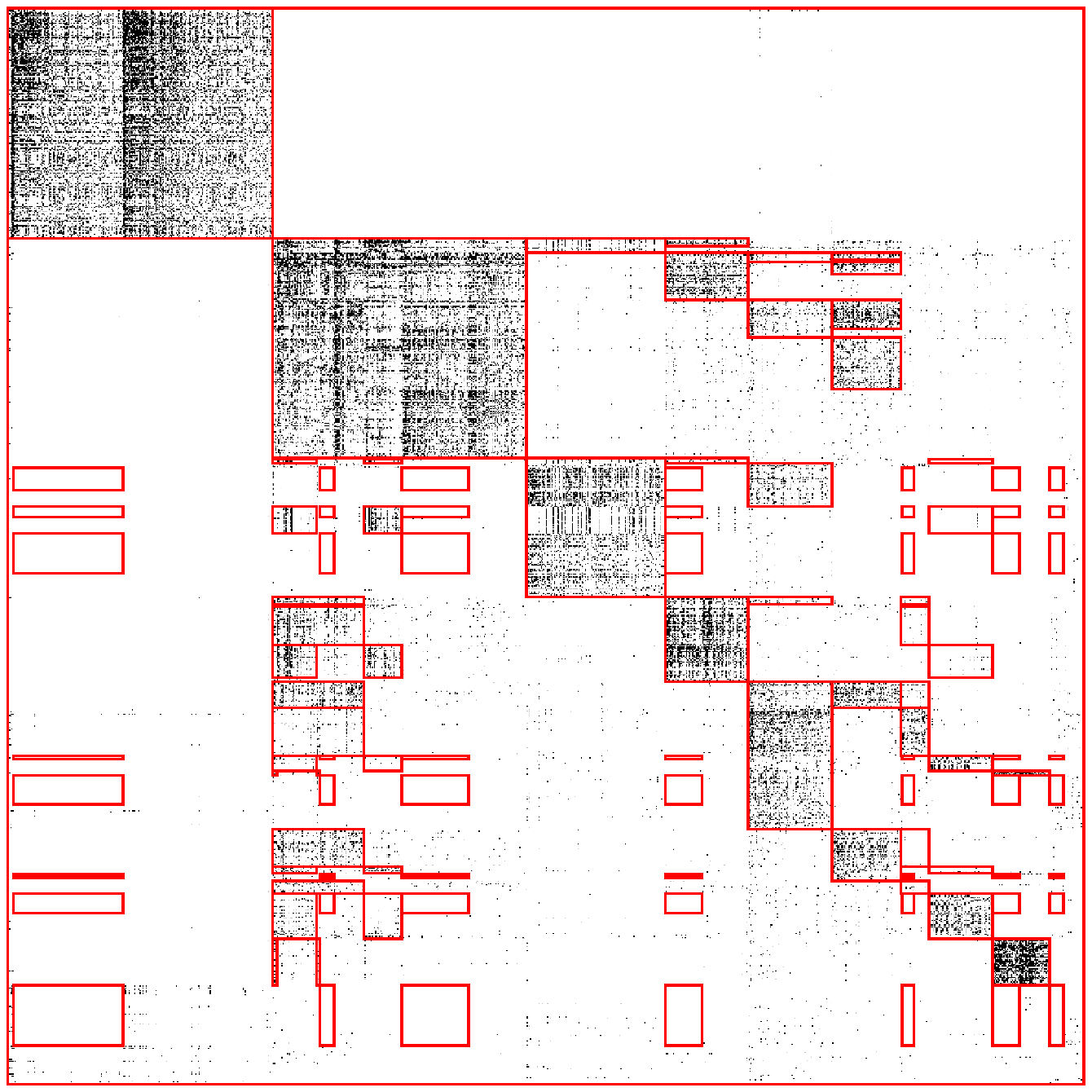}

  \includegraphics[width = 0.18 \textwidth, bb = 82 212 465 595, clip]{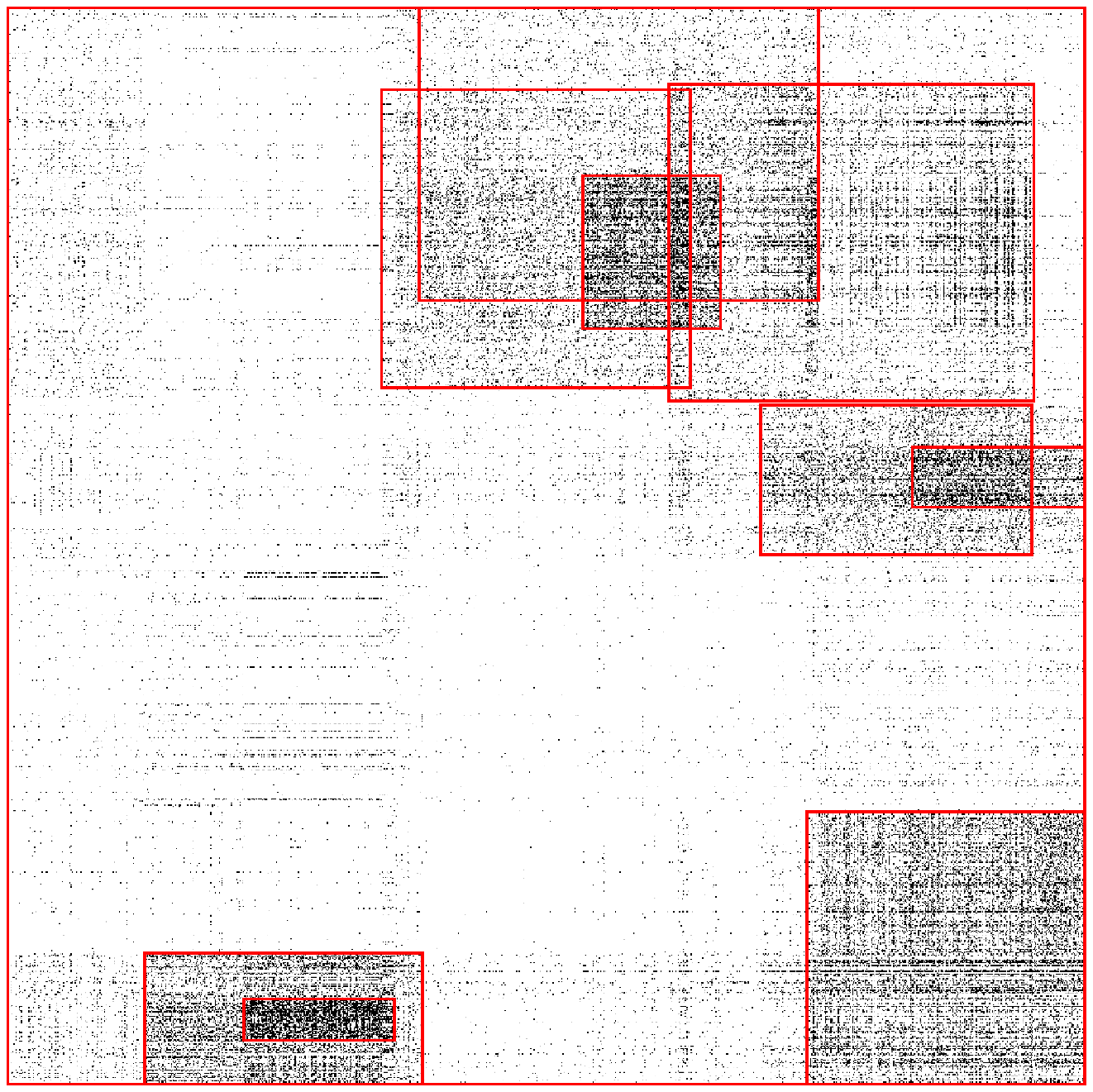}
  \includegraphics[width = 0.18 \textwidth, bb = 82 212 465 595, clip]{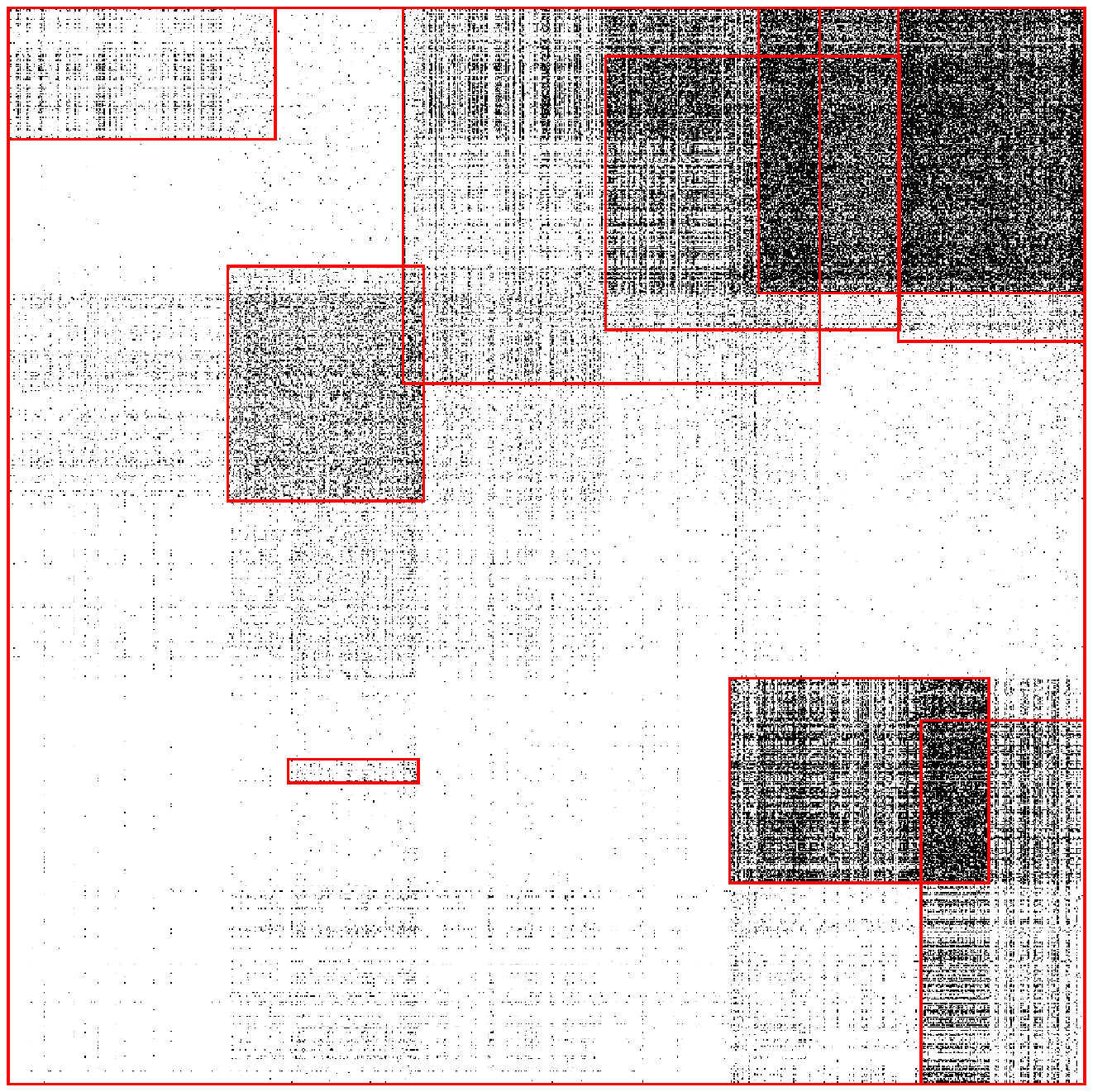}
  \includegraphics[width = 0.18 \textwidth, bb = 82 212 465 595, clip]{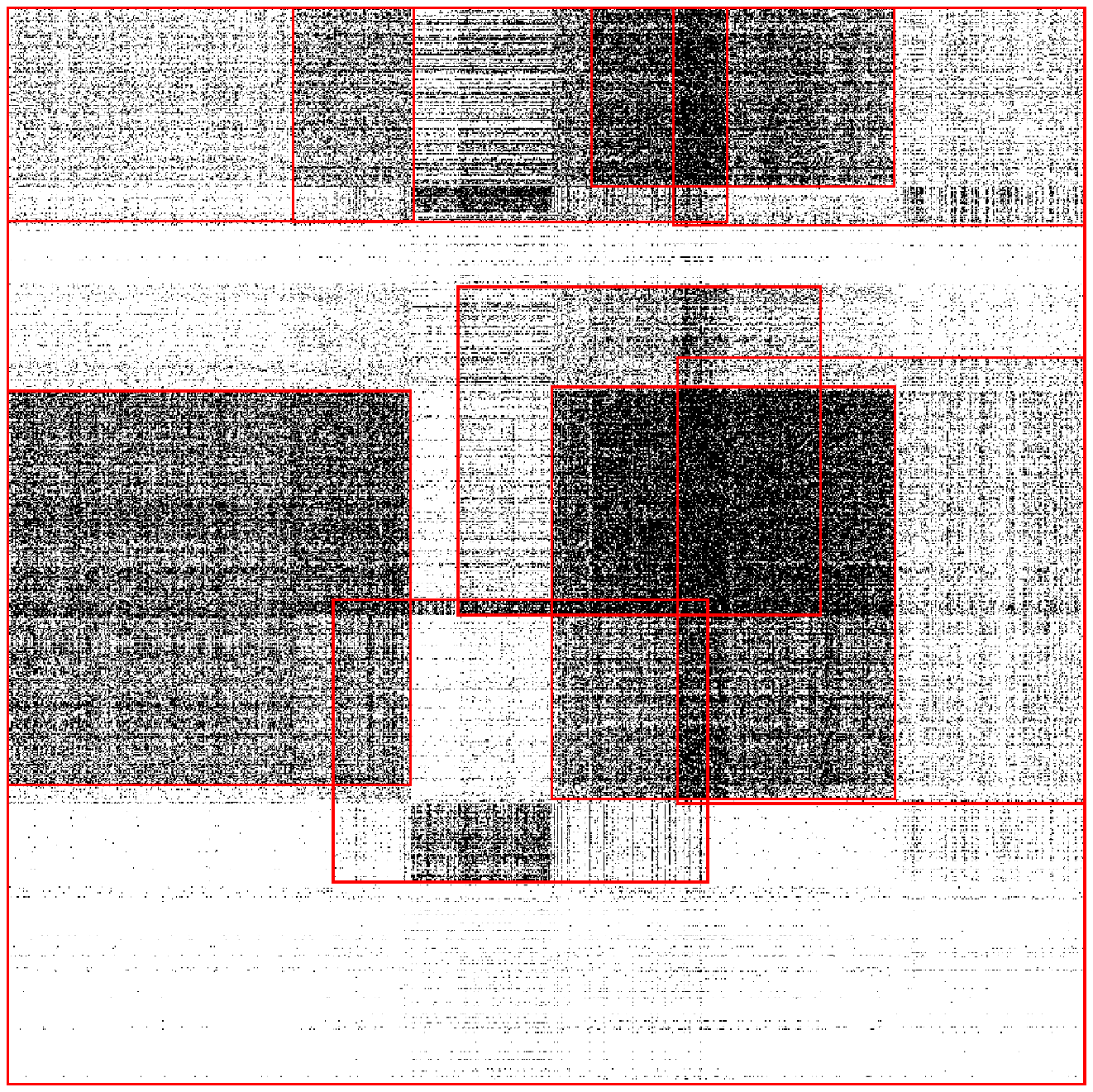}
  \includegraphics[width = 0.18 \textwidth, bb = 82 212 465 595, clip]{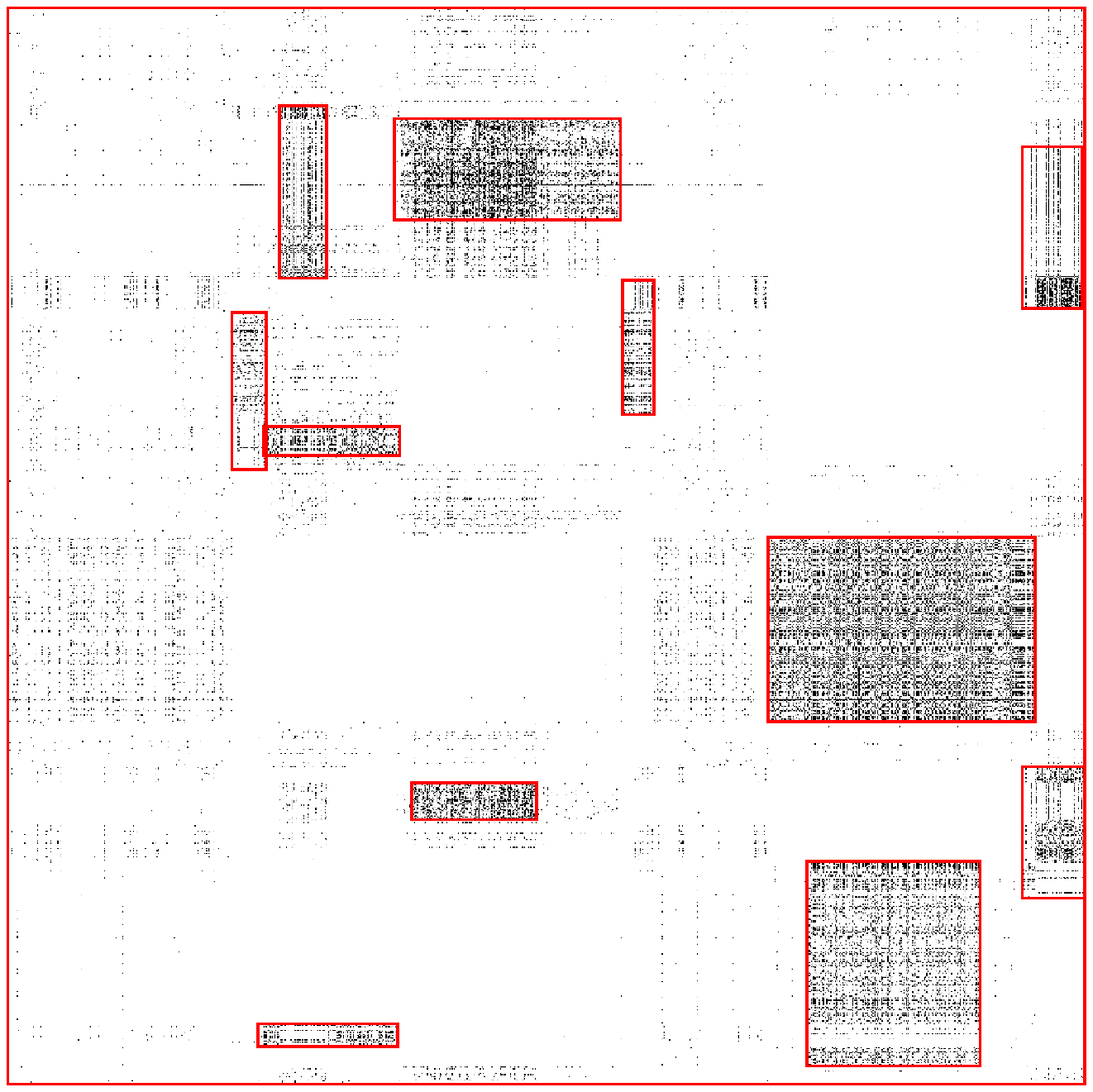}
  \includegraphics[width = 0.18 \textwidth, bb = 116 216 493 593, clip]{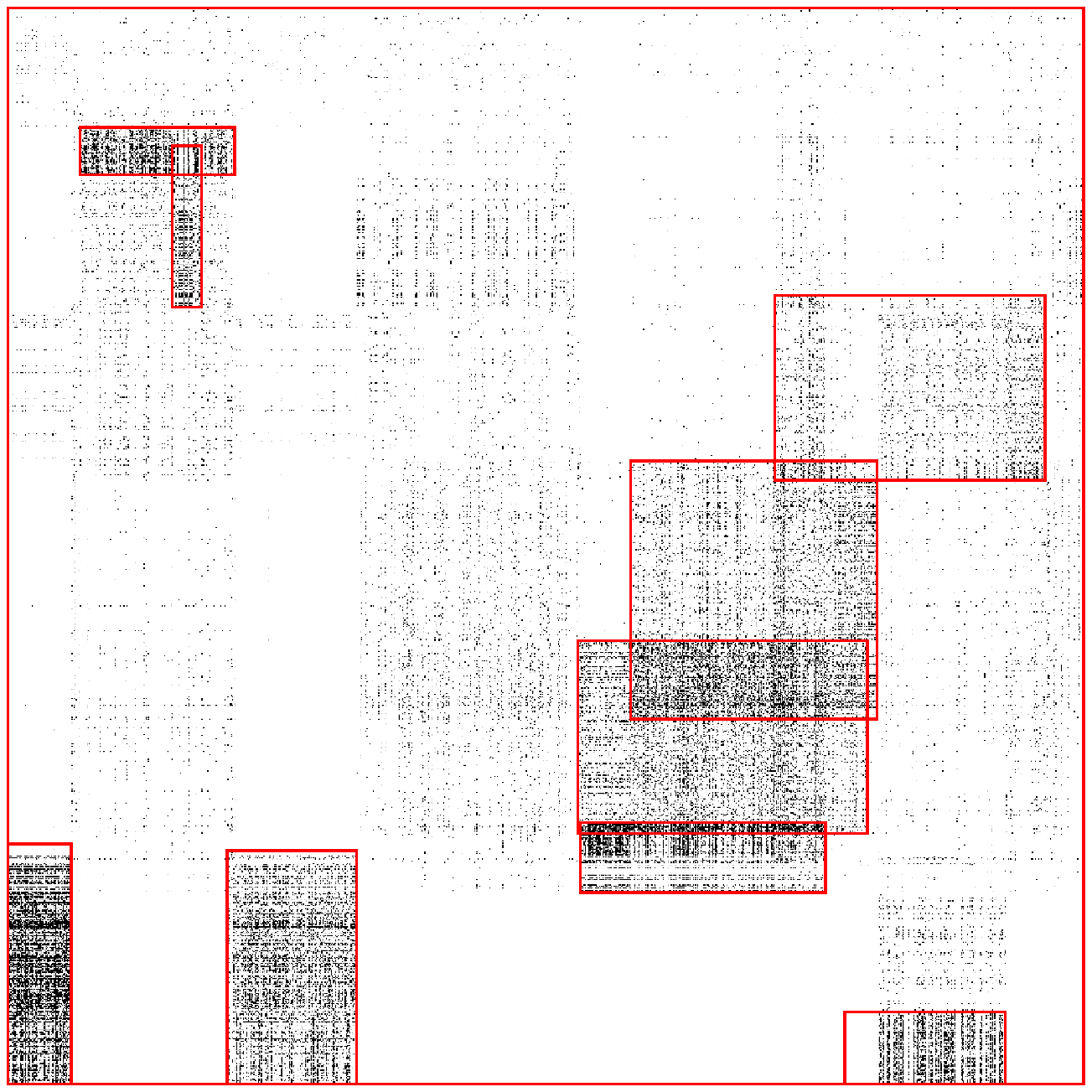}

  \includegraphics[width = 0.18 \textwidth, bb = 115 275 479 569, clip]{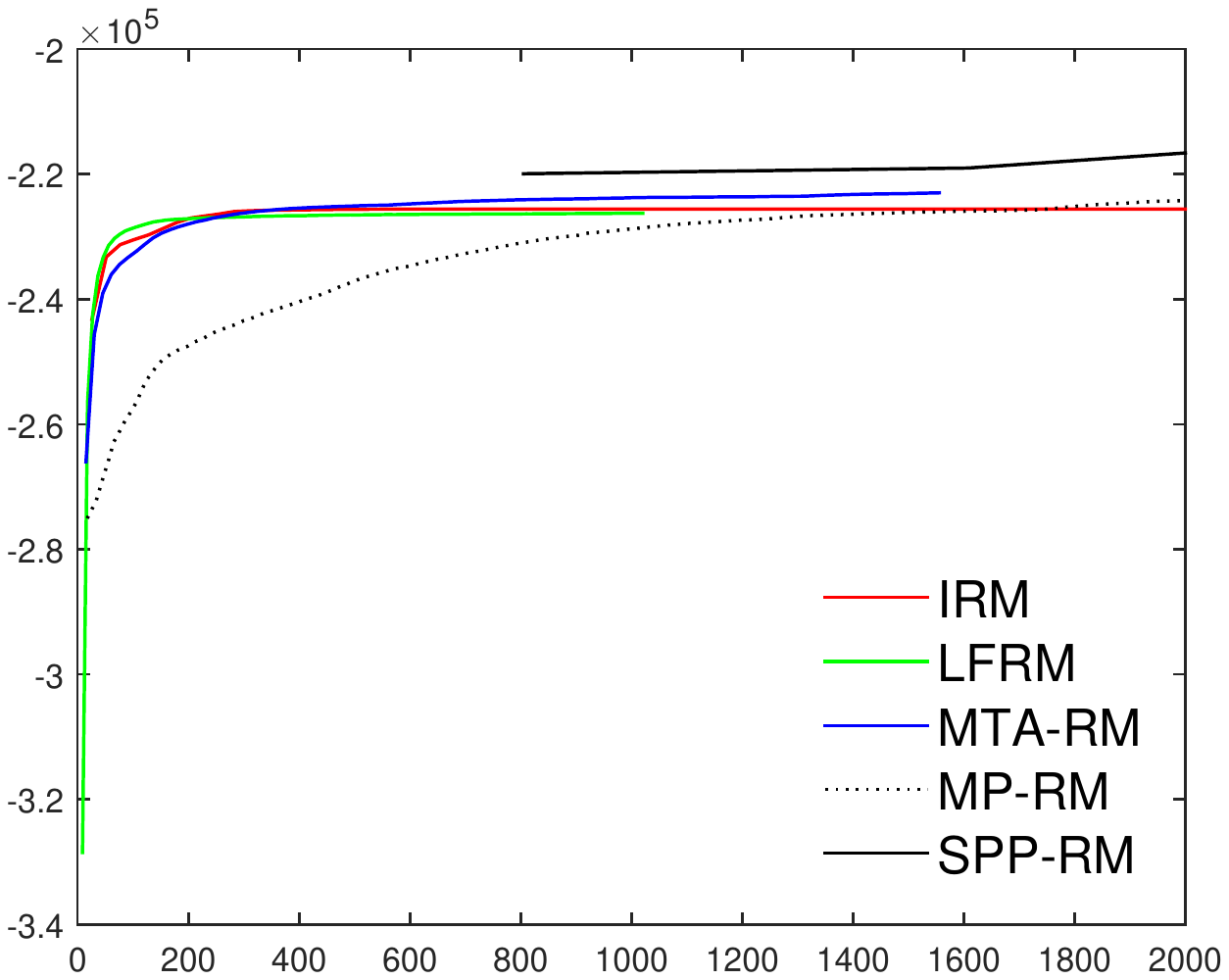}
  \includegraphics[width = 0.18 \textwidth, bb = 115 275 479 569, clip]{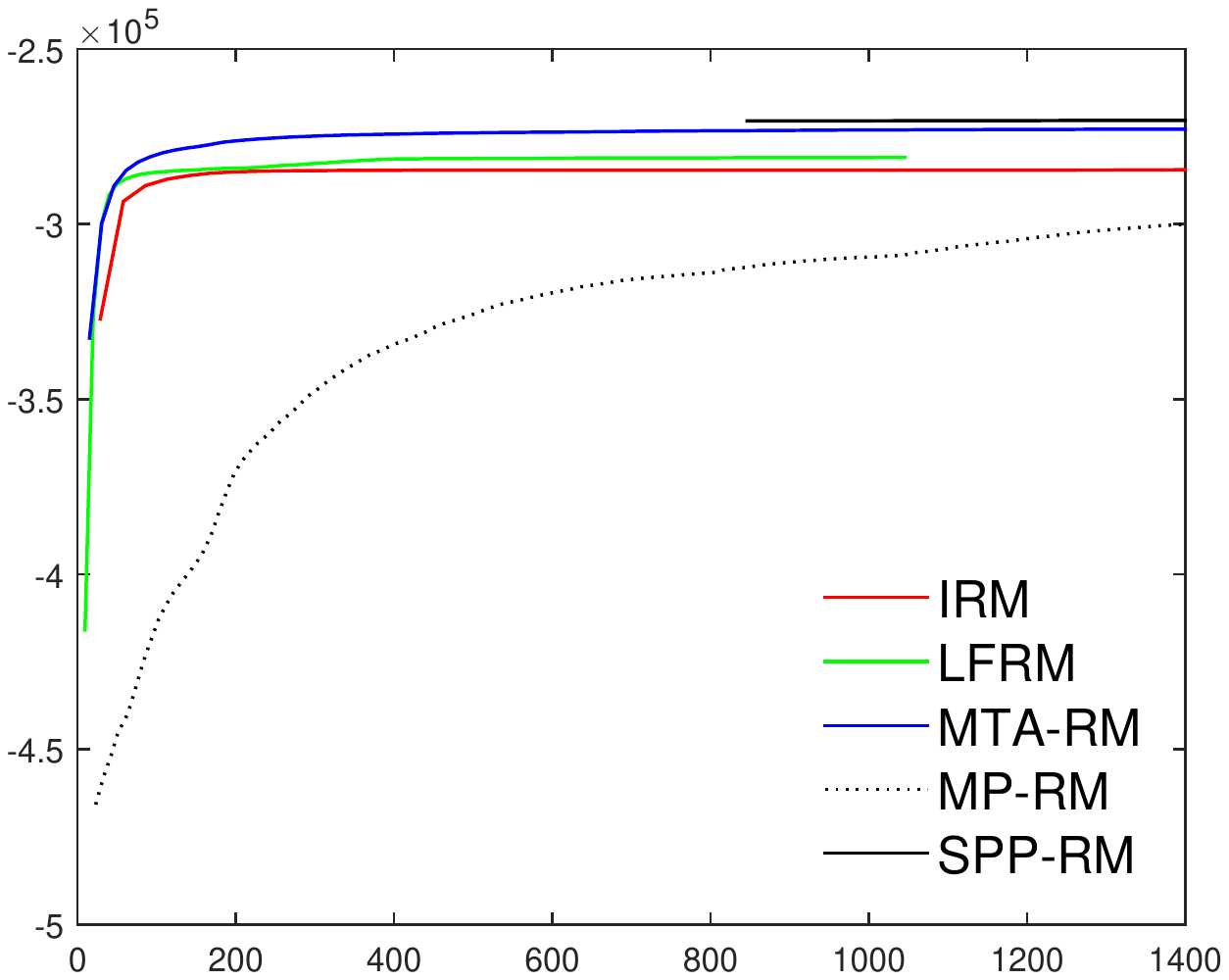}
  \includegraphics[width = 0.18 \textwidth, bb = 115 275 479 569, clip]{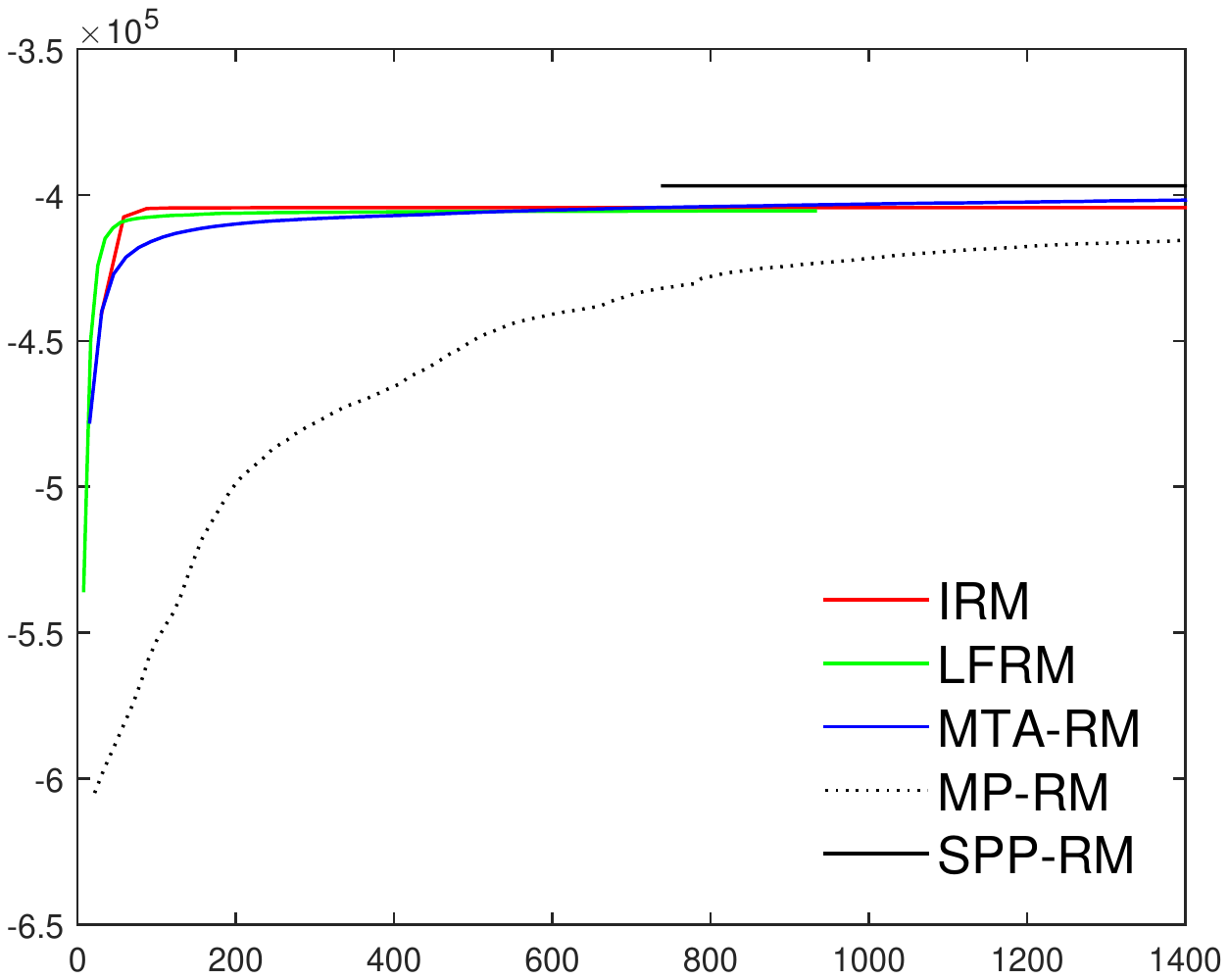}
  \includegraphics[width = 0.18 \textwidth, bb = 115 275 479 569, clip]{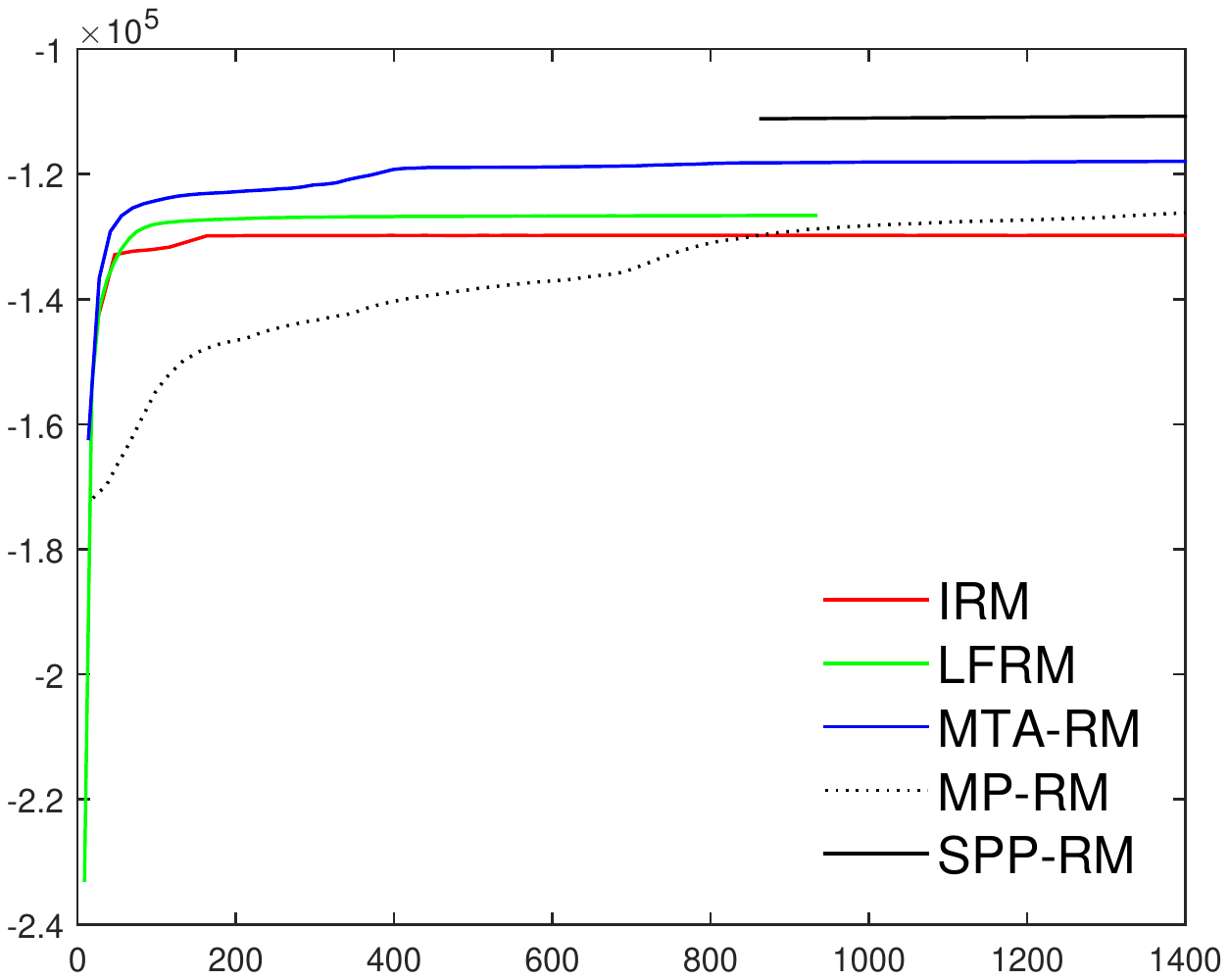}
  \includegraphics[width = 0.18 \textwidth, bb = 115 275 479 569, clip]{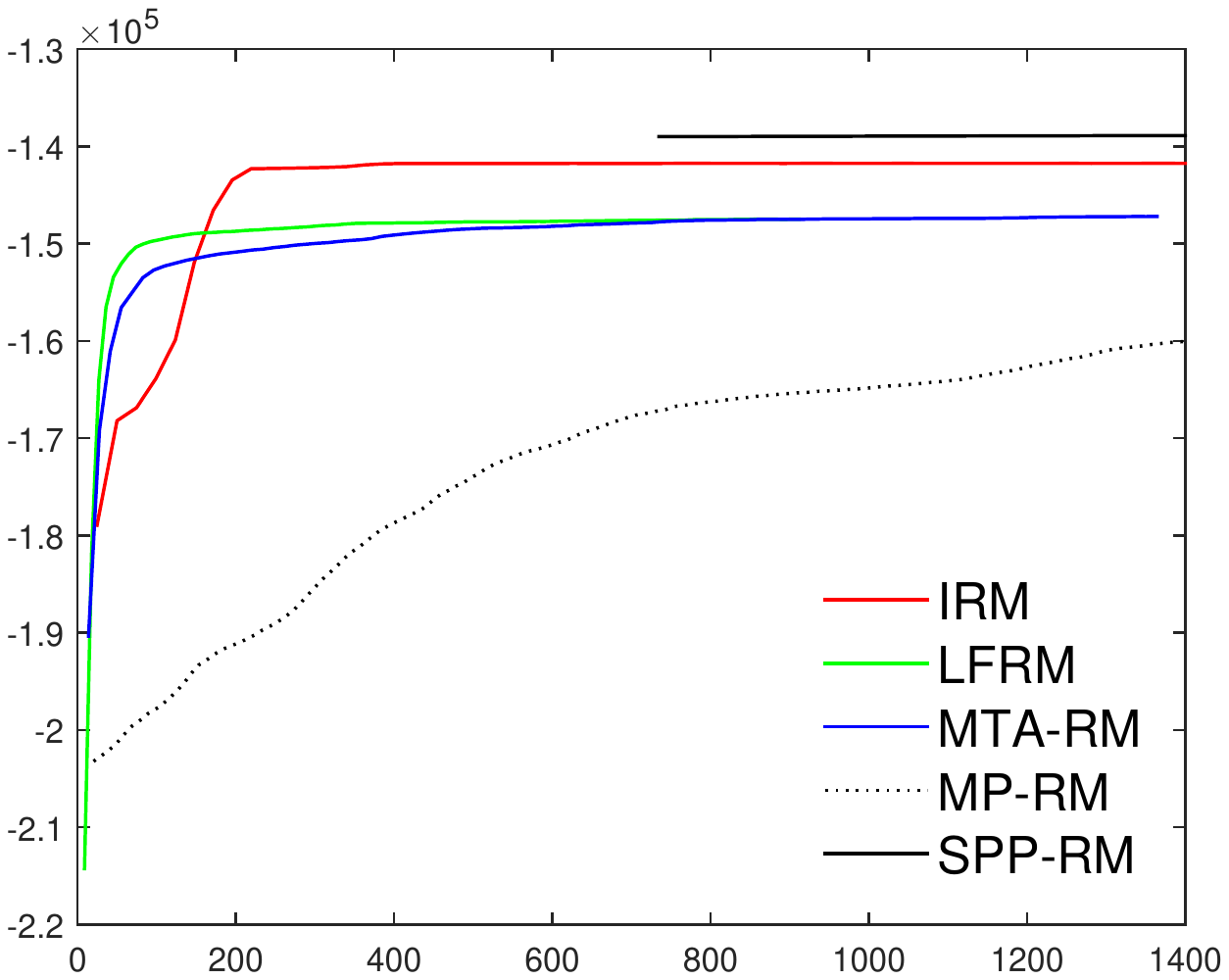}

  \includegraphics[width = 0.18 \textwidth, bb = 115 275 479 562, clip]{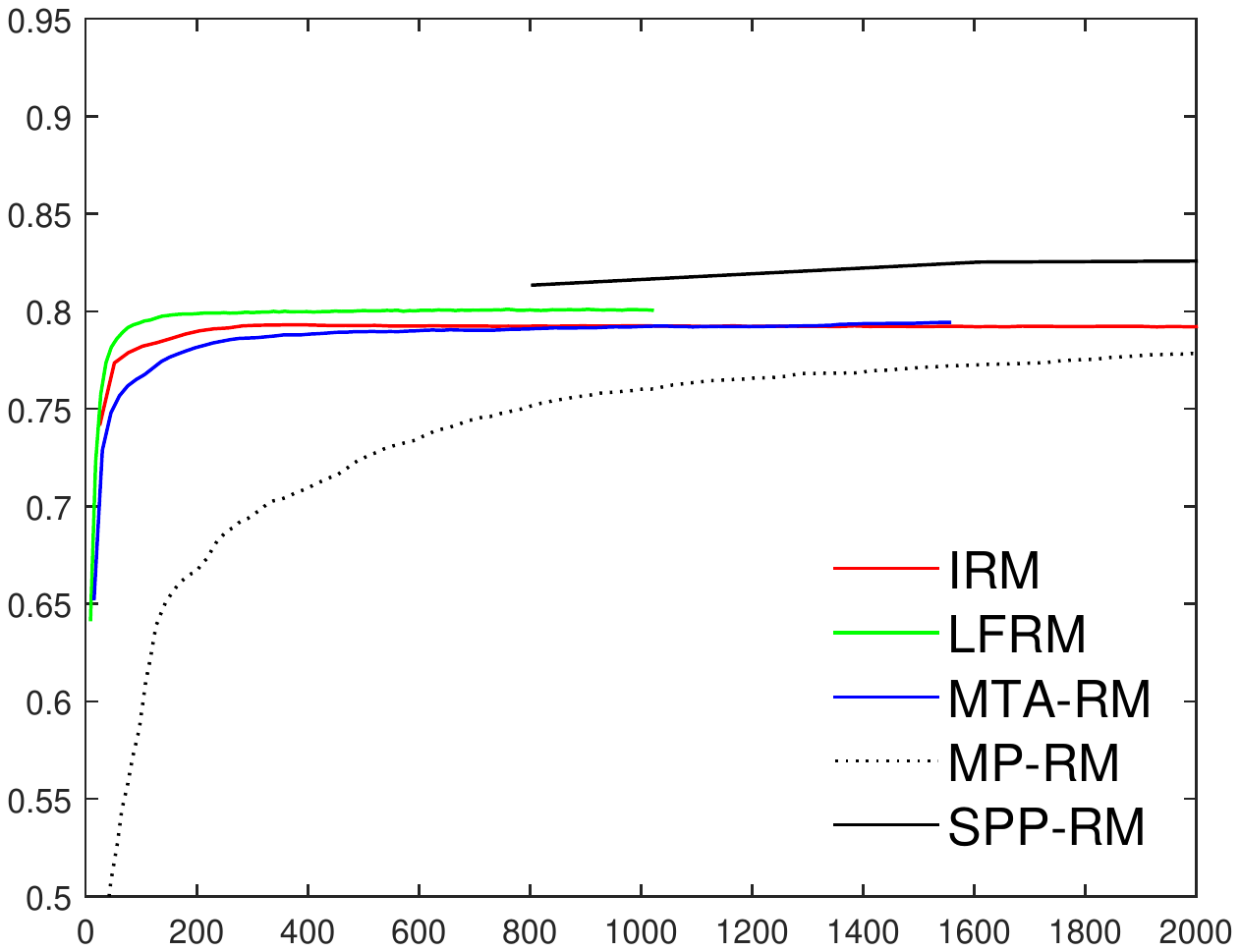}
  \includegraphics[width = 0.18 \textwidth, bb = 115 275 479 562, clip]{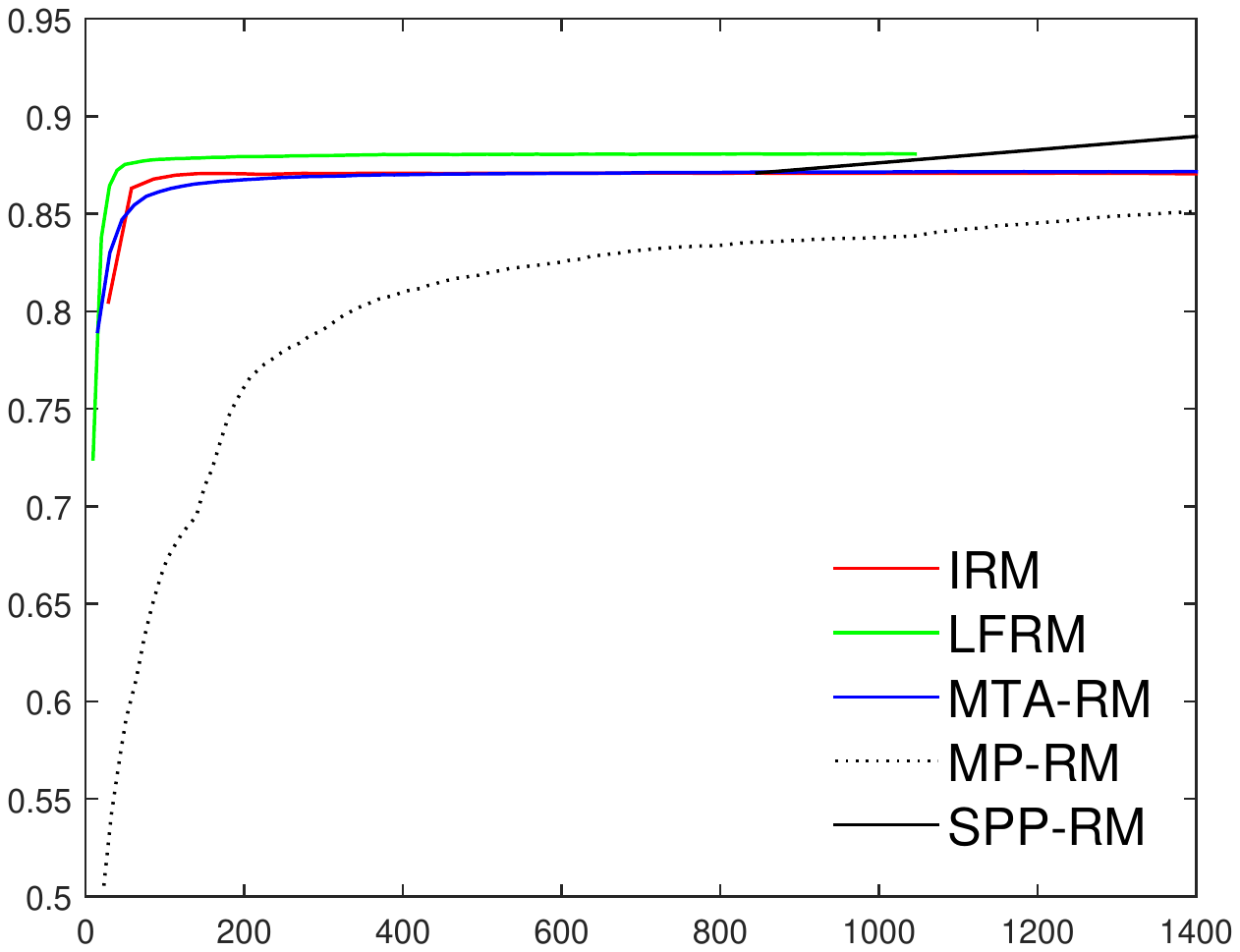}
  \includegraphics[width = 0.18 \textwidth, bb = 115 275 479 562, clip]{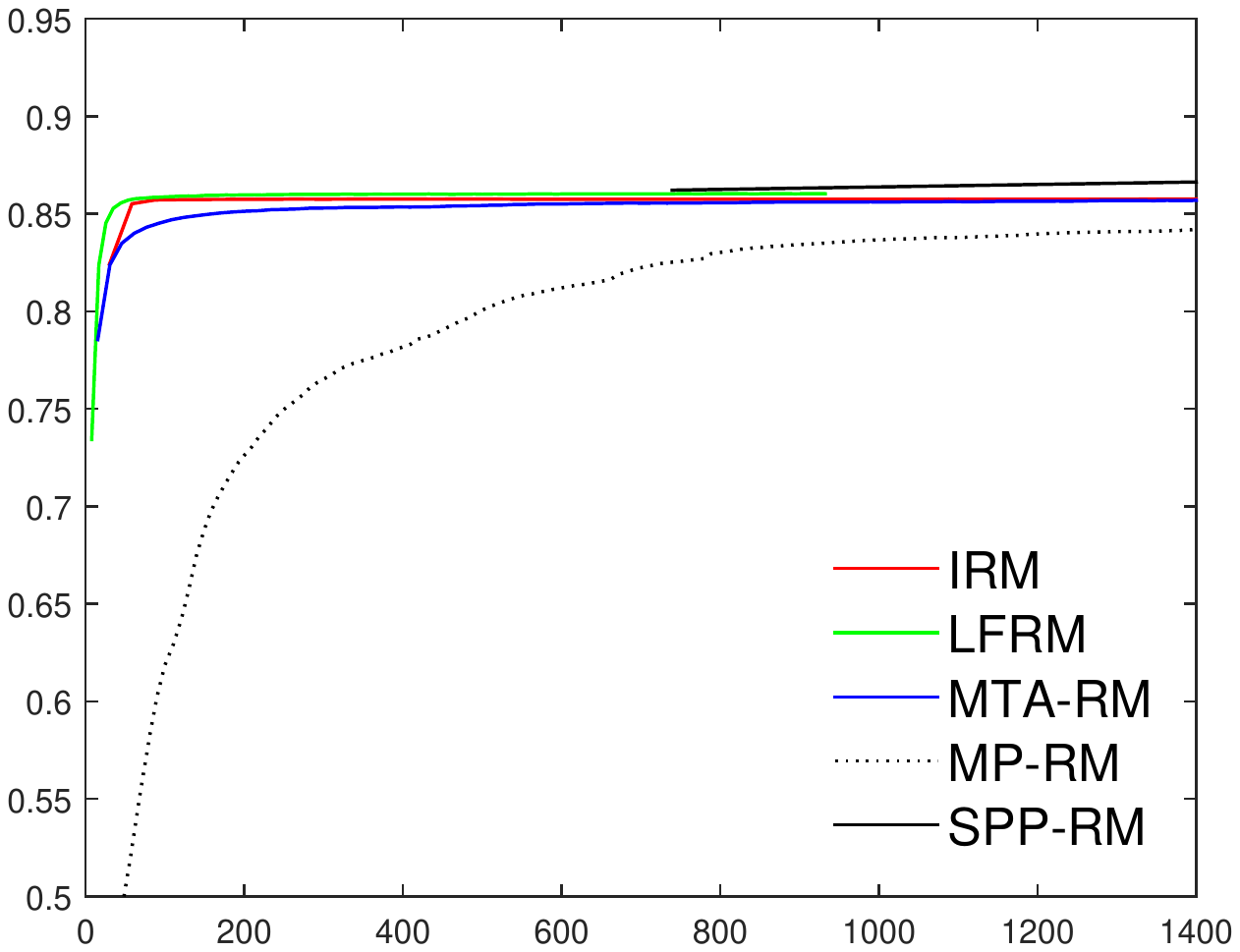}
  \includegraphics[width = 0.18 \textwidth, bb = 115 275 479 562, clip]{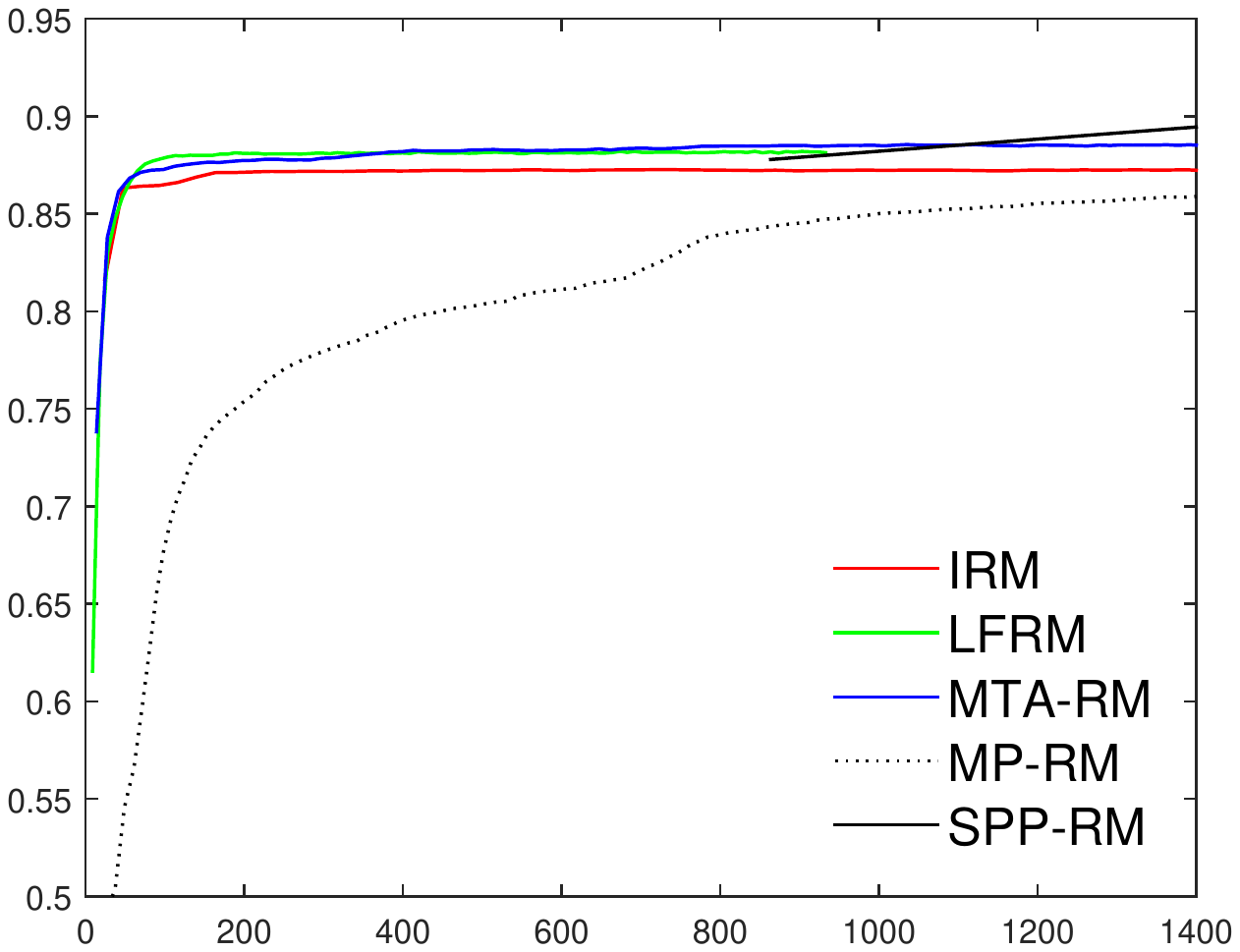}
  \includegraphics[width = 0.18 \textwidth, bb = 115 275 479 562, clip]{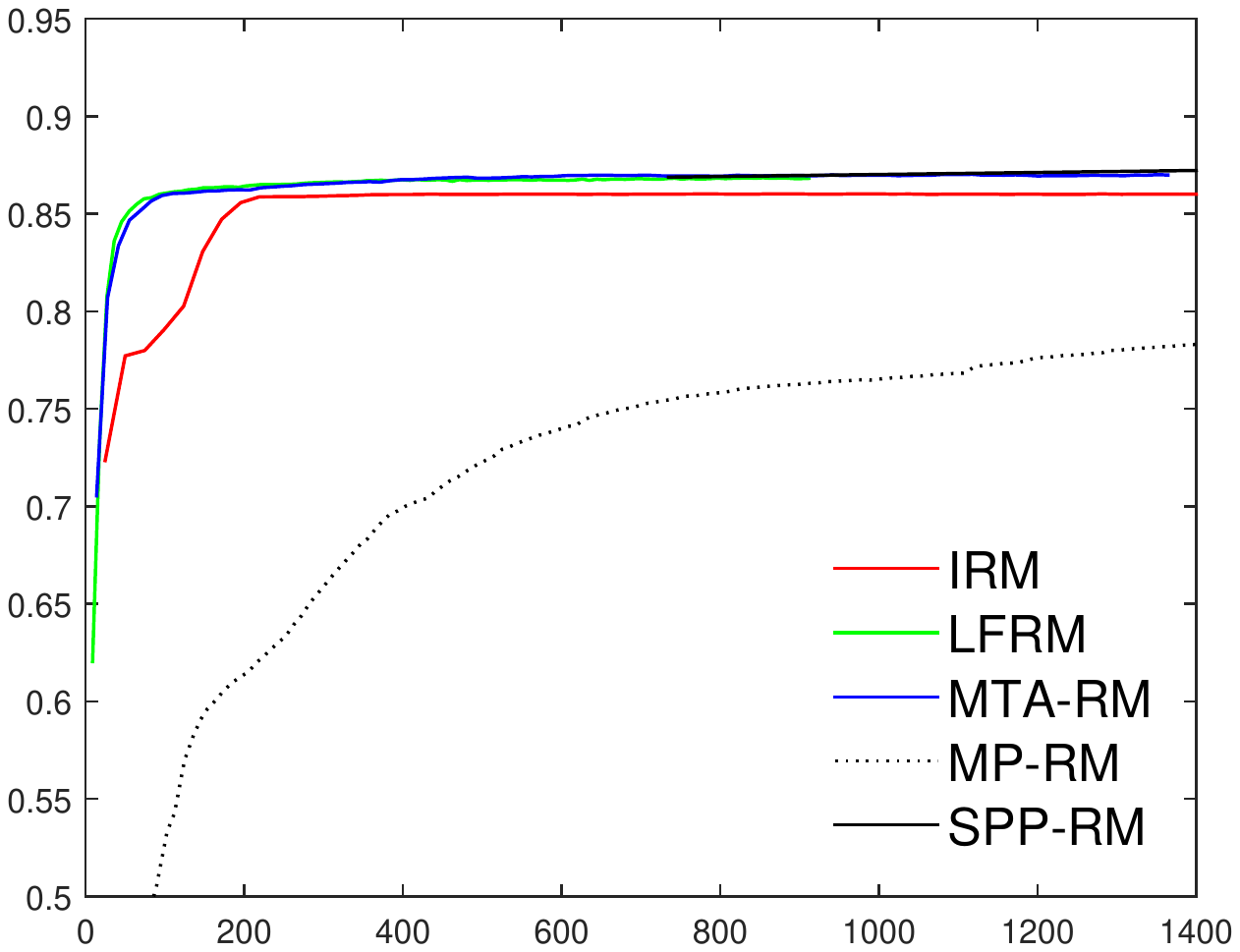}
  \caption{Partition structure visualization and performance comparison on the five data sets: (from left to right) Digg, Flickr, Gplus, Facebook and Twitter. The rows correspond to (from top to bottom) (1) IRM, (2) LFRM, (3) MP-RM, (4) MTA-RM, (5) SPP-RM, (6) training log-likelihood vs. wall-clock time (s) and (7) testing AUC vs. wall-clock time (s). Note that the curves of SPP-RM start from around 800s because the C-SMC sampling in each iteration takes longer time.}
\label{PartiionGraph}
\end{figure*}

  Figure~\ref{PartiionGraph} (rows 1$\sim$5) illustrates the visual patterns of the partition results. As expected, our bounding-based method SPP-RM indeed focuses on describing dense regions of relational data matrices with fewer patches, while the two representative cutting-based methods, IRM and MP-RM, cut sparse regions into more blocks. An interesting observation of SPP-RM is that overlapped patches are very useful in describing inter-community interactions (e.g., patches in Digg, Flickr, and Gplus) and community-in-community interactions (e.g., upper-right corner in Flickr and Gplus). Thus, in addition to improved performance, SPP-RM also produce parsimonious partitions.

  Figure~\ref{PartiionGraph} (rows 6$\sim$7) plot the average performance versus the wall-clock time for investigating the convergence behavior of the compared methods. IRM and LFRM converge fastest because of efficient collapsed Gibbs sampling. MTA-RM also converges fast because it is trained using a simple iterative algorithm. Although SPP-RM takes longer time for each sampling epoch compared to the other methods, it usually can obtain competitive performance in few iterations, thanks to the C-SMC sampler.
  %MP-RM and SPP-RM have similar convergence rate after first 200 seconds. SPP-RM has an inferior initial performance since it starts with no patches; however it overtakes MP-RM very soon because SPP-RM updates all the patches simultaneously in each iteration while MP-RM only updates one leaf block of the $k$d-tree partition structure.

%==========================================================
\section{Conclusion} \label{sec:con}
%==========================================================

  A parsimonious partition process, named Stochastic Patching Process (SPP), is proposed. Instead of the cutting-based strategy, we adopt a bounding-based strategy to attach \emph{i.i.d.} rectangular patches to model dense data regions in the space such that it can avoid unnecessary dissections in sparse regions. We apply SPP to relational modeling and find that SPP can achieve clear performance gain with fewer patches (blocks) compared to the state-of-the-art relational modeling methods.

\section{Acknowledgement}

  We thank the anonymous meta-reviewer of ICML-16 for his/her constructive and very detail comments, which helped us to significantly improve the manuscript.

\newpage
{\small
\bibliography{StochasticPartitionProcessBase}

\begin{thebibliography}{10}

\bibitem{adams2009tractable}
Ryan~P. Adams, Iain Murray, and David~J.C. MacKay.
\newblock Tractable nonparametric {Bayesian} inference in {Poisson} processes
  with {Gaussian} process intensities.
\newblock In {\em ICML}, pages 9--16, 2009.

\bibitem{airoldi2009mixed}
Edoardo~M. Airoldi, David~M. Blei, Stephen~E. Fienberg, and Eric~P. Xing.
\newblock Mixed membership stochastic blockmodels.
\newblock In {\em NIPS}, pages 33--40, 2009.

\bibitem{aldous1981representations}
David~J. Aldous.
\newblock Representations for partially exchangeable arrays of random
  variables.
\newblock {\em Journal of Multivariate Analysis}, 11(4):581--598, 1981.

\bibitem{andrieu2010particle}
Christophe Andrieu, Arnaud Doucet, and Roman Holenstein.
\newblock Particle {Markov chain Monte Carlo} methods.
\newblock {\em Journal of the Royal Statistical Society: Series B (Statistical
  Methodology)}, 72(3):269--342, 2010.

\bibitem{bochner1955harmonic}
Salomon Bochner.
\newblock {\em Harmonic Analysis and the Theory of Probability}.
\newblock University of California Press, 1955.

\bibitem{chung2001course}
Kai~Lai Chung.
\newblock {\em A Course in Probability Theory}.
\newblock Academic Press, 2001.

\bibitem{Givoni06}
Inmar Givoni, Vincent Cheung, and Brendan Frey.
\newblock Matrix tile analysis.
\newblock In {\em UAI}, pages 200--207, 2006.

\bibitem{hoover1979relations}
Douglas~N. Hoover.
\newblock Relations on probability spaces and arrays of random variables.
\newblock {\em Preprint, Institute for Advanced Study, School of Mathematics,
  Princeton, NJ}, 1979.

\bibitem{ishiguro2010dynamic}
Katsuhiko Ishiguro, Tomoharu Iwata, Naonori Ueda, and Joshua~B. Tenenbaum.
\newblock Dynamic infinite relational model for time-varying relational data
  analysis.
\newblock In {\em NIPS}, pages 919--927, 2010.

\bibitem{karrer2011stochastic}
Brian Karrer and Mark~E.J. Newman.
\newblock Stochastic blockmodels and community structure in networks.
\newblock {\em Physical Review E}, 83(1):016107, 2011.

\bibitem{kemp2006learning}
Charles Kemp, Joshua~B. Tenenbaum, Thomas~L. Griffiths, Takeshi Yamada, and
  Naonori Ueda.
\newblock Learning systems of concepts with an infinite relational model.
\newblock In {\em AAAI}, volume~3, pages 381--388, 2006.

\bibitem{LakRoyTeh2014a}
Balaji Lakshminarayanan, Daniel~M. Roy, and Yee~Whye Teh.
\newblock {Mondrian} forests: Efficient online random forests.
\newblock In {\em NIPS}, pages 3140--3148, 2014.

\bibitem{LakOryTeh2015ParticleGibbs}
Balaji Lakshminarayanan, Daniel~M. Roy, and Yee~Whye Teh.
\newblock Particle {Gibbs} for {Bayesian} additive regression trees.
\newblock In {\em AISTATS}, pages 553--561, 2015.

\bibitem{balaji2016aistats}
Balaji Lakshminarayanan, Daniel~M. Roy, and Yee~Whye Teh.
\newblock Mondrian forests for large-scale regression when uncertainty matters.
\newblock In {\em AISTATS}, pages 1478--1487, 2016.

\bibitem{leskovec2010predicting}
Jure Leskovec, Daniel Huttenlocher, and Jon Kleinberg.
\newblock Predicting positive and negative links in online social networks.
\newblock In {\em WWW}, pages 641--650, 2010.

\bibitem{liu2000multiple}
Jun~S. Liu, Faming Liang, and Wing~Hung Wong.
\newblock The multiple-try method and local optimization in {Metropolis}
  sampling.
\newblock {\em Journal of the American Statistical Association},
  95(449):121--134, 2000.

\bibitem{facebook_mcauley2012learning}
Julian~J. McAuley and Jure Leskovec.
\newblock Learning to discover social circles in ego networks.
\newblock In {\em NIPS}, volume 2012, pages 548--556, 2012.

\bibitem{miller2009nonparametric}
Kurt Miller, Michael~I. Jordan, and Thomas~L. Griffiths.
\newblock Nonparametric latent feature models for link prediction.
\newblock In {\em NIPS}, pages 1276--1284, 2009.

\bibitem{nakano2014rectangular}
Masahiro Nakano, Katsuhiko Ishiguro, Akisato Kimura, Takeshi Yamada, and
  Naonori Ueda.
\newblock Rectangular tiling process.
\newblock In {\em ICML}, pages 361--369, 2014.

\bibitem{nowicki2001estimation}
Krzysztof Nowicki and Tom~A.B. Snijders.
\newblock Estimation and prediction for stochastic block structures.
\newblock {\em Journal of the American Statistical Association},
  96(455):1077--1087, 2001.

\bibitem{TPAMI2014peterdaniel}
Peter Orbanz and Daniel~M. Roy.
\newblock {Bayesian} models of graphs, arrays and other exchangeable random
  structures.
\newblock {\em IEEE Transactions on Pattern Analysis and Machine Intelligence},
  37(02):437--461, 2015.

\bibitem{porteous2008multi}
Ian Porteous, Evgeniy Bart, and Max Welling.
\newblock {Multi-HDP}: A non parametric {Bayesian} model for tensor
  factorization.
\newblock In {\em AAAI}, pages 1487--1490, 2008.

\bibitem{roy2011thesis}
Daniel~M. Roy.
\newblock {\em Computability, Inference and Modeling in Probabilistic
  Programming}.
\newblock PhD thesis, MIT, 2011.

\bibitem{roy2007learning}
Daniel~M. Roy, Charles Kemp, Vikash Mansinghka, and Joshua~B. Tenenbaum.
\newblock Learning annotated hierarchies from relational data.
\newblock In {\em NIPS}, pages 1185--1192, 2007.

\bibitem{roy2009mondrian}
Daniel~M. Roy and Yee~Whye Teh.
\newblock The {Mondrian} process.
\newblock In {\em NIPS}, pages 1377--1384, 2009.

\bibitem{nonpa2013schmidt}
Mikkel~N. Schmidt and Morten M{\o}rup.
\newblock Nonparametric {Bayesian} modeling of complex networks: An
  introduction.
\newblock {\em IEEE Signal Processing Magazine}, 30(3):110--128, 2013.

\bibitem{wang2011nonparametric}
Pu~Wang, Kathryn~B. Laskey, Carlotta Domeniconi, and Michael~I. Jordan.
\newblock Nonparametric {Bayesian} co-clustering ensembles.
\newblock In {\em SDM}, pages 331--342, 2011.

\bibitem{Zafarani+Liu:2009}
Reza Zafarani and Huan Liu.
\newblock Social computing data repository at {ASU}, 2009.

\end{thebibliography}
\bibliographystyle{plain}
}

\section*{Alternative Construction of SPP}

    An alternative construction of SPP which is equivalent to the one introduced in Section 3 is as follows:
\begin{enumerate}
  \item Sample the number of \emph{nonempty} patches $K_{\tau}\sim\mbox{Poisson}(\tau S_X\cdot P(S_{\Box^X}>0))$, where $P(S_{\Box^X}>0) = \prod_{d=1}^D \frac{1}{N_X^{(d)}} \cdot \left[\theta+(1-\theta)N_X^{(d)}\right]$;
  \item Given $K_{\tau}$, sample \emph{i.i.d.} \emph{nonempty} patches $\{\Box_k\}_{k=1}^{K_\tau}$. For $k = 1,\ldots,K_\tau$, $d = 1,\ldots,D$
\begin{enumerate}
  \item Sample the initial position $s_k^{(d)}$ of $\Box_k$ from $\{1,2,\ldots,N_X^{(d)}\}$ in proportion to $\{1, (1-\theta), \ldots, (1-\theta)\}$;
  \item Sample the side-length $l_k^{(d)}$ using the distribution
      %\footnote{$P(l_{k}^{(d)})$ follows a geometric distribution for $1\le l_{k}^{(d)} < L_*$; while the probability for $l_{k}^{(d)} = L_*$ is the sum of the probability mass of the geometric distribution for $l_{k}^{(d)} \ge L_*$.}
\begin{eqnarray}
  P(l_{k}^{(d)})=\left\{\begin{array}{ll}
  \theta^{l_{k}^{(d)}-1}(1-\theta), & 1\le l_{k}^{(d)} < L_*; \\
  \theta^{l_{k}^{(d)}-1}, & l_{k}^{(d)}= L_*, \\
  \end{array}\right. \nonumber
    \end{eqnarray}
where $L_*=N_X^{(d)}-s_{k}^{(d)}+1$;
\end{enumerate}
\item Sample $K_{\tau}$ \emph{i.i.d.} time points uniformly in $(0,\tau]$ and index them to satisfy $t_1<\ldots<t_{K_{\tau}}$. Set the cost of $\Box_k$ as $m_k = t_{k}-t_{k-1} (t_0=0)$ and the rate of $\Box_k$ as $\omega_k = m_k/S_{\Box_k}$, where $S_{\Box_k}=\prod_{d=1}^D l_k^{(d)}$.
\end{enumerate}
  In this way, one can directly sample \emph{nonempty} patches through thinning the Poisson process which is used for generating candidate patches.

\section*{Proof for Proposition 1}

\begin{proof}
  $\forall d\in \{1, \cdots, D\}$, we have the probability of side-length $l^{(d)}>0$ as
\begin{eqnarray} \label{eq_nonemptypatch0}
  P(l^{(d)}>0) = \frac{1}{N_X^{(d)}} + \frac{1-\theta}{N_X^{(d)}} \cdot (N_X^{(d)}-1)=  \frac{1}{N_X^{(d)}} \cdot \left[\theta+(1-\theta)N_X^{(d)}\right].
\end{eqnarray}
According to the thinning property of the Poisson process, we thus have the following expected number of nonempty patches:
\begin{eqnarray} \label{eq_nonintensity}
\mathbb{E}(K_{\tau}) = \tau\cdot S_X\cdot \prod_{d=1}^D P(l^{(d)}>0) =  \tau\cdot\prod_{d=1}^D \left[\theta+(1-\theta)N_X^{(d)}\right].
\end{eqnarray}

For the expectation of side-length $l^{(d)}$, we need to consider its all possible initial positions $s^{(d)}$:
\begin{eqnarray}
 \mathbb{E}(l^{(d)}) & =& \sum_{l^{(d)}=1}^{N_X^{(d)}} l^{(d)} \cdot P_l(l^{(d)}) =  \sum_{l^{(d)}=1}^{N_X^{(d)}} l^{(d)} \cdot \sum_{s^{(d)}=1}^{N_X^{(d)}- l^{(d)}+1} P(l^{(d)}|s^{(d)})P_s(s^{(d)}) \nonumber \\
& = &\sum_{s^{(d)}=1}^{N_X^{(d)}}P_s(s^{(d)})\cdot\sum_{l^{(d)}=1}^{N_X^{(d)}- s^{(d)}+1} l^{(d)}\cdot P(l^{(d)}|s^{(d)}).
\end{eqnarray}
For convenience of notation, we let
\begin{equation}
E_n = P_s(s^{(d)}=n)\cdot\sum_{l^{(d)}=1}^{N_X^{(d)}- s^{(d)}+1} l^{(d)}\cdot P(l^{(d)}|s^{(d)}=n).
\end{equation}

In the case of $s^{(d)}=1$ (the initial position is on the boundary of $X$), we have $P_s(s^{(d)}=1) = \frac{1}{1+(N_X^{(d)}-1)(1-\theta)}$ according to Step 2(a) in ``Alternative Construction of SPP''; and the expected side-length is
\begin{equation}
\begin{split}
E_1 = & \frac{1}{1+(N_X^{(d)}-1)(1-\theta)}\left[(1-\theta)+2\theta(1-\theta)+3\theta^2(1-\theta)+\cdots\right.\\
& \left.+(N_X^{(d)}-1)\theta^{N_X^{(d)}-2}(1-\theta)+N_X^{(d)}\theta^{{N_X^{(d)}-1}}\right].
\end{split}
\end{equation}
To simplify the above equation, we can compute $E_1-\theta E_1$ to cancel out many terms and obtain the following equation
\begin{equation}
E_1-\theta E_1 =
\frac{1-\theta}{\theta+(1-\theta)N_X^{(d)}}\left[1+\theta+\theta^2+\cdots+\theta^{{N_X^{(d)}}-1}\right].
%\begin{split}
%E_1-\theta E_1 = & \frac{1}{\theta+(1-\theta)N_X^{(d)}}\left[(1-\theta)+\theta(1-\theta)+\theta^2(1-\theta)+\cdots\right.\\
%& \left.+\theta^{{N_X^{(d)}}-2}(1-\theta)+N_X^{(d)}\theta^{{N_X^{(d)}}-1}-(N_X^{(d)}-1)\theta^{{N_X^{(d)}-1}}(1-\theta)-N_X^{(d)}\theta^{{N_X^{(d)}}}\right]\\
%=&\frac{1-\theta}{\theta+(1-\theta)N_X^{(d)}}\left[1+\theta+\theta^2+\cdots+\theta^{{N_X^{(d)}}-1}\right]
%\end{split}
\end{equation}
Thus we have
\begin{eqnarray}
E_1 = \frac{1}{\theta+(1-\theta)N_X^{(d)}}\cdot\frac{1-\theta^{N_X^{(d)}}}{1-\theta}.
\end{eqnarray}

Similarly, in the case of $s^{(d)}=n\in\{2, 3, \cdots, N\}$ (the initial position is not on the boundary of $X$), we have
\begin{equation}
E_n = \frac{1-\theta}{\theta+(1-\theta)N_X^{(d)}}\cdot\frac{1-\theta^{N_X^{(d)}-n+1}}{1-\theta}.
\end{equation}

Now the expectation of $l^{(d)}$ becomes
\begin{equation} \label{expecedlength}
\begin{split}
\mathbb{E}(l^{(d)}) = & \frac{1}{\theta+(1-\theta)N_X^{(d)}}\cdot\frac{1-\theta^{N_X^{(d)}}}{1-\theta}+\frac{1-\theta}{\theta+(1-\theta)N_X^{(d)}}\cdot\sum_{n=2}^{N_X^{(d)}}\frac{1-\theta^{N_X^{(d)}-n+1}}{1-\theta}\\
= & \frac{1}{\theta+(1-\theta)N_X^{(d)}}\left[\frac{1-\theta^{N_X^{(d)}}}{1-\theta}+\sum_{n=1}^{N_X^{(d)}-1}(1-\theta^n)\right]\\
= & \frac{N_X^{(d)}}{\theta+(1-\theta)N_X^{(d)}}.
\end{split}
\end{equation}

By combining Eq.~(\ref{eq_nonintensity}) and Eq.~(\ref{expecedlength}), we can obtain the expected volume of all patches:
\begin{equation}
\mathbb{E}(K_{\tau})\cdot \prod_{d=1}^D \mathbb{E}(l^{(d)}) = \tau\cdot\prod_{d=1}^D \left[\theta+(1-\theta)N_X^{(d)}\right] \cdot \frac{N_X^{(d)}}{\theta+(1-\theta)N_X^{(d)}} = \tau\cdot\prod_{d=1}^D N_X^{(d)},
\end{equation}
which concludes the proof.
  \end{proof}

\section*{Proof for Proposition 2}

\begin{proof}
  According to the definition, the candidate patches sampled from $\mbox{SPP}(Y,\tau)$ (or $\mbox{SPP}(X,\tau)$) follows a homogeneous Poisson process with intensity $S_Y$ (or $S_X$). Since there exists possibility to generate empty patches, we use intensity $S_X\cdot P(S_{\Box^X}>0)$ for thinning the Poisson process to generate nonempty patches. Given the same budget $\tau$, Proposition~\ref{expectednumber} holds if we can prove the following equality of the two Poisson process intensities
\begin{eqnarray} \label{eq:poissonrateequality1}
  S_Y\cdot P(S_{\pi_{Y,X}(\Box^Y)}>0) = S_X\cdot P(S_{\Box^X}>0).
\end{eqnarray}
  Due to the independence of dimensions, $P(S_{\Box^X}>0)$ can be rewritten as
\begin{eqnarray} \label{eq_dimensiondecomposition}
  P(S_{\Box^X}>0) = \prod_d P(l_X^{(d)}>0)
\end{eqnarray}
  Assuming $N_X^{(d)} \ge 2$, we have
\begin{eqnarray} \label{eq_nonemptypatch}
  P(l^{(d)}_X>0) = \frac{1}{N_X^{(d)}} + \frac{N_X^{(d)}-1}{N_X^{(d)}} \cdot (1-\theta) =  \frac{1}{N_X^{(d)}} \cdot \left[\theta+N_X^{(d)}(1-\theta)\right].
\end{eqnarray}

  W.l.o.g, we assume that the two arrays, $X$ and $Y$, have the same shape apart from the $d'$th dimension where $Y$ has one additional column (the general case of more columns follows by induction), then $S_X/S_Y = N_X^{(d')}/N_Y^{(d')}$.

  There are two cases to consider: (1) $X$ and $Y$ share the terminal boundary in the $d'$th dimension; (2) $X$ and $Y$ share the initial boundary in the $d'$th dimension. In either case, by independence of dimensions, we have
\begin{eqnarray} \label{eq_nonemptypatch2}
  P(S_{\pi_{Y,X}(\Box^Y)}>0) = P(\pi_{Y,X}(l^{(d')}_Y)>0) \cdot \prod_{d\neq d'} P(l_Y^{(d)}>0).
\end{eqnarray}

  In case (1) where $X$ and $Y$ share the terminal boundary in the $d'$th dimension, we have
\begin{eqnarray} \label{eq_patchcross}
  P(\pi_{Y,X}(l^{(d')}_Y)>0) = \frac{\theta}{N_Y^{(d')}} + \frac{N_X^{(d')}}{N_Y^{(d')}} \cdot (1-\theta) =  \frac{1}{N_Y^{(d')}} \cdot \left[\theta+N_X^{(d')}(1-\theta)\right],
\end{eqnarray}
  where the first term $\frac{\theta}{N_Y^{(d')}}$ corresponds to the case that the initial position is on the boundary of $Y$ and $l^{(d')}_Y \geq 2$ (otherwise $\Box^Y$ will not cross into $X$); while the second term $\frac{N_X^{(d')}}{N_Y^{(d')}} \cdot (1-\theta)$ corresponds to the cases that the initial position is not on the boundary of $Y$. By combining Eqs.~(\ref{eq_dimensiondecomposition}), (\ref{eq_nonemptypatch}), (\ref{eq_nonemptypatch2}) and (\ref{eq_patchcross}), we have
\begin{eqnarray}
  \frac{P(S_{\pi_{Y,X}(\Box^Y)}>0)}{P(S_{\Box^X}>0)} = \frac{P(\pi_{Y,X}(l^{(d')}_Y)>0)}{P(l^{(d')}_X>0)} = \frac{N_X^{(d')}}{N_Y^{(d')}}=\frac{S_X}{S_Y}.
\end{eqnarray}
  Thus we have $S_Y\cdot P(S_{\pi_{Y,X}(\Box^Y)}>0) = S_X\cdot P(S_{\Box^X}>0)$.

  In case (2) where $X$ and $Y$ share the initial boundary in the $d'$th dimension, we have
\begin{eqnarray} \label{eq_patchcross_share}
  P(\pi_{Y,X}(l^{(d')}_Y)>0) = \frac{1}{N_Y^{(d')}}+\frac{N_X^{(d')}-1}{N_Y^{(d')}} \cdot (1-\theta) = \frac{1}{N_Y^{(d')}} \cdot \left[\theta+N_X^{(d')}(1-\theta)\right].
\end{eqnarray}
  The conclusion can be similarly derived.
\end{proof}

Because of the same Poisson process intensity Eq.~(\ref{eq:poissonrateequality1}), the following equality also holds
  \begin{equation} \label{k_tau_m_k1}
   P^Y_{K_{\tau}, \{m_k\}_k}\left(\pi_{Y,X}^{-1}\left(K_{\tau}^{X}, \{m_{k}^{X}\}_{k=1}^{K_{\tau}^{X}}\right)\right)= P^X_{K_{\tau}, \{m_k\}_k}\left(K_{\tau}^{X}, \{m_{k}^{X}\}_{k=1}^{K_{\tau}^{X}}\right)
\end{equation}

\section*{Proof for Proposition 3}

\begin{proof}
  W.l.o.g, we assume that the two arrays, $X$ and $Y$, have the same shape apart from the $d'$th dimension where $Y$ has one additional column (the general case follows by induction). For dimensions $d \neq d'$, it is obvious that the law of patches are consistent under projection because the projection is the identity. Given the same budget $\tau$, Proposition~\ref{expecteddistribution} holds if we can prove the following equality
\begin{equation} \label{eq19}
   P^Y_{u}\left(\pi_{Y,X}^{-1}(u^{(d')}_{X}) \bigm| |\pi_{Y,X}(u^{(d')}_{Y})| \geq 1\right) =  P^X_{u}(u^{(d')}_{X} \bigm| |u^{(d')}_{X}|\geq 1),
\end{equation}
  where $u^{(d')}_{X}$ indicates the initial position, $s^{(d')}_{X}$, and the side-length, $l^{(d')}_{X}$, of the $d'$th side of $\Box^X$; $|u^{(d')}_{X}|\geq 1$ means that there is at least one ``1'' entry in $u^{(d')}_{X}$ as $\Box^X$ is a \emph{nonempty} patch.

  There are two cases to consider: (1) $X$ and $Y$ share the initial boundary in the $d'$th dimension; (2) $X$ and $Y$ share the terminal boundary in the $d'$th dimension. In each case, there are two cases (denoted as A \& B in the following) regarding whether the terminal/initial (for Case 1/2, respectively) position locates at the boundary of $X$. In total we have four cases to discuss as follows.

  In case (1) where $X$ and $Y$ share the initial boundary, $|\pi_{Y,X}(u^{(d')}_{Y})| \geq 1$ implies the condition of $s_Y^{(d')}\in X$ because $s_Y^{(d')}\notin X$ would lead to $|\pi_{Y,X}(u^{(d')}_{Y})|=0$. Thus, the left term of Eq. (\ref{eq19}) can be expressed as
\begin{eqnarray} \label{eq21}
   P^Y_{u}\left(\pi_{Y,X}^{-1}(u^{(d')}_{X}) \bigm| |\pi_{Y,X}(u^{(d')}_{Y})|\geq 1\right) = P^Y_{s,l}\left((s^{(d')}_{Y}, l^{(d')}_{Y}) \bigm| s^{(d')}_{Y}\in X\right) \nonumber\\
   = P^Y_{s}\left(s^{(d')}_Y \bigm| s^{(d')}_Y\in X\right)P^Y_{l}\left(l^{(d')}_Y \bigm| s^{(d')}_Y, s^{(d')}_Y\in X\right).
\end{eqnarray}
  For convenience of notation, we let $\theta_\dag = P^Y_{s}(s^{(d')}_Y \bigm| s^{(d')}_Y\in X)$, specifically $\theta_\dag = \frac{1}{\theta+(1-\theta)N^{(d')}_X}$ if $s^{(d')}_Y=1$; $\theta_\dag = \frac{1-\theta}{\theta+(1-\theta)N^{(d')}_X}$ if $s^{(d')}_Y>1$.

  [Case 1.A] For $0<s^{(d')}_X+l^{(d')}_X-1<N_X^{(d')}< N_Y^{(d')}$,
\begin{equation}
   P^Y_{u}\left(\pi_{Y,X}^{-1}(u^{(d')}_{X}) \bigm| |\pi_{Y,X}(u^{(d')}_{Y})|\geq 1\right) = \theta_\dag \theta^{l^{(d')}_X-1}(1-\theta) =  P^X_{u}(u^{(d')}_{X} \bigm| |u^{(d')}_{X}|\geq 1).
\end{equation}

  [Case 1.B] For $0<s^{(d')}_X+l^{(d')}_X-1=N_X^{(d')}< N_Y^{(d')}$,
\begin{eqnarray}
\begin{split}
   &P^Y_{u}\left(\pi_{Y,X}^{-1}(u^{(d')}_{X}) \bigm| |\pi_{Y,X}(u^{(d')}_{Y})| \geq 1\right)\\
   = &P^Y_{s}(s^{(d')}_Y \bigm| s^{(d')}_Y\in X)P^Y_{l}(l^{(d')}_Y=l^{(d')}_X \bigm| s^{(d')}_Y, s^{(d')}_Y\in X)\\
   &+ P^Y_{s}(s^{(d')}_Y \bigm| s^{(d')}_Y\in X)P^Y_{l}(l^{(d')}_Y=l^{(d')}_X+1 \bigm| s^{(d')}_Y, s^{(d')}_Y\in X)\\
  = & \theta_\dag \theta^{l^{(d')}_X-1}(1-\theta)+\theta_\dag \theta^{l^{(d')}_X} = \theta_\dag \theta^{l^{(d')}_X-1} =  P^X_{u}(u^{(d')}_{X} \bigm| |u^{(d')}_{X}| \geq 1)
\end{split}
\end{eqnarray}

  In case (2) where $X$ and $Y$ share the terminal boundary, $|\pi_{Y,X}(u^{(d')}_{Y})| \geq 1$ implies the condition of $s^{(d')}_{Y} \cdot l^{(d')}_{Y} > 1$ because $(s^{(d')}_{Y}=1, l^{(d')}_{Y}=1)$ would lead to $|\pi_{Y, X}(u_Y^{(d')})|=0$.
  Thus, the left term of Eq.~(\ref{eq19}) can be expressed as
\begin{eqnarray} \label{eq22}
   P^Y_{u}\left(\pi_{Y,X}^{-1}(u^{(d')}_{X}) \bigm| |\pi_{Y,X}(u^{(d')}_{Y})| \geq 1\right) =P^Y_{s,l}\left((s^{(d')}_{Y}, l^{(d')}_{Y})\bigm|s^{(d')}_{Y} \cdot l^{(d')}_{Y} > 1\right).
\end{eqnarray}
  Given the condition of $s^{(d')}_{Y} \cdot l^{(d')}_{Y} > 1$ and the assumption $N_Y^{(d')} = N_X^{(d')} + 1$, we have
  \begin{eqnarray}
&&  P^Y_{s}\left(s^{(d')}_Y =  1 \bigm| s^{(d')}_{Y} \cdot l^{(d')}_{Y} > 1\right)=\frac{P^Y_{s,l}(s^{(d')}_Y =  1, l^{(d')}_{Y}\ge 2)}{P^Y_{s,l}(s^{(d')}_{Y} \cdot l^{(d')}_{Y} > 1)} \nonumber\\
  &=&\frac{1\cdot\theta}{1+(N_Y^{(d')}-1)(1-\theta)-{1\cdot(1-\theta)}}=  \frac{\theta}{\theta+(1-\theta)N^{(d')}_X}
  \end{eqnarray}
  and
  \begin{eqnarray}
&&  P^Y_{s}\left(s^{(d')}_Y =  2 \bigm| s^{(d')}_{Y} \cdot l^{(d')}_{Y} > 1 \right)= \frac{P^Y_{s}(s^{(d')}_Y = 2)}{P^Y_{s,l}(s^{(d')}_{Y} \cdot l^{(d')}_{Y} > 1)} \nonumber\\
  &=&\frac{1-\theta}{1+(N_Y^{(d')}-1)(1-\theta)-{1\cdot(1-\theta)}}=  \frac{1-\theta}{\theta+(1-\theta)N^{(d')}_X}.
  \end{eqnarray}

  [Case 2.A] For $\pi_{Y,X}(s^{(d')}_Y)=1$, we have $s^{(d')}_X=1$,
\begin{eqnarray}
\begin{split}
   & P^Y_{u}\left(\pi_{Y,X}^{-1}(u^{(d')}_{X})\bigm||\pi_{Y,X}(u^{(d')}_{Y})|\geq1\right) =P^Y_{s,l}\left((s^{(d')}_{Y}, l^{(d')}_{Y})\bigm| s^{(d')}_{Y} \cdot l^{(d')}_{Y} > 1\right)\\
   = &  P^Y_{s}(s^{(d')}_Y =  2 \bigm| s^{(d')}_{Y} \cdot l^{(d')}_{Y} > 1)P^Y_{l}(l^{(d')}_Y=l^{(d')}_X \bigm| s^{(d')}_Y=2)\\
 & +  P^Y_{s}(s^{(d')}_Y =  1 \bigm| s^{(d')}_{Y} \cdot l^{(d')}_{Y} > 1)P^Y_{l}(l^{(d')}_Y=l^{(d')}_X+1 \bigm| s^{(d')}_Y=1)\\
  = & \frac{1-\theta}{\theta+(1-\theta)N^{(d')}_X}\cdot\theta^{l^{(d')}_X-1}\theta_\ddag + \frac{\theta}{\theta+(1-\theta)N^{(d')}_X}\cdot\theta^{l^{(d')-1}_X}\theta_\ddag\\
  =&\frac{\theta^{l^{(d')}_X-1}\theta_\ddag}{\theta+(1-\theta)N^{(d')}_X}= P^X_{u}(u^{(d')}_{X}\bigm| |u^{(d')}_{X}|\geq1),
\end{split}
\end{eqnarray}
  where $\theta_\ddag = 1$ if $0<\pi_{Y,X}(s^{(d')}_Y)+l^{(d')}_X-1=N_X^{(d')}$; $\theta_\ddag = 1-\theta$ if $0<\pi_{Y,X}(s^{(d')}_Y)+l^{(d')}_X-1<N_X^{(d')}$.

  [Case 2.B] For $\pi_{Y,X}(s^{(d')}_Y)>1$, we have  $s^{(d')}_X>1$,
\begin{eqnarray}
&&   P^Y_{u}\left(\pi_{Y,X}^{-1}(u^{(d')}_{X})\bigm| |\pi_{Y,X}(u^{(d')}_{Y})|\geq1\right) = \frac{1-\theta}{\theta+(1-\theta)N^{(d')}_X}\cdot\theta^{l^{(d')}_X-1}\theta_\ddag \nonumber \\
   &=&  P^X_{u}(u^{(d')}_{X} \bigm| |u^{(d')}_{X}|\geq1).
\end{eqnarray}

  Consider all $D$ dimensions, for each case, we have $ P^Y_{\Box}(\pi_{Y,X}^{-1}(\Box^X) \bigm| S_{\pi_{Y,X}(\Box^Y)}>0) =  P^X_{\Box}(\Box^X \bigm| S_{\Box^X}>0)$.
\end{proof}

\section*{Detail Proof of $ P^Y_\boxplus(\pi_{Y,X}^{-1}(\boxplus_X))= P^X_\boxplus(\boxplus_X)$}

\begin{eqnarray}
  & &  P^Y_\boxplus(\pi_{Y,X}^{-1}(\boxplus_X))=  P^Y_\boxplus\left(\pi_{Y,X}^{-1}\left(K_{\tau}^{X}, \{m_{k}^{X},\Box_{k}^{X}\}_{k=1}^{K_{\tau}^{X}}\right)\right) \nonumber \\
  &=&  P^Y_{K_{\tau}, \{m_k\}_k}\left(\pi_{Y,X}^{-1}\left(K_{\tau}^{X}, \{m_{k}^{X}\}_{k=1}^{K_{\tau}^{X}}\right)\right) \nonumber \\
  &&\cdot P^Y_{\Box}\left(\pi_{Y,X}^{-1}\left(\{\Box_{k}^{X}\}_{k=1}^{K_{\tau}^{X}}|K_{\tau}^{X}, \{m_{k}^{X}\}_{k=1}^{K_{\tau}^{X}}\right) \bigm| S_{\pi_{Y,X}(\Box^Y)}>0 \right) \label{app1} \\
  &=&  P^Y_{K_{\tau}, \{m_k\}_k}\left(\pi_{Y,X}^{-1}\left(K_{\tau}^{X}, \{m_{k}^{X}\}_{k=1}^{K_{\tau}^{X}}\right)\right) \nonumber \\
  &&\cdot P^Y_{\Box}\left(\pi_{Y,X}^{-1}\left(\{\Box_{k}^{X}\}_{k=1}^{K_{\tau}^{X}}|K_{\tau}^{X}\right) \bigm| S_{\pi_{Y,X}(\Box^Y)}>0 \right) \label{app2} \\
  &=&  P^Y_{K_{\tau}, \{m_k\}_k}\left(\pi_{Y,X}^{-1}\left(K_{\tau}^{X}, \{m_{k}^{X}\}_{k=1}^{K_{\tau}^{X}}\right)\right)
   \nonumber\\
  &&\cdot\prod_{k=1}^{K_{\tau}^{X}} P^Y_{\Box}\left(\pi_{Y,X}^{-1}\left(\Box_{k}^{X} \right) \bigm| S_{\pi_{Y,X}(\Box^Y)}>0 \right) \label{app3} \\
  &=&  P^X_{K_{\tau}, \{m_k\}_k}\left(K_{\tau}^{X}, \{m_{k}^{X}\}_{k=1}^{K_{\tau}^{X}}\right) \nonumber\\
  &&
  \cdot\prod_{k=1}^{K_{\tau}^{X}} P^Y_{\Box}\left(\pi_{Y,X}^{-1}\left(\Box_{k}^{X}\right)
  \bigm| S_{\pi_{Y,X}(\Box^Y)}>0 \right) \label{app4} \\
  &=&  P^X_{K_{\tau}, \{m_k\}_k}\left(K_{\tau}^{X}, \{m_{k}^{X}\}_{k=1}^{K_{\tau}^{X}}\right) \cdot\prod_{k=1}^{K_{\tau}^{X}} P^X_{\Box}\left(\Box_{k}^{X} \bigm| S_{\Box^X}>0 \right) \label{app5} \\
  & =&  P^X_\boxplus \left(K_{\tau}^{X}, \{m_{k}^{X},\Box_{k}^{X}\}_{k=1}^{K_{\tau}^{X}}\right) =  P^X_\boxplus(\boxplus_X). \nonumber
\end{eqnarray}
  We can obtain Eq.~(\ref{app2}) from Eq.~(\ref{app1}) because
\begin{equation}
  P\left(\{m_{k},\Box_{k}\}_{k=1}^{K_{\tau}}|K_{\tau}\right)= P\left(\{m_{k}\}_{k=1}^{K_{\tau}}|K_{\tau}\right)\cdot P\left(\{\Box_{k}\}_{k=1}^{K_{\tau}}|K_{\tau}\right) \nonumber
\end{equation}
  which indicates
\begin{equation}
  P\left(\{\Box_{k}\}_{k=1}^{K_{\tau}}|K_{\tau},\{m_{k}\}_{k=1}^{K_{\tau}}\right)= P\left(\{\Box_{k}\}_{k=1}^{K_{\tau}}|K_{\tau}\right) \nonumber
\end{equation}
  We can obtain Eq.~(\ref{app3}) from Eq.~(\ref{app2}) because of independence of patches. Eq.~(\ref{app4}) is derived from Eq.~(\ref{app3}) by applying Proposition 2 and Eq.~(\ref{app5}) is derived from Eq.~(\ref{app4}) by applying Proposition 3.

\section*{Multiple-Try Metropolis for Sampling $\{r_i\}_i, \{c_j\}_{j}$ (in Section 5.2)}

The prior distributions of $\{r_i\}_i$ and $\{c_j\}_{j}$ are discrete uniform distributions over all the $N!$ permutations. To cooperate with the C-SMC sampler for higher acceptance ratio, we adopt the Multiple-Try Metropolis method~\cite{liu2000multiple} for sampling the row and column indices of the relational data. For each $r_i$, we propose an exchange between $r_i$ and $Z$ proposal rows $\{r_{i'_z}\}_{z=1}^Z$, which are randomly chosen. The detail for sampling $\{r_i\}_i$ is summarized in Algorithm~\ref{mtm_ranks} (sampling $\{c_j\}_j$ is similar).
%Since the prior distribution and the proposal distribution are identical, the acceptance ratio $\min(1,\alpha)$ of the exchange is simply a likelihood ratio
%\begin{equation}
%  \alpha = \frac{\prod_{j} \ell(r_{i}^*,c_j,\rho_{ij}) \ell(r_{i'}^*,c_j,\rho_{{i'}j})}
%  {\prod_{j} \ell(r_i,c_j,\rho_{ij}) \ell(r_{i'},c_j,\rho_{{i'}j})}
%\end{equation}
%  where $r^*_i = r_{i'}$ and $r^*_{i'} = r_i$ (row index exchange).

\begin{algorithm}[h]
\caption{Multiple-Try Metropolis for Sampling $\{r_i\}_i$}
\label{mtm_ranks}
\begin{algorithmic}
  \REQUIRE Relational data $R$, number of proposals $Z$, $\{\rho_{ij}\}_{i,j}$, $\{r_i\}_i$ and $\{c_j\}_j$
  \ENSURE New assignments of $\{r_i\}_i$
  \FOR{$i=1,\cdots, N$}
    \STATE Propose $Z$ independent proposal indices $\{r_{i'_z}\}_{z=1}^Z$ for exchanging with $r_i$;
    \STATE Compute the weights $\{\omega(r_{i'_z})\}_{z=1}^Z$ as
    \begin{eqnarray}
    \omega(r_{i'_z}) = \frac{\prod_{j} \ell(r_{i'_z},c_j,\rho_{ij}) \ell(r_i,c_j,\rho_{{i'_z}j})}
    {\prod_{j} \ell(r_i,c_j,\rho_{ij}) \ell(r_{i'_z},c_j,\rho_{{i'_z}j})};
\end{eqnarray}
    \STATE Sample $r_{i^*}$ from $\{r_{i'_z}\}_{z=1}^Z$ with probability in proportional to $\{\omega(r_{i'_z})\}_{z=1}^Z$;
    \STATE Suppose $r_i$ and $r_{i^*}$ are exchanged; propose $Z-1$ new independent proposal indices $\{r_{i''_z}\}_{z=1}^{Z-1}$ and set $r_{i''_{Z}}=r_i$ for exchanging with $r_{i^*}$;
    \STATE Compute the weights $\{\omega(r_{i''_z})\}_{z=1}^Z$ as
    \begin{eqnarray}
  \omega(r_{i''_z}) = \frac{\prod_{j} \ell(r_{i''_z},c_j,\rho_{ij}) \ell(r_{i^*},c_j,\rho_{{i''_z}j}) }
  {\prod_{j} \ell(r_{i^*},c_j,\rho_{ij}) \ell(r_{i''_z},c_j,\rho_{{i''_z}j})}
\end{eqnarray}
    \STATE Accept the exchange between $r_i$ and $r_{i^*}$ with the ratio
    \begin{eqnarray}
    \alpha = \min\left(1, \frac{\omega(r_{i'_1})+\cdots+\omega(r_{i'_Z})}{\omega(r_{i''_1})+\cdots+\omega(r_{i''_Z})}\right)
    \end{eqnarray}
  \ENDFOR
\end{algorithmic}
\end{algorithm}

\end{document}